\documentclass[12pt]{article}

\usepackage{amssymb}
\usepackage{amsmath}
\usepackage{amscd}
\usepackage{latexsym}
\usepackage{graphicx}

\usepackage{enumerate}

\numberwithin{equation}{section}
\usepackage{cite}
\usepackage{enumitem}

\newcommand{\be}{\begin{equation}}
\newcommand{\ee}{\end{equation}}
\newcommand{\Dlt}{\Delta}
\newcommand{\dlt}{\delta}
\newcommand{\cH}{{\cal H}}
\newcommand{\cP}{{\cal P}}

\newcommand{\ra}{\rightarrow}

\newcommand{\bt}{\beta}
\newcommand{\vp}{\varphi}
\newcommand{\ep}{\varepsilon}
\newcommand{\al}{\alpha}

\newcommand{\prt}{\partial}
\newcommand{\gm}{\gamma}

\newcommand{\lbd}{\lambda}

\newcommand{\rgl}{\rangle}
\newcommand{\lgl}{\langle}

\begin{document}

\vskip 2mm

{\it Review}

\begin{center}

{\Large{\bf Quantum Operation of Affective Artificial Intelligence} \\ [5mm]

V.I. Yukalov } \\ [3mm]

{\it
Bogolubov Laboratory of Theoretical Physics, \\
Joint Institute for Nuclear Research, Dubna 141980, Russia \\ 

and \\

Instituto de Fisica de S\~ao Carlos, Universidade de S\~ao Paulo, \\
CP 369,  S\~ao Carlos 13560-970, S\~ao Paulo, Brazil}  \\ [2mm]

{\bf e-mail}: yukalov@theor.jinr.ru  \\ [3mm]

\end{center}

\vskip 2cm

\begin{abstract}

The review analyzes the fundamental principles which Artificial Intelligence should 
be based on in order to imitate the realistic process of taking decisions by humans 
experiencing emotions. Two approaches are considered, one based on quantum theory
and the other employing classical terms. Both these approaches have a number of 
similarities, being principally probabilistic. The analogies between quantum 
measurements under intrinsic noise and affective decision making are elucidated. 
It is shown that cognitive processes have many features that are formally similar 
to quantum measurements. This, however, in no way means that for the imitation of 
human decision making Affective Artificial Intelligence has necessarily to rely on 
the functioning of quantum systems. The analogies between human decision making and 
quantum measurements merely demonstrate formal common properties in their functioning. 
It is in this sense that one has to understand quantum operation of Artificial 
Intelligence. Appreciating the common features between quantum measurements and 
decision making helps for the formulation of an axiomatic approach employing only 
classical notions. Artificial Intelligence, following this approach, operates 
similarly to humans, by taking into account the utility of the considered alternatives 
as well as their emotional attractiveness. Affective Artificial Intelligence, whose 
operation takes account of the cognition-emotion duality, avoids numerous behavioural 
paradoxes of traditional decision making. A society of intelligent agents, interacting 
through the repeated multistep exchange of information, forms a network accomplishing 
dynamic decision making based on the evaluation of utility and affected by the emotional 
attractiveness of alternatives. The considered intelligent networks can characterize 
the operation of either a human society of affective decision makers, or the brain 
composed of neurons, or a typical probabilistic network of an artificial intelligence.   
\end{abstract}

\vskip 1cm
{\parindent=0pt
{\bf Keywords}: artificial intelligence, quantum measurements, quantum intrinsic noise, 
affective decision making, cognition-emotion duality, behavioural paradoxes, 
dynamic decision making, collective decision making, probabilistic networks}

\newpage

{\parindent=0pt
{\bf {\Large Contents} }

\vskip 5mm
{\bf 1}. Introduction

\vskip 3mm
{\bf 2}. Measurements under intrinsic noise

\vskip 2mm
\hspace{5mm}   2.1. Quantum algebra of events

\vskip 2mm
\hspace{5mm}   2.2. Operationally testable events 

\vskip 2mm
\hspace{5mm}   2.3. Modes of intrinsic noise

\vskip 2mm
\hspace{5mm}   2.4. Noise-decorated alternatives

\vskip 2mm
\hspace{5mm}   2.5. Quantum probability space

\vskip 2mm
\hspace{5mm}   2.6. Quantum-classical correspondence

\vskip 2mm
\hspace{5mm}   2.7. Probability of superposition states

\vskip 2mm
\hspace{5mm}   2.8. Alternative-noise entanglement

\vskip 2mm
\hspace{5mm}   2.9. Entanglement production by measurements

\vskip 2mm
\hspace{5mm}   2.10. Time dependence of probability

\vskip 2mm
\hspace{5mm}   2.11. Quantum state reduction

\vskip 2mm
\hspace{5mm}   2.12. Consecutive measurements of alternatives

\vskip 2mm
\hspace{5mm}   2.13. Immediate consecutive measurements

\vskip 2mm
\hspace{5mm}   2.14. Synchronous noiseless measurements

\vskip 2mm
\hspace{5mm}   2.15. Synchronous measurements under noise

\vskip 2mm
\hspace{5mm}   2.16. Swap order relations

\vskip 2mm
\hspace{5mm}   2.17. Quantum versus classical probabilities

\vskip 2mm
\hspace{5mm}   2.18. Quantum decision theory

\vskip 3mm
{\bf 3}. Affective decision making

\vskip 2mm
\hspace{5mm}   3.1. Evolutionary origin of emotions

\vskip 2mm
\hspace{5mm}   3.2. Problems in decision making

\vskip 2mm
\hspace{5mm}   3.3. Behavioural probabilities of alternatives

\vskip 2mm
\hspace{5mm}   3.4. Quantification of utility factor

\vskip 2mm
\hspace{5mm}   3.5. Magnitude of attraction factor

\vskip 2mm
\hspace{5mm}   3.6. Multiple attraction factors 

\vskip 2mm
\hspace{5mm}   3.7. Problems in classifying attractiveness

\vskip 2mm
\hspace{5mm}   3.8. Explicit attraction factors

\vskip 2mm
\hspace{5mm}   3.9. Buridan's donkey problem

\vskip 2mm
\hspace{5mm}   3.10. Kahneman-Tversky lotteries

\vskip 2mm
\hspace{5mm}   3.11. Verification of quarter law
 
\vskip 2mm
\hspace{5mm}   3.12. Contextuality of attraction factors

\vskip 2mm
\hspace{5mm}   3.13. Choice between bundled alternatives

\vskip 2mm
\hspace{5mm}   3.14. Quantum versus classical consciousness

\vskip 3mm
{\bf 4}. Resolution of behavioural paradoxes

\vskip 2mm
\hspace{5mm}   4.1. St. Petersburg paradox

\vskip 2mm
\hspace{5mm}   4.2. Martingale illusion

\vskip 2mm
\hspace{5mm}   4.3. Allais paradox

\vskip 2mm
\hspace{5mm}   4.4. Independence paradox

\vskip 2mm
\hspace{5mm}   4.5. Ellsberg paradox

\vskip 2mm
\hspace{5mm}   4.6. Prisoner dilemma

\vskip 2mm
\hspace{5mm}   4.7. Disjunction effect

\vskip 2mm
\hspace{5mm}   4.8. Conjunction fallacy

\vskip 2mm     
\hspace{5mm}   4.9. Disposition effect

\vskip 2mm     
\hspace{5mm}   4.10. Ariely paradox

\vskip 2mm
\hspace{5mm}   4.11. Decoy effect

\vskip 2mm
\hspace{5mm}   4.12. Planning paradox

\vskip 2mm
\hspace{5mm}   4.13. Preference reversal

\vskip 2mm
\hspace{5mm}   4.14. Preference intransitivity 

\vskip 2mm
\hspace{5mm}   4.15. Order effects

\vskip 3mm
{\bf 5}. Networks of intelligent agents

\vskip 2mm
\hspace{5mm}   5.1. Multistep decision making 

\vskip 2mm
\hspace{5mm}   5.2. Types of interactions and memory

\vskip 2mm
\hspace{5mm}   5.3. Networks with uniform memory 

\vskip 2mm
\hspace{5mm}   5.4. Network with mixed memory

\vskip 2mm
\hspace{5mm}   5.5. Dynamic regimes of preferences

\vskip 2mm
\hspace{5mm}   5.6. Attenuation of emotion influence

\vskip 2mm
\hspace{5mm}   5.7. Continuous decision making

\vskip 2mm
\hspace{5mm}   5.8. Discrete versus continuous processes

\vskip 2mm
\hspace{5mm}   5.9. Time discounting of utility

\vskip 2mm
\hspace{5mm}   5.10. Collective network operation

\vskip 3mm
{\bf 6}. Conclusion

}

\vskip 1cm

\section{Introduction}

Artificial Intelligence is understood as intelligence demonstrated by machines, 
as opposed to natural intelligence displayed by animals including humans. The main
Artificial Intelligence textbooks define the field as the study of artificial 
intelligent systems perceiving the information obtained from the environment and 
taking decisions and actions for the goal attainment 
\cite{Nilsson_1,Luger_2,Rich_3,Poole_4,Neapolitan_5,Russell_6}. There is wide 
agreement among artificial intelligence researchers that to be called intelligence, 
it is required to be able to use logical strategy and make judgments under 
uncertainty.

A system possessing intelligence is termed an intelligent agent. That system, 
evaluating the available information, is able to take autonomous actions and 
decisions directed to the achievement of the desired goals and may improve its 
performance with learning or using obtained knowledge 
\cite{Nilsson_1,Luger_2,Rich_3,Poole_4,Neapolitan_5,Russell_6}. Often, the term 
intelligent agent is applied to systems possessing artificial intelligence. 
However the intelligent agent paradigm is closely related and employed with respect 
to agents in economics, in cognitive science, ethics, philosophy, as well as in many 
interdisciplinary socio-cognitive modeling and simulations. Generally, from the 
technical or mathematical point of view, the notion of intelligent agent can be 
associated with either real or artificial intelligence. An intelligent agent could 
be anything that makes decisions, as a person, firm, machine, or software.

In this review, we concentrate on one of the most difficult and important problems 
of Artificial Intelligence, that is on the mechanism of taking decisions similarly 
to this process in humans, whose decisions practically always are accompanied 
by emotions. The achievement of human-level machine intelligence has been a principal 
goal from the beginning of works on Artificial Intelligence 
\cite{Nilsson_1,Luger_2,Rich_3,Poole_4,Neapolitan_5,Russell_6}. The key-point of 
the present review is the description of how affective decision making could be 
mathematically formalized to the level sufficient for the functioning of Artificial 
Intelligence imitating human decision processes in which emotions are an inevitable 
part. Below, talking about Artificial Intelligence we keep in mind Affective 
Artificial Intelligence. 

In order to formulate the basic operational algorithms of Affective Artificial 
Intelligence, it is necessary to develop a mathematical description of human 
affective decision making. The problem of emotion quantification consists of two 
sides. One side is the assessment of emotions experienced by a subject as reactions 
on external events, e.g., hearing voice or looking at pictures. The arising emotions 
can include happiness, anger, pleasure, disgust, fear, sadness, astonishment, pain, 
and so on. The severity or intensity of such emotions can be estimated by studying 
the expressive forms manifesting themselves in motor reactions, such as facial 
expressions, pantomime, and general motor activity, and by measuring physiological 
reactions, such as the activity of the sympathetic and parasympathetic parts of the 
autonomic nervous system, as well as the activity of the endocrine glands. Vegetative 
manifestations of emotions can be noticed by studying changes in the electrical 
resistance of the skin, the frequency and strength of heart contractions, blood pressure, 
skin temperature, hormonal and chemical composition of the blood, and like that. 
There exists a vast literature on the methods of emotion detection and appraisal in 
speech, facial expressions, and body gestures \cite{Moors_2019,Mumenthaler_8}. The study 
and development of systems and devices that can recognize, interpret, process, and 
simulate human affects is named Affective Computing \cite{Picard_9,Calvo_10}. These 
problems will not be touched in the review.

The other side of the story is the challenge of characterizing how emotions influence
decision making. To formulate the principles of functioning of Affective Artificial 
Intelligence in the process of taking decisions, it is necessary to be able to quantify 
the role of emotions in this process. It is this objective that is in the center of 
the present review.
 
This goal confronts the basic problem of how emotions, arising in the process of 
decision making, could be defined and quantified. It seems to be too difficult, if 
possible at all, to develop a formalized quantification of emotions allowing for the 
selection, in the presence of emotions, of an optimal alternative in the cognitive 
process of making decisions. The mathematical description of emotion influence in 
the process of decision making is the hard problem that has not found yet a 
comprehensive solution \cite{Lerner_11}.  

Difficulties start with the fact that there is no a unique generally accepted 
definition of what is emotion as compared to cognition. It is possible to mention 
the long-standing dispute about whether emotion is primary and independent of cognition 
\cite{Zajonc_12,Zajonc_13}, or secondary and always dependent upon cognition 
\cite{Lazarus_14,Lazarus_15}, although there is the point of view that this dispute 
is largely semantic, being induced by dissimilar definitions \cite{Leventhal_16}. 

The studies on brain organization often support the assumption that there is a 
considerable degree of functional specialization and that many regions of brain can 
be conceptualized as either affective or cognitive. Popular examples are the amygdala 
in the domain of emotion and the lateral prefrontal cortex in the case of cognition. 
However, there are arguments \cite{Pessoa_17,Brosch_18} that complex 
cognitive-emotional behaviours have their basis in dynamic coalitions of networks 
of brain areas, none of which should be conceptualized as specifically affective or 
cognitive. Different brain areas exhibit a high degree of connectivity for regulating 
the flow and integration of information between brain regions, which results in the 
intense cognitive-emotional interactions. Usually, under ``emotions" one understands 
just a placeholder for something much broader than emotions in a narrow sense, 
including affective processes in general \cite{Wentura_19}. There are arguments that 
the notions of emotion, cognition, and the related phenomena can be more precisely 
defined in a functional framework, for example in terms of behavioural principles 
\cite{Houwer_20}, with respect to emotion taxonomy \cite{Keltner_21}, to emotion 
regulation \cite{Gyurak_22}, or studying the emotion appraisal during the dynamics 
of the emotion process \cite{Scherer_2009,Scherer_2019,Moors_2019}.
More references on the definition of emotions and their relation to cognition can 
be found in the surveys \cite{Koole_26,Koole_27,Reisenzein_28}. 

The functional framework keeps in mind the operational separation of cognition and 
emotion as the notions related to the process of decision making that comprises two 
sides, reasoning and affective \cite{Lerner_11,Schwarz_29}. Under the {\it reasoning} 
side one means the ability of formulating explicit rules allowing for a {\it normative} 
choice. And the {\it affective} side implies the possibility of making a choice being 
influenced by emotions that not always allow for explicit formal prescriptions. The 
reasoning-affective dichotomy in decision making is often called rational-irrational 
duality \cite{Ariely_30}. As is explained above, there is no strictly speaking uniquely 
defined and absolutely separated notions of cognitive and affective, as well as of 
rational and irrational. However, our goal is not to plunge into semantic debates, 
but to describe an approach taking into account two aspects of decision making, normative 
allowing for the explicit evaluation of utility and affective that seems to avoid the 
characterization by prescribed formal rules. The kaleidoscope of emotions can be quite 
ramified and not allowing for sharp categorical definitions, because of which it is 
labeled \cite{Scherer_2009,Scherer_2019,Moors_2019} as idiosyncratic and fuzzy. This
fuzziness is the main obstacle in the attempts of quantifying the influence of 
emotions on decision making. 

Thus the principal difference between a standard programmed robot or computer and 
human-type intelligence is the cognition-emotion duality of human consciousness in 
the process of taking decisions. For clarity, one can talk about human intelligence, 
although the same duality in decision making is typical of practically all alive beings, 
as numerous empirical studies prove. The animals likely feel a full range of emotions, 
including fear, joy, happiness, shame, embarrassment, resentment, jealousy, rage, anger, 
love, pleasure, compassion, respect, relief, disgust, sadness, despair, and grief 
\cite{Bekoff_31}. 

The cognition-emotion duality of human consciousness, exhibited when taking decisions, 
combines rational conscious evaluation of utility of the intended actions with irrational 
subconscious emotions. The latter are especially noticeable in decisions under risk and 
uncertainty. This duality is the cause of a number of behavioural paradoxes in classical 
decision making, when human actions contradict expected utility theory. So, in order 
to formulate explicit algorithms for the operation of Affective Artificial Intelligence, 
comprising cognition-emotion duality, it is necessary to develop an adequate theory of 
affective decision making that could give realistic predictions under uncertainty.   

The existence of the cognition-emotion duality in decision making hints on the possibility
of its description by resorting to the techniques of quantum theory, in which there also
exists duality, the so-called particle-wave duality \cite{Dittel_31}. Although the nature 
of these notions in physics and decision theory is rather different, but, probably, the 
mathematical techniques of quantum theory could hint on the similar description of both 
phenomena. Bohr \cite{Bohr_32,Bohr_33} was the first to assume that the functioning of 
the human brain could be described by the techniques of quantum theory. Since then, 
there have appeared numerous publications discussing the possibility of directly 
applying quantum techniques for characterizing the process of human decision making. 
These discussions, assuming that consciousness is quantum or quantum-like have been 
summarized in many review works, e.g. 
\cite{Khrennikov_34,Busemeyer_35,Agrawal_36,Bagarello_37,Ashtiani_38,Wendt_38}, where 
numerous references on different attempts of applying quantum techniques to the 
description of consciousness are cited.

It is necessary to accept that many researchers are rather sceptical with regard 
to the parallelism between quantum physics and cognitive processes because of the 
following reasons: 

\vskip 2mm
(i) First of all, according to the current neurophysiological knowledge, the brain 
is in no way a quantum system, hence, it has nothing to do with quantum consciousness. 
The assumption that the brain's neurons act as miniature quantum devices, thus that 
the brain functions similarly to a quantum computer \cite{Penrose_39,Hameroff_40} has 
been justly criticized \cite{Tegmark_41} by showing that decoherence effects do not 
allow for neurons to act as quantum objects. This does not exclude that some quantum 
processes do exist in the brain, which are studied in quantum biophysics 
\cite{Al_42,Jedlicka_43}. Nevertheless, the brain as a whole and its functioning seem 
to have nothing to do with quantum theory. 

\vskip 2mm
(ii) The above objection is usually refuted by saying that the possibility of describing 
human thinking processes by means of quantum theory does not require the assumption that 
human brains are some quantum systems. Instead, it holds that, although the brain is not 
a quantum object, but cognition and the process of human thinking can be mathematically 
formalized into the language of quantum theory. This is similar to the situation presented 
by the theory of differential equations, which was initially developed for describing the 
motion of planets. But now the theory of differential equations is employed everywhere, 
being just an efficient mathematical tool not necessarily related to planet motion. In 
the same way, quantum theory may provide a convenient framework for the mathematical 
description of thinking processes. The critics, however, insist that the analogies are 
superficial, do not prescribe practical recipes, and sometimes even contradict empirical 
data qualitatively \cite{Boyer_44,Boyer_45}.   

\vskip 2mm
(iii) Moreover, the simple logic teaches us that, if the brain is a classical object,
then its functioning should be described by classical equations, since it is exactly
its properties, including functioning, that classify an object as classical or quantum. 
If the properties of an object cannot in principle be described by a classical theory, 
but allow for only quantum description, then this object is quantum, which contradicts 
our current knowledge on the brain.      

\vskip 2mm 
(iv) The direct use of quantum theory for describing decision making introduces a large 
number of unknown parameters and ambiguous notions that cannot be characterized on the 
level of observable quantities associated with decision making. For instance, what is a 
Hamiltonian in psychological processes? How to define and measure numerous coefficients
entering wave functions describing the brain states? What is the evolution equation for 
statistical operators characterizing the brain? And a lot of other ambiguously defined 
notions appear \cite{Zheltikov_46}. 

\vskip 2mm
(v) The most important goal of any theory is the ability of predicting quantitative
results that could be verified in experiment. However, none of the purely quantum 
variants of decision making has ever predicted some numerical data. The maximum what 
can be done is the consideration of particular cases and fitting parameters for the 
assumed interpretation of these cases. In order to extract quantitative information 
from the derived quantum relations, it is necessary to complement them by a number of 
assumptions not related to quantum techniques. In that sense the complicated quantum 
substructure becomes excessive, similarly to the excessiveness of nonlocal hidden 
variables for explaining quantum phenomena \cite{Dakic_47}.         

\vskip 2mm
(vi) The fact that some events in decision making can qualitatively be interpreted as 
due to quantum processes does not exclude the possibility of other interpretations in
terms of classical language. According to the Occam's razor principle, the simplest 
of competing theories has to be preferred to the more complex, so that explanations 
of unknown phenomena should be sought first in terms of known quantities. Hence quite 
complicated theories based on quantum formulas are to be disregarded in favor of much
simpler explanations based on classical notions, provided these exist. Entities should 
not be multiplied beyond necessity. The simplest theory is the best \cite{Sober_48}. 

\vskip 2mm
To understand whether the functioning of consciousness is described by quantum or 
classical rules is important, since, depending on the involved formalism, the operation 
of artificial intelligence has to be characterized in the same language. Examining the
above objections to the use of quantum techniques for the formalization of decision
making, it is possible to say the following: First, although at the present time, the 
influence of quantum effects on the functioning of the brain has not been convincingly 
argued, it cannot be absolutely excluded. Second, even if actual quantum effects play 
no role in the brain operation and consciousness does not need quantum description, the 
investigation of the analogies between decision making and quantum processes can enrich 
both of them suggesting their more profound comprehension. The peculiarities of quantum 
phenomena, that are better conceived, can give hints on the ways of characterizing 
consciousness functioning.          

The point of view advocated in this review can be summarized as follows: The brain is
a classical object, hence its basic property, that is consciousness, by definition, has 
to be classical. Otherwise it would be a meaninglessness to say that a classical object
has quantum properties. Nevertheless, there exists a number of formal analogies in the 
description of quantum measurements and decision making. These analogies need to be 
carefully investigated for two reasons:

\vskip 2mm
(i) Although being formal, the analogies between different phenomena very often suggest 
concrete practical ways for describing these phenomena. 

\vskip 2mm
(ii) Borrowing some ideas from the nominal analogies between two different approaches
helps to compare these approaches and to choose the more efficient and simple theory. 

\vskip 2mm
The formal analogy between quantum and conscious phenomena has been noticed long 
time ago by von Neumann, who mentioned that the quantum theory of measurements can 
be interpreted as decision theory \cite{Neumann_49}. This concept has been developed 
by other researchers, for instance by Benioff \cite{Benioff_50,Benioff_51}. Thus 
{\it quantum measurement} is analogous to {\it decision making}, hence the 
{\it measurement of an observable} is similar to the {\it choice of an alternative} 
in decision making. Accepting these analogies, we can go further. Keeping in mind that 
{\it emotions} appear subconsciously during the process of decision making, they can 
be associated with {\it intrinsic noise} produced by a measuring device during the 
measurement procedure. In that way, the {\it observable-noise duality} is equivalent 
to the {\it cognition-emotion duality}. In the same way as in physical measurements 
the detection of signals can be either hindered by noise or the addition of the 
appropriate amount of noise can boost a signal and hence facilitate its detection
\cite{Buchleitner_52,Gardiner_53}, in decision processes, emotions can either hinder 
or facilitate decision making. 

In quantum measurements, there can exist {\it observable-noise entanglement}, 
which in decision making corresponds to correlations mimicking {\it cognition-emotion 
entanglement}. If the intrinsic noise is presented as a superposition of several 
modes, then there appears {\it noise interference}, hence there can arise the 
{\it emotion interference}. In that way, it is possible to expect different similarities 
between quantum measurements and decision making. So, even if consciousness does not 
function exactly by the same rules as quantum measurements, but anyway the many 
found similarities can provide useful hints for formalizing the operation of decision 
procedures, hence for the creation of artificial intelligence.     

Concluding, in order to avoid confusion, it is necessary to stress what the review 
is about and what are not the review aims. This is in no way a survey of the general 
field of quantum techniques applied to the characterization of consciousness, because 
of which thousands of articles on such applications are not discussed, but only the 
main books are cited, where the vast number of references can be found. Concentrating 
on the ideas and methods for emotion quantification, citations are given only to those 
works where the role of emotions in decision making is studied and especially where 
practical methods of their description are discussed, but we do not plunge into the 
ocean of papers where no these problems are touched upon. While in the majority of 
works discussing the applications of quantum theory to consciousness neither the role 
of emotions is considered, nor their quantification is touched at all.   
 
The very first requirement in the way of creating human-like artificial intelligence
is the formulation of explicit mathematical rules of its operation. This paper does 
not pretend to describe all technical stages of actual artificial intelligence 
functioning, but it aims at formulating explicit mathematical algorithms for the 
operation of a human-like artificial intelligence in the process of taking decisions.
Without a mathematical description of such rules and algorithms, no device can be 
modeled. But in order to mathematically formulate the process of choice in human-like 
decision making to be implemented by an artificial intelligence, it is compulsory to 
understand and mathematically describe the process of choice by humans, whose actions
an artificial intelligence is planned to mimic. Therefore the pivotal aim of the paper 
is to analyze the combination of the following problems, whose solution is necessary 
for the mathematical formulation of decision making by an intelligence, whether 
human-like artificial or human:

(1) The analysis of the role of emotions in decision making and survey of the 
related literature, whether it employs quantum or classical language. This is 
necessary for understanding the basic qualitative principles of Affective 
Intelligence Processing

(2) The exposition of a practical way for emotion quantification in the process 
of taking decisions. This is a prerequisite for the formation of an Affective 
Artificial Intelligence requiring, for its functioning, the existence of explicit 
quantitative algorithms. 

(3) The comparison of two ways, quantum and classical, for the formulation of the 
practical principles of Affective Decision Making. This is compulsory for selecting 
the most appropriate method that would be self-consistent, simple, and providing 
quantitative recipes for its operation.  

(4) The comprehension of how the classical approach has to be modified in order to
provide the same practical results as with the use of quantum techniques. Again, 
this is impossible to understand without a comparison of both approaches, quantum 
and classical. Otherwise, the reader would constantly exclaim: Why this or that 
assumption has been made? Where this or that formula has appeared from?      

These goals are realized in the review. An exhaustive survey of literature discussing 
the role of emotions in decision making is given. The attempts of emotion 
quantification are described being based on the available literature. As is evident 
from numerous given citations, there are plenty of papers discussing the role of 
emotions in classical terms. The detailed comparison of quantum and classical 
techniques is given. It is shown that the classical approach can be modified by 
taking account of emotions, in such a way that to give the same results as in the 
language of quantum decision theory. For example, all paradoxes of classical decision 
making can be quantitatively explained without any use of quantum theory. 

However, without comparing the two different approaches for taking into account 
emotions, it would be impossible: First, to make a conclusion which of them is 
preferable, and second, how it would be necessary to modify the classical theory 
so that it would give the same results as the quantum approach. Therefore all 
parts of the review are of equal importance and would lose sense being separated.
Thus it is impossible to justify one of the approaches without comparing it with 
the other. On the other side, after different approaches are formulated, they can 
be employed independently and their effectiveness compared. 
      
The layout of the review is as follows. In Sec. $2$, the general theory of quantum
measurements in the presence of intrinsic noise is introduced. The analogies with 
decision making are emphasized. Assuming that the functioning of noisy quantum 
measurements is similar to that of affective decision making suggests the general 
framework for the latter. The comparison of a quantum and a modified classical 
approaches does not merely provide interesting analogies, but it allows for the 
formulation of the most simple and effective theory of Affective Decision Making. 

Quantum techniques, of course, are not a common knowledge and can strongly hinder the 
use of quantum theory for practical applications. Therefore, if the same phenomena
can be described in quantum language and also in classical terms, it is reasonable 
to resort to the simpler classical approach, but not to play science specially 
complicating the consideration with fashionable terminology. The theory has to be 
as simple as possible, in order that it could be straightforwardly employed by anyone, 
including those who may not know quantum techniques. This concerns decision theory 
as well, which can be developed as a branch of quantum theory or can be reformulated 
into an axiomatic form that, from one side mimics some quantum operations and structures,
but, from the other side, does not require the knowledge of quantum terminology. 
Section 3 accomplishes this goal showing that the theory of affective decision making 
can be formulated in an axiomatic way that does not need to resort to quantum theory. 
Being formulated in mathematical terms, affective decision theory can be implemented 
for the operation of artificial intelligence. Section 4 considers the famous behavioural 
paradoxes in decision making and shows that, on the aggregate level, these paradoxes 
do not arise in the frame of the affective decision theory. In that sense, an artificial 
intelligence, obeying the rules of this theory, will act as a typical human decision 
maker. In Section 5, the structure of networks composed of intelligent agents taking 
decisions in the presence of emotions is described. Section 6 concludes.

\section{Measurements under intrinsic noise}

One of the main points advocated in the present review is the existence of an analogy
between human emotions in decision making and intrinsic noise in quantum measurements.
This obliges us to investigate the structure of quantum probability for measurements
under intrinsic noise in order to find out the answers to two principal questions:

\vskip 2mm
1. Can this analogy be employed for developing affective decision theory that could 
be sufficiently formalized for being useful for describing the operation of human-level
artificial intelligence?

\vskip 2mm
2. Whether this analogy is merely nominal or it goes deeper than that, requiring the
use of quantum techniques for adequately representing behavioural decision making?

\vskip 2mm      
In physics, noise is modeled by introducing concrete noise terms into Hamiltonians 
or evolution equations and prescribing the corresponding distributions 
\cite{Buchleitner_52,Gardiner_53,Horsthemke_54,Gardiner_55}. Our aim here is not the
analysis of concrete models, but the study of the general structure of probabilities 
for quantum events, decorated by the presence of intrinsic noise \cite{Yukalov_56}. 
This is because we wish to compare these probabilities with those arising in decision 
theory. However the explicit nature of intrinsic noise, mimicking emotions, appearing 
in the process of taking decisions, is not known. Thus only general structures can be 
compared. In parallel with the physics terminology, we shall mention decision-making 
analogies of the considered notions. The mentioned analogies do not imply that the 
process of taking decisions by humans has to be necessarily treated as a quantum 
procedure, but conversely, this rather means that quantum measurements can be handled 
as formally similar to decision making 
\cite{Benioff_50,Benioff_51,Yukalov_57,Yukalov_58,Yukalov_59}. The most important 
analogy is between intrinsic noise in quantum measurements and emotions in decision 
making \cite{Yukalov_59,Yukalov_Emotions}.

\subsection{Quantum algebra of events}

Let us consider quantum events $A_n$, enumerated with the index $n=1,2,\ldots$. For 
concreteness, we consider a discrete index $n$, while in general it could be continuous.
Events can be the results of measurements for observable quantities. In decision theory, 
an event can be the choice of a particular alternative from the given set of alternatives. 
The collection of quantum events forms a ring \cite{Birkhoff_60,Hughes_61}, 
\be
\label{2.1} 
\mathbb{A} = \{ A_n : ~ n = 1,2, \ldots \}
\ee
possessing two binary operations, addition and conjunction. Addition, or union, 
or disjunction, implies that for any $A, B \in \mathbb{A}$ there is the union 
$A \cup B \in \mathbb{A}$, meaning either $A$ or $B$ and enjoying the properties
$$
A \bigcup B = B \bigcup A   \qquad  ( {\rm commutativity} ) \; ,
$$
$$
A \; \bigcup \; \left( B \; \bigcup \; C \right)  =  
\left( A \; \bigcup \; B \right) \; \bigcup \; C  
\qquad  
( {\rm associativity} ) \; ,
$$
$$
A \; \bigcup \; A =  A   \qquad  ( {\rm idempotency} ) \;  .
$$  
Conjunction, or multiplication, means that for any $A, B \in \mathbb{A}$ there exists
$A \cap B \in \mathbb{A}$ implying both $A$ and $B$ and having the properties
$$
A \; \bigcap \; \left( B \; \bigcap \; C \right)  =  
\left( A \; \bigcap \; B \right) \; \bigcap C \;
 \qquad  
( {\rm associativity} ) \; ,
$$
$$
A \; \bigcap \; A =  A   \qquad  ( {\rm idempotency} ) \;   .
$$
In general, conjunction is not commutative,
$$
A \; \bigcap \; B \neq B \; \bigcap \; A \qquad  ( {\rm no \; commutativity} ) \;  , 
$$
and not distributive,
$$
A \; \bigcap \; \left( B \; \bigcup \; C \right)  \neq  
\left( A \; \bigcap \; B \right) \; \bigcup \; \left( A \; \bigcap \; C \right) 
\qquad  ( {\rm no \; distributivity} ) \;   .
$$

The ring $\mathbb{A}$ includes the identical event $1 \in \mathbb{A}$, which is an 
event that is identically true. For this event,
$$
A \; \bigcap \; 1 = 1 \; \bigcap \; A = A \; , \qquad A \; \bigcup \; 1 = 1 \; , 
\qquad 
1 \; \bigcup \; 1 = 1 \; .
$$
Also, there exists an impossible event $0 \in \mathbb{A}$, which is identically 
false, so that
$$
A \; \bigcap \; 0 = 0 \; \bigcap \; A = 0 \; , \qquad  A \; \bigcup \; 0 = A \; , 
\qquad 0 \; \bigcup \; 1 = 1 \;   .
$$
The events for which $A \cap B = B \cap A = 0$ are called disjoint or orthogonal.
Note that one often simplifies the above notation by denoting the addition as 
$A \cup B \equiv A + B$ and the conjunction as $A \cap B \equiv AB$. 

For each event $A \in \mathbb{A}$, there exists a complementary, or negating, event 
$\overline{A} \in \mathbb{A}$, for which 
$$
A \; \bigcup \; \overline A = 1 \; , \qquad 
A \; \bigcap \; \overline A = \overline A \; \bigcap \; A = 0 \; , \qquad 
\overline 0 = 1 \; , \qquad \overline 1 = 0 \;  .
$$

The absence of distributivity can be demonstrated by the simple example 
\cite{Birkhoff_60}. Consider two events $B_1$ and $B_2$ whose union forms unity, 
$B_1\cup B_2=1$. And assume that both $B_1$ and $B_2$ are orthogonal to a non-trivial 
event $A \neq 0$, so that $A\cap B_1=A\cap B_2 = 0$. By this definition, 
$A\cap (B_1\cup B_2) = A \cap 1 = A$. If the property of distributivity were true, 
then it would be $(A \cap B_1) \cup (A \cap B_2)=0$. Since, by assumption, $A\neq 0$, 
the property of distributivity does not hold. 

The concept of non-distributivity in quantum physics can be illustrated by the example 
of spin measurements \cite{Hughes_61}. Let the spin projection of a particle with spin 
1/2 be measured. Suppose $B_1$ is the event of measuring the spin in the up state with 
respect to the $z-$axis, whereas $B_2$ is the event of measuring the spin in the down 
state along this axis. The spin can be either up or down, hence $B_1\cup B_2=1$. Assume 
that $A$ is the event of measuring the spin along an axis in the plane orthogonal to the 
$z-$axis, say along the $x-$axis. Since the spin cannot be measured simultaneously along 
two orthogonal axes, it is found either measured along one axis or along another axis, 
but cannot have components on both axes simultaneously. Hence, $A\cap B_1=A\cap B_2=0$. 
At the same time, $A\cap (B_1\cup B_2)= A\neq 0$. Therefore, there is no distributivity 
of events in the spin measurement.

\subsection{Operationally testable events}

An event is termed operationally testable, when it can be quantified by means of 
measurements. In physics, one measures observable quantities. For example, in quantum 
physics, one can measure the eigenvalues of a Hermitian operator corresponding to an 
observable. In decision making, one makes decisions by choosing preferable alternatives
from the given set. Quantum measurements can be treated as a kind of decision making
\cite{Neumann_49,Benioff_50,Benioff_51,Yukalov_57,Holevo_62,Holevo_63}.  

Let us consider a set of alternatives (\ref{2.1}) representing, e.g., a set of 
eigenvalues of an operator in quantum physics or a set of alternatives in decision 
theory. In quantum theory, each $A_n$ can be put into correspondence to a vector
(state of an alternative) $|A_n\rangle$ in a Hilbert space $\mathcal{H}_A$. For 
simplicity, we keep in mind nondegenerate spectra of Hermitian operators. The vectors 
$|A_n\rangle$ can be orthonormalized,
\be
\label{2.2}
 \lgl \; A_m \; | \; A_n \; \rgl = \dlt_{mn} \; .
\ee
Here and in what follows, the Dirac bracket notation \cite{Dirac_64} is employed. The 
Hilbert space $\mathcal{H}_A$ can be defined as the closed linear envelope
\be
\label{2.3}  
 \cH_A = {\rm span}\; \{ | \; A_n \; \rgl \} \;  .
\ee
In quantum decision theory, this is called the space of alternatives.  

Each alternative $A_n$ is represented by a projection operator
\be
\label{2.4}
  \hat P(A_n) =  | \; A_n \; \rgl \lgl \; A_n \; |
\ee
enjoying the property
$$
\hat P(A_m) \; \hat P(A_n) = \dlt_{mn} \; \hat P(A_n) \; .
$$
The latter means that the projection operators are idempotent and the alternatives 
of the ring $\mathbb{A}$ are mutually incompatible. The projection operators satisfy 
the resolution of unity
\be
\label{2.5}
\sum_n  \hat P(A_n) = \hat 1_A \; ,
\ee
where $\hat{1}_A$ is the identity operator in $\cH_A$. The complete family of the 
projection operators forms the projection-valued operator measure on $\cH_A$ with 
respect to set (\ref{2.1}).  

In quantum physics there can happen degenerate spectra, when for an $A_n$ there 
correspond several vectors $|A_{n_i}\rangle$, with $i = 1,2, \ldots$. Then the 
projection operator associated with $A_n$ is
$$
 \hat P(A_n) = \sum_i \hat P(A_{n_i})  \; , \qquad
\hat P(A_{n_i}) \equiv  | \; A_{n_i} \; \rgl \lgl \; A_{n_i} \; | \;   .
$$
If one wishes to avoid the problem of degeneracy, one slightly modifies the considered 
system by introducing infinitesimal terms lifting the degeneracy connected with some 
kind of symmetry, as has been mentioned by von Neumann \cite{Neumann_49} and elaborated 
by Bogolubov \cite{Bogolubov_65,Bogolubov_66,Bogolubov_67}. Similarly, in decision theory, 
the problem of degeneracy can be easily avoided by reclassifying the alternatives under 
consideration \cite{Yukalov_68,Yukalov_69}. Thus for decision theory, it is sufficient 
to consider the situation with no degeneracy.

\subsection{Modes of intrinsic noise}

Any measurement is accompanied by some kind of noise that can be of two types, 
extrinsic and intrinsic 
\cite{Buchleitner_52,Gardiner_53,Horsthemke_54,Gardiner_55,Buchleitner_70}.
Here we are interested in intrinsic noise that is generated by the measurement 
device in the process of measurement. Because of the intrinsic noise, what is 
measured is not a pure result for an alternative, but a combination of the data 
related to the testable event of interest and the influence of noise. The intrinsic 
noise also can be called instrumental or self-induced. In decision theory, the 
analogy of the intrinsic noise is the collection of emotions arising in the process 
of decision making, of subconscious allusions, intuitive guesses, gut feelings, and 
like that \cite{Yukalov_56,Yukalov_71}.     
 
Suppose the intrinsic noise is characterized by a set of elementary modes 
\be
\label{2.6}
 \mathbb{E} = \{ e_\mu : ~ \mu=1,2,\ldots \} \;  .
\ee
In decision theory, this would be a family of different emotions. Each noise mode 
$e_\mu$ is put into correspondence with a vector (noise state) $|e_\mu \rangle$ 
of a Hilbert space of noise $\mathcal{H}_E$ that can be represented as the closed 
linear envelope 
\be
\label{2.7} 
 \cH_E = {\rm span}\; \{ |\; e_\mu \; \rgl \} \;  .
\ee
The vectors of elementary nodes are assumed to be orthonormalized,
\be
\label{2.8}
 \lgl \; e_\mu \; | \; e_\nu \; \rgl = \dlt_{\mu\nu} \;  .
\ee
In quantum decision theory, space (\ref{2.7}) is called the emotion space. Emotion 
modes represent different types of emotions, such as joy, sadness, anger, fear, disgust,
trust, etc.

The projection operator for a mode $e_\mu$ is 
\be
\label{2.9}
\hat P(e_\mu) =  |\; e_\mu \; \rgl \lgl \; e_\mu \;|  \;  ,
\ee
which is idempotent and orthogonal to the projectors of other modes,
\be
\label{2.10}
\hat P(e_\mu) \hat P(e_\nu) =  \dlt_{\mu\nu} \; \hat P(e_\mu) \; .
\ee
The family of these projectors is complete satisfying the resolution of unity
\be
\label{2.11}
\sum_\mu \hat P(e_\mu) = \hat 1_E \;   ,
\ee
where $\hat{1}_E$ is the unity operator in $\cH_E$. Thus the family of projectors 
(\ref{2.9}) forms the projection-valued operator measure with respect to the 
set (\ref{2.6}).

The measurement of an alternative $A_n$ generates the intrinsic noise $z_n$ 
represented by the vector
\be
\label{2.12}
| \; z_n \; \rgl = \sum_\mu a_{n\mu} \; | \; e_\mu \; \rgl \;  .
\ee
In decision making, this corresponds to the collection of emotions arising under 
the choice between alternatives. The noise vector (\ref{2.12}) can be normalized,
\be
\label{2.13}
 \lgl \; z_n \;  | \; z_n \; \rgl = \sum_\mu |\; a_{n\mu} \; |^2 = 1  \; ,
\ee
although the noise vectors generated by different measurements are not necessarily 
mutually orthogonal, so that the product
\be
\label{2.14}
 \lgl \; z_m \;  | \; z_n \; \rgl = \sum_\mu a^*_{m\mu} a_{n\mu}
\ee
is not obligatory a Kronecker delta. Equivalently, the collections of emotions 
generated under the choice of different alternatives do not need to be mutually 
exclusive.

Strictly speaking, emotions are contextual and are subject to temporal variations, 
which means that the coefficients $a_{n\mu}$, generally, can vary with time, depending 
on the state of a decision maker and the corresponding surrounding.

The noise projectors 
\be
\label{2.15}
\hat P(z_n) =    | \; z_n \; \rgl \lgl \; z_n \; |    
\ee
are idempotent,
\be
\label{2.16}
 \hat P^2(z_n) = \hat P(z_n) \; ,
\ee
however, generally, are not mutually orthogonal,
\be
\label{2.17}
  \hat P(z_m) \hat P(z_n) = 
( \; \lgl \; z_m \;  | \; z_n \; \rgl \; ) \; | \; z_m \;  \rgl \lgl \; z_n \; |
\ee
because of property (\ref{2.14}). 

Note the important difference between projectors (\ref{2.9}) and (\ref{2.15}). 
The family of projectors (\ref{2.9}) is complete with respect to the set (\ref{2.6}) 
due to the resolution of unity (\ref{2.11}). However, the set of projectors 
(\ref{2.15}) is not complete with respect to the set 
\be
\label{2.18}
\mathbb{Z} = \{ z_n : ~ n = 1,2,\ldots \} \;   ,
\ee
since the sum
\be
\label{2.19}
\sum_n \hat P(z_n) = \sum_n \; \sum_{\mu\nu} a_{n\mu} a^*_{n\nu} 
| \; e_\mu \;  \rgl \lgl \; e_\nu \; |
\ee
is not a unity operator in $\mathcal{H}_E$. The latter is clear from the equality
$$
 \lgl \; e_\mu \; | \; \sum_n \hat P(z_n) \; | \; e_\nu \; \rgl =
\sum_n a^*_{n\mu} a_{n\nu} \;  ,
$$
which is not a Kronecker delta.

\subsection{Noise-decorated alternatives}

When a measurement of an alternative $A_n$ is accompanied by inevitable intrinsic 
noise $z_n$, what is actually observed is not a pure event $A_n$ but this event 
decorated with the noise, that is the combined event $A_n z_n$, whose representation 
is given by the vector
\be
\label{2.20} 
| \; A_n z_n \; \rgl = | \; A_n  \; \rgl \; \bigotimes \; | \; z_n \; \rgl =
\sum_\mu a_{n\mu} \; | \; A_n e_\mu \; \rgl
\ee
in the Hilbert space
\be
\label{2.21}
 \cH = \cH_A \; \bigotimes \; \cH_E = 
{\rm span} \; \{\; | \; A_n e_\mu \; \rgl \; \} .
\ee
The vectors defined in equation (\ref{2.20}) are mutually orthogonal,
\be
\label{2.22}
 \lgl \; z_m A_m \; | \; A_n z_n \; \rgl = \dlt_{mn} \;  .
\ee

The set of the noise-decorated events
\be
\label{2.23}
\mathbb{A}_Z = \{ A_n z_n : ~ n = 1,2,\ldots \}
\ee
is characterized by the family of the projectors
\be
\label{2.24}
 \hat P(A_n z_n) =    | \; A_n z_n \; \rgl \lgl \; z_n A_n \; | =
\hat P(A_n) \; \bigotimes \; \hat P(z_n)
\ee
that also can be written as
\be
\label{2.25}
 \hat P(A_n z_n) = \sum_{\mu\nu} a_{n\mu} a^*_{n\nu} \;
 | \; A_n e_\mu \; \rgl \lgl \; e_\nu A_n \; | \;  .
\ee
These projectors are idempotent and mutually orthogonal,
\be
\label{2.26}
 \hat P(A_m z_m) \hat P(A_n z_n) = \dlt_{mn} \; \hat P(A_n z_n) .
\ee
However, since the vectors (\ref{2.20}) do not form a basis in space (\ref{2.21}), 
the projectors(\ref{2.24}) do not sum to one,
\be
\label{2.27}
 \sum_n  \hat P(A_n z_n) = \sum_n \; \sum_{\mu\nu} a_{n\mu} a^*_{n\nu} \;
| \; A_n e_\mu \; \rgl \lgl \; e_\nu A_n \; | \;  ,
\ee
which is seen from the equality
$$
 \lgl \; e_\mu A_m \; | \; \sum_k \hat P(A_k z_k) \; | \;
A_n e_\nu \; \rgl = \dlt_{mn} \; a^*_{n\mu} a_{n\nu} \;  .
$$
Thus the family of projectors (\ref{2.24}) is idempotent, orthogonal, but not 
complete, hence does not compose a standard operator-valued measure and requires 
some additional conditions for introducing the probability of alternatives 
\cite{Yukalov_71,Yukalov_72}.

\subsection{Quantum probability space}

Statistical properties of the considered system are characterized by a statistical 
operator $\hat{\rho}$ that depends on the context and the observer's knowledge about 
the state of the system, because of which it is also called the state-of-knowledge 
operator. The operator $\hat{\rho}$ is a positive-semidefinite trace-one operator. 
It is also called the system state, or often simply a state. 

The general representation of $\hat{\rho}$ in the basis $\{|A_n e_\mu \rgl \}$ of 
orthonormalized vectors, that are not necessarily eigenvectors of $\hat{\rho}$, has 
the form
\be
\label{2.28}
 \hat\rho = \sum_{mn} \; \sum_{\mu\nu} \rho_{mn}^{\mu\nu} \;
|\; A_m e_\mu \; \rgl \lgl \; e_\nu A_n \; | \;  ,
\ee
with
\be
\label{2.29}
 \rho_{mn}^{\mu\nu} \equiv 
\lgl \; e_\mu A_m \; | \; \hat\rho \; | \; A_n e_\nu \; \rgl \; .
\ee
The trace normalization condition can be written as
\be
\label{2.30}
{\rm Tr}_\cH \hat\rho = \sum_{n\mu} \rho_{nn}^{\mu\mu} = 1 \;  .
\ee
A positive-semidefinite operator on a complex Hilbert space is necessarily 
self-adjoint \cite{Conway_73}, which imposes the constraint
\be
\label{2.31}
\left( \rho_{mn}^{\mu\nu} \right)^* = \rho_{nm}^{\nu\mu} \qquad 
( \hat\rho^+ = \hat\rho )  \; .
\ee
Also, let us require that the family of projectors (\ref{2.24}) be complete on 
average, so that
\be
\label{2.32}
 {\rm Tr}_\cH \hat\rho \; \left[\; \sum_n \hat P(A_nz_n) \; \right] = 1 \;  ,
\ee
which acquires the form
\be
\label{2.33}
 \sum_n \; \sum_{\mu\nu} a_{n\mu}^* a_{n\nu} \rho^{\mu\nu}_{nn} = 1 \;  ,
\ee
with the trace over the space (\ref{2.21}).

To be self-consistent, the system of constraints should not be overdefined. This 
means that the number of the involved parameters cannot be smaller than number of 
the constraint equations. The vectors $|z_n\rangle$ include the complex coefficients 
$a_{n \mu}$ containing $2 d_A d_E$ real components, where
\be
\label{2.34}
d_A \equiv {\rm dim} \; \cH_A \; , \qquad 
d_E \equiv {\rm dim} \; \cH_E \; .
\ee
The statistical operator $\hat{\rho}$ comprises the coefficients $\rho_{mn}^{\mu\nu}$ 
containing $d_Ad_E$ real diagonal elements and $d_A^2d_E^2-d_Ad_E$ complex off-diagonal 
elements, hence in total $2d_A^2d_E^2-d_Ad_E$ real components. Thus the total number of 
real components in $\hat{\rho}$ is $2d_A^2 d_E^2 - d_A d_E$. Conditions (\ref{2.31}) 
impose $d_A^2 d_E^2 - d_A d_E$ restrictions. In addition, there are two normalization 
conditions (\ref{2.30}) and (\ref{2.33}). In this way, there are in total 
$2d_A^2 d_E^2 + d_A d_E$ real parameters and $d_A^2 d_E^2 - d_A d_E + 2$ equations. 
From here, the condition of self-consistency becomes
\be 
\label{2.35}
d_A^2 d_E^2 +2 d_A d_E \geq 2 \;  ,
\ee
which holds for any $d_A d_E \geq 1$. 
   
The pair $\{ \cH, \hat\rho \}$ is called {\it quantum statistical ensemble} 
\cite{Yukalov_74}. Adding here the family 
\be
\label{2.36}
\cP_{AZ} = \{ \hat P(A_n z_n) : ~ n =1,2,\ldots \}
\ee
of projectors (\ref{2.24}) gives the {\it quantum probability space} 
$$
\{ \cH, \; \hat\rho, \; \cP_{AZ} \} \; .
$$ 

The probability of observing an event $A_n z_n$ reads as
\be
\label{2.37}
 p(A_n z_n)  = {\rm Tr}_\cH \; \hat\rho \; \hat P(A_n z_n) \;  .
\ee
The normalization condition (\ref{2.32}) guarantees the validity of the 
normalization condition
\be
\label{2.38}
 \sum_n p(A_n z_n) = 1 \; , \qquad 0 \leq p(A_n z_n) \leq 1 \;  .
\ee
Explicitly, equation (\ref{2.37}) takes the form
\be
\label{2.39}
 p(A_n z_n) = \sum_{\mu\nu} a^*_{n\mu} a_{n\nu} \rho^{\mu\nu}_{nn} \; .
\ee
The form (\ref{2.37}) defines the quantum probability of observing an alternative 
$A_n$ decorated by intrinsic noise.

\subsection{Quantum-classical correspondence}

Quantum theory reduces to classical under the effect of decoherence 
\cite{Zurek_75,Schlosshauer_76}. Then quantum decision theory reduces to classical 
and the quantum probability (\ref{2.37}) reduces to classical probability 
\cite{Yukalov_68,Yukalov_77} .

In expression (\ref{2.39}), it is possible to separate the diagonal part
\be
\label{2.40}
f(A_n z_n) = \sum_\mu |\; a_{n\mu}\; |^2 \; \rho^{\mu\mu}_{nn}
\ee
and the remaining off-diagonal part
\be
\label{2.41}
 q(A_n z_n) = \sum_{\mu\neq\nu} a^*_{n\mu} a_{n\nu} \; \rho^{\mu\nu}_{nn} \;  .
\ee
Then probability (\ref{2.37}) becomes the sum
\be
\label{2.42}
 p(A_n z_n) = f(A_n z_n) + q(A_n z_n) \; .
\ee
The first term here is semi-positive (non-negative), while the second one is not 
sign-defined. The term $q(A_n z_n)$ is due to the interference of noise modes and 
is zero if there is just a single mode or when $\rho_{nn}^{\mu \nu}$ is diagonal 
in the upper indices. This is why it can be called quantum interference term or 
quantum coherence term. In the present case, it is caused by the noise interference. 

The disappearance of the quantum coherence term is named decoherence, when there 
occurs the reduction of quantum probability to the classical form 
\cite{Zurek_75,Schlosshauer_76,Yukalov_77} associated with expression (\ref{2.40}). 
Interpreting the latter as classical probability implies the validity of the 
conditions
\be
\label{2.43}
 \sum_n f(A_n z_n) = 1 \; , \qquad 0 \leq f(A_n z_n) \leq 1 \; .
\ee
Therefore, in view of conditions (\ref{2.38}) and (\ref{2.43}), the interference 
term satisfies the conditions
\be
\label{2.44}
 \sum_n q(A_n z_n) = 0 \; , \qquad -1 \leq q(A_n z_n) \leq 1 \;  .
\ee
More precisely, it fulfills the inequality
\be
\label{2.45}
  - f(A_n z_n) \leq q(A_n z_n) \leq 1 - f(A_n z_n) \; .
\ee
In decision theory, the first equation in (\ref{2.44}) is called the alternation 
law \cite{Yukalov_78,Yukalov_79}.

The quantum-classical correspondence can be formulated as the reduction of quantum 
probability to classical under decoherence, when 
\be
\label{2.46}
 p(A_n z_n) \mapsto f(A_n z_n) \; , \qquad  q(A_n z_n) \mapsto 0 \; .
\ee
Thus the appearance of an additional term $q(A_n z_n)$ is due to the interference of 
noise modes. The phenomenon of mode interference is well known in quantum physics
\cite{Dittel_31,Vedral_81,Yukalov_82,Yukalov_83,Yukalov_84,Griffiths_80}. The absence 
of intrinsic noise accompanying measurements corresponds to the absence of emotions 
in the choice between alternatives.

\subsection{Probability of superposition states}

For the purpose of quantum information processing 
\cite{Williams_85,Nielsen_86,Vedral_87,Keyl_88,Mintert_89,Guhne_90,Wilde_91}, one creates
quantum states in the form of superpositions. Then it is admissible to define the 
probability of observing these states. For illustration, let us consider the binary
combinations of states
\be
\label{2.47}
 A_{mn} z_{mn} = A_m z_m \; \bigcup \; A_n z_n \qquad ( m \neq n) \;  .
\ee
Following the general procedure, each member (\ref{2.47}) can be put into correspondence 
to the vector
\be
\label{2.48}
 | \;  A_{mn} z_{mn} \; \rgl = c_m \; | \;  A_m z_m \; \rgl +
c_n \; | \;  A_n z_n \; \rgl
\ee
and characterized by the projector
\be
\label{2.49}
 \hat P( A_{mn} z_{mn} ) = 
| \;  A_{mn} z_{mn} \; \rgl \lgl \; z_{mn} A_{mn} \; | \; .
\ee
Vector (\ref{2.48}) is assumed to be normalized to one, which requires the 
condition
$$
| \; c_m \; |^2 + | \; c_n \; |^2  = 1 \; .
$$

It is worth emphasizing that this type of composite states could be introduced 
by combining the members from two different sets, say $\{A_n\}$ and $\{B_k\}$ such 
that their related vectors $|A_n\rgl$ and $|B_k\rgl$ pertain to the same basis in the 
Hilbert space, which is required by the necessity of defining the vector (\ref{2.48}) 
as a superposition with respect to a common basis. In the physics language, this means 
that, if $A_n$ and $B_k$ are the eigenvalues of some operators $\hat{A}$ and $\hat{B}$, 
then these operators have to commute with each other, since only then they enjoy the 
common family of orthonormalized eigenvectors. Noncommuting operators cannot form 
such linear combinations. In that sense, the corresponding events and observables are 
called incompatible as far as they cannot be measured simultaneously 
\cite{Neumann_49,Griffiths_80}.          

The projector (\ref{2.49}) reads as
$$
\hat P(A_{mn} z_{mn}) = | \; c_m \; |^2 \hat P(A_m z_m) + 
| \; c_n \; |^2 \hat P(A_n z_n) \; + 
$$
\be
\label{2.50}
+ \;
c_m c^*_n \; | \; A_m z_m \; \rgl \lgl \; z_n A_n \; | +
 c_n c^*_m \; | \; A_n z_n \; \rgl \lgl \; z_m A_m \; |  \; .
\ee
Then the probability of observing the composite state, corresponding to $A_{mn}z_{mn}$, 
becomes
$$
p(A_{mn} z_{mn}) = | \; c_m \; |^2 p(A_m z_m) + 
| \; c_n \; |^2 p(A_n z_n) \; +
$$
\be
\label{2.51}
+ \;
2{\rm Re}\; \left( c^*_m c_n \sum_{\mu\nu} 
a^*_{m\mu} a_{n\nu} \; \rho^{\mu\nu}_{mn} \right) 
\qquad ( m \neq n ) \;   .
\ee
This expression includes the terms due to the interference of noise modes as well 
as to the interference of the alternatives $A_m$ and $A_n$. Even when there is just 
a single noise mode, hence there is no noise interference, there remains the 
interference of alternatives. Thus in the case of a single noise mode $e_0$, when 
\be
\label{2.52}
 a_{m\mu} = \dlt_{\mu 0} \qquad ( z_n \mapsto e_0 ) \; ,
\ee
so that the noise interference disappears, probability (\ref{2.51}) reduces to the 
form
\be
\label{2.53}
p(A_{mn} z_{mn}) \mapsto 
| \; c_m \; |^2 p(A_m e_0) + | \; c_n \; |^2 p(A_n e_0) 
+ 2{\rm Re}\; \left( c^*_m c_n \; \rho_{mn} \right) 
\qquad ( m \neq n ) \;  ,
\ee
where
$$
\rho_{mn} \; \equiv \;
\lgl \; e_0 A_m \; | \; \hat\rho \; | \; A_n e_0 \; \rgl \; .
$$
The last term in probability (\ref{2.53}) describes the interference of alternatives. 
   
It is natural to ask whether the linear combinations of alternatives, hence the 
alternative interference, exist in human decision making, similarly to quantum physics. 
In the latter, the superpositions of wave functions representing quantum states do 
exist. However this quantum notion does not exist in human decision making. For 
instance, we can consider a set of fruits and vegetables deciding what to buy, an 
apple or a banana. However the seller will be quite astonished if we ask him/her 
to give us a quantum superposition of a banana and an apple. It looks that quantum 
superpositions of alternatives do not exist in the real life outside quantum 
experiments.     

Note that in many works of quantum cognition, one considers exactly the interference 
of alternatives, but not the interference of emotions (intrinsic noise). This is the 
principal difference between our approach and the works of other authors. In our 
approach, the operationally testable events, that is the observed alternatives, do 
not interfere. These are emotions that can interfere.

\subsection{Alternative-noise entanglement}

The notion of entanglement plays an important role in quantum information processing 
and quantum computing 
\cite{Williams_85,Nielsen_86,Vedral_87,Keyl_88,Mintert_89,Guhne_90,Wilde_91}.
Entanglement happens when the considered Hilbert space is represented as a tensor 
product of several Hilbert spaces. In our case, the Hilbert space (\ref{2.21}) is the 
product of the spaces characterizing the alternatives and intrinsic noise. Therefore 
there may exist entanglement between alternatives and noise. Generally, the statistical 
operator (\ref{2.28}) is entangled. The state that is not entangled is termed 
{\it separable}. In the present case, the state would be separable if it would have 
the form
\be
\label{2.54}
 \hat\rho_{sep} = \sum_i \lbd_i \; \hat\rho^i_A \bigotimes \hat\rho^i_E \;  ,
\ee
where the first factor in (\ref{2.54}) is a state acting on the space $\cH_A$, while 
the second is a state acting on the space $\mathcal{H}_E$, and 
$$
\sum_i \lbd_i = 1 \; , \qquad 0 \leq \lbd_i \leq 1 \;   .
$$
The factor states can be represented as
\be
\label{2.55}
\hat\rho^i_A = 
\sum_{mn} \rho^i_{mn} |\; A_m \; \rgl \lgl \; A_n \; | \; , 
\qquad
\hat\rho^i_E = 
\sum_{\mu\nu} \rho_i^{\mu\nu} |\; e_\mu \; \rgl \lgl \; e_\nu \; | \; ,
\ee
with the normalization
$$
\sum_n \rho_{nn}^i = 1 \; , \qquad \sum_\mu \rho^{\mu\mu}_i = 1 \;  .
$$
Then the state (\ref{2.54}) reads as
\be
\label{2.56}
 \hat\rho_{sep} = \sum_i \lbd_i \;  \sum_{mn} \; \sum_{\mu\nu} 
\rho^i_{mn}\; \rho_i^{\mu\nu} \;
 |\; A_m e_\mu \; \rgl \lgl \; e_\nu A_n \; | \; .
\ee
This means that in the representation (\ref{2.28}), the coefficient is
\be
\label{2.57}
 \rho_{mn}^{\mu\nu} = \sum_i \lbd_i \; \rho^i_{mn}\; \rho_i^{\mu\nu} \; .
\ee

For the separable state (\ref{2.54}), the classical limit (\ref{2.40}) becomes
\be
\label{2.58}
 f(A_n z_n) =   \sum_i \lbd_i \; 
\sum_\mu |\; a_{n\mu} \; |^2 \; \rho^i_{nn}\; \rho_i^{\mu\mu} 
\ee
and the quantum noise interference term is
\be
\label{2.59}
  q(A_n z_n) =   \sum_i \lbd_i \; \sum_{\mu\neq\nu}  
a^*_{n\mu}  \; a_{n\nu} \; \rho^i_{nn}\; \rho_i^{\mu\nu} \;  .
\ee

Thus, generally, the alternatives and noise are entangled with each other. The noise 
interference is not related to whether the state is entangled or not. The state can 
be separable, while the noise interference present. In quantum decision theory, the 
alternative-noise entanglement is equivalent to the entanglement of alternatives 
and emotions \cite{Yukalov_92,Yukalov_93}.

\subsection{Entanglement production by measurements}

It is necessary to keep in mind that, for operators, there exist two different 
notions, state entanglement and operator entanglement production. An entangled 
state is a state that is not separable, as is explained above. In that sense, 
entanglement is the property of the state structure. While entanglement production 
by an operator is the ability of generating entangled functions from disentangled 
ones.

A vector of a tensor-product Hilbert space is named disentangled if it can be 
represented as a tensor product of vectors pertaining to the factor Hilbert spaces. 
Just as an example, the basis vector $|A_n e_\mu\rgl=|A_n\rgl \otimes |e_\mu\rgl$ 
is disentangled. Disentangled vectors are often called separable vectors.    

An operator is called entangling, when there exists at least one disentangled vector 
such that it becomes entangled under the action of this operator. Conversely, one says 
that an operator is not entangling if its action on any disentangled vector yields 
again a disentangled one. It has been proved \cite{Marcus_94,Westwick_95,Johnston_96} 
that the only operators preserving vector separability are the operators having the
form of tensor products of local operators and a swap operator permuting Hilbert spaces 
in the tensor product describing the total Hilbert space of a composite system. The 
action of the swap operator is trivial, in the sense that it merely permutes the indices 
labeling the spaces. The vector separability preservation by product operators has been 
proved for binary \cite{Marcus_94,Beasley_97,Alfsen_98} as well as for multipartite 
vectors \cite{Westwick_95,Johnston_96,Friedland_99}. The operators preserving vector 
separability are called nonentangling \cite{Gohberg_100,Crouzeux_101}. While an operator 
transforming at least one disentangled vector into an entangled vector is termed 
entangling \cite{Fan_102,Dao_103}. The strongest type of an entangling operator is 
a universal entangler that makes all disentangled vectors entangled \cite{Chen_104}.  

Entanglement of vectors can be generated in the process of measurements by the action 
of statistical operators. The measure of entanglement production for arbitrary operators
has been introduced in Refs. \cite{Yukalov_105,Yukalov_106}. This measure is applicable 
to any system, whether bipartite or multipartite, and to any trace-class operator. It
has been applied for characterizing different physical systems 
\cite{Yukalov_107,Yukalov_108,Yukalov_109,Yukalov_110,Yukalov_111}, as reviewed in 
Ref. \cite{Yukalov_112}. Entanglement production generated in the process of decision 
making is studied in Ref. \cite{Yukalov_113}.

The measure of entanglement production by the statistical operator (\ref{2.28}) acting
on the Hilbert space (\ref{2.21}) is calculated as follows. We define the partially
traced operators
\be
\label{2.60}
\hat\rho_A \equiv {\rm Tr}_E \; \hat\rho = 
\sum_{mn} \rho_{mn} |\; A_m \; \rgl \lgl \; A_n \; | \;   ,
\qquad
\hat\rho_E \equiv {\rm Tr}_A \; \hat\rho = 
\sum_{\mu\nu} \rho^{\mu\nu} |\; e_\mu \; \rgl \lgl \; e_\nu \; | \;   ,
\ee
in which the traces are over $\mathcal{H}_E$ or $\mathcal{H}_A$, respectively, and 
$$
\rho_{mn} \equiv \sum_\mu \rho^{\mu\mu}_{mn} \; , \qquad
\rho^{\mu\nu} \equiv \sum_n \rho^{\mu\nu}_{nn} \;  .
$$
The nonentangling state is given by the product
\be
\label{2.61}
 \hat\rho^\otimes \equiv \hat\rho_A \; \bigotimes \; \hat\rho_E \;  .
\ee
Comparing the action of the statistical operator $\hat{\rho}$ with that of its 
nonentangling counterpart (\ref{2.61}) we have the measure of entanglement production  
by the statistical operator
\be
\label{2.62}
 \ep(\hat\rho) \equiv 
\log \; \frac{||\; \hat\rho\;||}{||\; \hat\rho^\otimes\;||} \;  .
\ee
Here 
$$
||\; \hat\rho^\otimes\;|| = ||\; \hat\rho_A\;|| \cdot ||\; \hat\rho_E\;||  . 
$$
Keeping in mind the spectral norm yields
\be
\label{2.63}
 ||\; \hat\rho\;|| = \sup_{n\mu} \rho^{\mu\mu}_{nn}
\ee
and 
\be
\label{2.64}
||\; \hat\rho_A \;|| = \sup_n \; \sum_\mu \rho^{\mu\mu}_{nn} \; ,
\qquad
||\; \hat\rho_E \;|| = \sup_\mu \; \sum_n \rho^{\mu\mu}_{nn} \;  .
\ee
Therefore the measure of entanglement production (\ref{2.62}) turns into
\be
\label{2.65}
\ep(\hat\rho) = \log \; 
\frac{\sup_{n\mu}\rho^{\mu\mu}_{nn}}
{(\sup_n\sum_\mu\rho^{\mu\mu}_{nn})(\sup_\mu\sum_n\rho^{\mu\mu}_{nn}) } \; .
\ee

In this way, the statistical operator produces the alternative-noise entanglement 
by acting on the vectors of the Hilbert space. As an illustration, it is easy to 
show that even a separable state can produce entanglement. Thus, acting by the 
separable state (\ref{2.54}) on the disentangled basis vector $|A_n e_\mu\rangle$ 
gives the vector
\be
\label{2.66}
 \hat\rho_{sep} \; | \; A_n e_\mu\; \rgl =
\sum_{m\nu} \; \rho^{\mu\nu}_{mn} \; | \; A_m e_\nu \; \rgl \;  ,
\ee
where $\rho_{mn}^{\mu\nu}$ is given by (\ref{2.57}). This vector is entangled if 
at least two $\lambda_i$ are not zero. Similarly, in the process of making decisions 
the alternatives and emotions become entangled.

\subsection{Time dependence of probability}

In the previous sections, the measurements as well as decision making have been 
treated as occurring during so short time that it could be neglected. However, these 
processes do require some finite time. In addition, one can accomplish measurements 
or decisions at different moments of time. Therefore, for the correct description of 
measurements, as well as decision making, it is necessary to take account of time 
dependence of quantum probabilities. 

Time enters the probability through the time dependence of statistical operator 
$\hat\rho(t)$, whose time evolution is given by means of a unitary evolution operator 
$\hat{U}(t)$ according to the rule
\be
\label{2.67}     
 \hat\rho(t) = \hat U(t,0) \; \hat\rho(0) \; \hat U^+(t,0) \;  .
\ee
Alternatives decorated by intrinsic noise are represented by the family (\ref{2.36}) 
of projectors (\ref{2.24}) acting on the Hilbert space $\mathcal{H}$ defined in 
(\ref{2.21}). The quantum probability space reads as
\be
\label{2.68}
\{ \cH , \; \hat\rho(t) , \; \cP_{AZ} \} \;  ,
\ee
with the probability becoming dependent on time,
\be
\label{2.69}
 p(A_n z_n,t) = {\rm Tr}\; \hat\rho(t) \; \hat P(A_n z_n) \; .
\ee
Here and in what follows, the trace operation, without a notation of the related 
Hilbert space, is assumed to be over the whole space (\ref{2.21}).

As early, the probability is represented as the sum
\be
\label{2.70}
  p(A_n z_n,t) =  f(A_n z_n,t) +  q(A_n z_n,t)   
\ee
of the classical limit
\be
\label{2.71}
f(A_n z_n,t) = \sum_\mu |\; a_{n\mu} \; |^2 \; \rho^{\mu\mu}_{nn}(t)
\ee
and the quantum term caused by the noise interference
\be
\label{2.72}
q(A_n z_n,t) = \sum_{\mu\neq\nu}  
a^*_{n\mu} a_{n\nu} \; \rho^{\mu\nu}_{nn}(t) \; .
\ee
The dependence on time comes from the time dependence of the matrix elements 
(\ref{2.29}). The noise vector $|z_n\rangle$ can also depend on time through 
the coefficients $a_{n\mu}$, which we do not stress explicitly for the sake 
of notation compactness. The dependence of noise on time, as well as the 
time-dependence of emotion properties is rather natural, since they can vary 
with time.  
 
Because of the unitarity of the evolution operator, the normalization condition 
(\ref{2.38}) remains true:
\be
\label{2.73}
 \sum_n p(A_n z_n,t) = 1 \; , \qquad 0 \leq p(A_n z_n,t) \leq 1 \; .
\ee
Similarly, the normalization conditions (\ref{2.43}), (\ref{2.44}), and 
(\ref{2.45}) also remain valid.

\subsection{Quantum state reduction}

Suppose, the system, at time $t=0$, is prepared in a state $\hat{\rho}(0)$ and 
develops in time following the evolution equation (\ref{2.67}). Then at time $t_0$ 
it is subject to a measurement procedure for an observable allowing for the set 
of alternatives (\ref{2.1}). In the same way, one could be talking about taking 
a decision at time $t_0$ by choosing an alternative from the set of possible 
alternatives. 

At time $t_0 - 0$, just before the measurement, the a priori probabilities of the 
alternatives are given by the equation
\be
\label{2.74}
 p(A_n z_n,t_0-0) = {\rm Tr}\;\hat\rho(t_0-0) \; \hat P(A_n z_n) \;  .
\ee
Let us assume that at the moment of time $t_0$ an alternative $A_n$ is certainly 
observed. In decision making, this would imply that an alternative $A_n$ is certainly 
chosen. In any case, this means that, as a result of the interaction between the 
studied system and a measuring device, the a priori state has been reduced to an 
a posteriori state,
\be
\label{2.75}
 \hat\rho(t_0-0) \mapsto \hat\rho(A_n,t_0+0) \; ,
\ee
so that the a posteriori probability
\be
\label{2.76}
 p(A_n z_n,t_0+0) = {\rm Tr}\;\hat\rho(A_n,t_0+0) \; \hat P(A_n z_n) 
\ee
becomes unity, thus describing a certain event,
\be
\label{2.77}
p(A_n z_n,t_0+0) = 1 \; .
\ee

The above condition in the explicit form reads as
\be
\label{2.78}
{\rm Tr}\;\hat\rho(A_n,t_0+0) \; \hat P(A_n z_n) = 1 \; .
\ee
It is easy to verify that the solution to this equation can be written in the form
\be
\label{2.79}
 \hat\rho(A_n,t_0+0) = \frac{\hat P(A_n z_n)\hat\rho(t_0-0)\hat P(A_n z_n)}
{{\rm Tr}\hat\rho(t_0-0)\hat P(A_n z_n)} \; .
\ee

The transformation (\ref{2.75}) is called quantum state reduction 
\cite{Neumann_49,Luders_114,Balian_115} and the form (\ref{2.79}) is named the von 
Neumann-L\"uders state. This state becomes the initial state for the following state 
dynamics at times $t>t_0$,
\be
\label{2.80}
\hat\rho(A_n,t) = 
\hat U(t,t_0) \; \hat\rho(A_n,t_0+0) \; \hat U^+(t,t_0) \; .
\ee
The a priori probability of measuring an alternative $A_m$ for $t>t_0$ is
\be
\label{2.81}
p(A_m z_m,t) = {\rm Tr}\; \hat\rho(A_n,t)\; \hat P(A_m z_m) \qquad
( t > t_0 ) \;.
\ee

Thus the state reduction (\ref{2.75}), caused by the measurement process, implies 
the change of the initial condition for the state, hence the change of the state 
evolution at later times, which in turn presumes the alteration of the quantum 
probability space,
\be
\label{2.82}
\{ \cH, \; \hat\rho(t) , \; \cP_{AZ} \} \mapsto 
\{ \cH, \; \hat\rho(A_n,t) , \; \cP_{AZ} \}
\ee
and, respectively, the reduction of the probability,
\be
\label{2.83}
 p(A_n z_n,t_0-0) \mapsto p(A_n z_n,t_0+0) \; .
\ee
It is important that the existence of intrinsic noise does not disturb the standard 
scheme of quantum state reduction.  

{\it The quantum state reduction is nothing but the change of an a priori probability 
to an a posteriori probability due to the received information}.

\subsection{Consecutive measurements of alternatives}

Assume that at time $t_0$ an alternative $A_n$ has been certainly observed, as is 
described in the previous section. Then state (\ref{2.79}) plays the role of an 
initial condition for the following state dynamics. For times $t>t_0$, we have the 
state (\ref{2.80}). 

Suppose that after the moment of time $t_0$, we are interested in measuring another 
observable corresponding to the new set of alternatives
\be
\label{2.84}
 \mathbb{B} = \{ B_k: ~ k=1,2,\ldots \} \; .
\ee
The related operators $\hat{A}$ and $\hat{B}$ are not necessarily commuting, since 
their measurements are accomplished at different times. These alternatives are again
assumed to be decorated by intrinsic noise. The projectors
\be
\label{2.85}
\hat P(B_kz_k) = | \; B_k z_k \; \rgl \lgl \; z_k B_k \; |
\ee
compose the family
\be
\label{2.86}
 \cP_{BZ} = \{ \hat P(B_kz_k): ~ k=1,2,\ldots \} \; .
\ee
After the time $t_0$, the quantum probability space is
\be
\label{2.87}
 \{ \cH , \; \hat\rho(A_n,t) , \; \cP_{BZ} \} \qquad ( t > t_0 ) \;  .
\ee
The probabilities of alternatives from the set (\ref{2.84}) read as
\be
\label{2.88}
 p(B_kz_k,t)  = {\rm Tr}\; \hat\rho(A_n,t) \; \hat P(B_kz_k) \qquad 
( t > t_0 )
\ee
and, as any probability, they are normalized:
\be
\label{2.89}
\sum_k p(B_kz_k,t)  = 1 \;  .
\ee
  
From the other side, probability (\ref{2.88}) can be interpreted as a conditional 
probability of measuring an alternative $B_k$ at time $t$, after the alternative 
$A_n$ at time $t_0$ has been certainly observed. Thus the conditional probability 
is defined as the straightforward renotation
\be
\label{2.90}
  p(B_kz_k,t)  \equiv  p(B_kz_k,t|A_nz_n,t_0)  \qquad  ( t > t_0 ) \;  , 
\ee
with the related renotation of normalization (\ref{2.89}),
\be
\label{2.91}
 \sum_k p(B_kz_k,t|A_nz_n,t_0) = 1 \; .
\ee

Substituting the state (\ref{2.80}) into the expression
\be
\label{2.92}
p(B_kz_k,t|A_nz_n,t_0) = {\rm Tr}\; \hat\rho(A_n,t) \; \hat P(B_kz_k)
\ee
and using the notation
\be
\label{2.93}
p(B_kz_k,t,A_nz_n,t_0) \equiv  {\rm Tr}\; \hat U(t,t_0) \; \hat P(A_nz_n) \;
\hat\rho(t_0-0) \; \hat P(A_nz_n) \; \hat U^+(t,t_0) \; \hat P(B_kz_k)
\ee
results in the probability that can be called conditional, 
\be
\label{2.94}
 p(B_kz_k,t|A_nz_n,t_0) = \frac{p(B_kz_k,t,A_nz_n,t_0)}{p(A_nz_n,t_0-0)} \; .
\ee
Employing normalization (\ref{2.91}), we get the relation
\be
\label{2.95}
\sum_k p(B_kz_k,t,A_nz_n,t_0) = p(A_nz_n,t_0-0) \;  .
\ee
From here, the normalization condition follows:
$$
\sum_{nk} p(B_kz_k,t,A_nz_n,t_0) = 1 \; .
$$

These formulas suggest that probability (\ref{2.93}) can be named as joint probability.
Similar equations often are considered for a fixed moment of time $t=t_0$, which 
brings problems dealing with incompatible events corresponding to the simultaneous 
measurement of noncommuting operators \cite{Goodman_116}. These problems do not arise 
when considering a realistic situation of measurements at different moments of time. 
Taking into account the presence of intrinsic noise also does not complicate much 
the consideration \cite{Yukalov_56,Yukalov_71}. 

Note that, for $t >t_0$, neither the joint probability (\ref{2.93}) nor the conditional 
probability (\ref{2.92}) or (\ref{2.94}) are symmetric with respect to the interchange 
of the events $A_n$ and $B_k$. This asymmetry can explain the so-called order effects
in decision theory, when the probability of choice depends on the order of choosing 
alternatives \cite{Yukalov_83}.

\subsection{Immediate consecutive measurements}

One often considers two measurements occurring immediately one after another 
\cite{Neumann_49,Luders_114}. This is the limiting case of the consecutive measurements 
treated in the previous subsection, when at the moment of time $t_0$ an alternative 
$A_n$ has been certainly observed and the second measurement of another observable 
corresponding to the set of alternatives (\ref{2.84}) is measured at the time $t_0+0$ 
immediately following $t_0$. 
  
In the case of these immediate measurements, the evolution operator reduces to unity 
operator,
\be
\label{2.96}
\hat U(t_0+0,t_0) = \hat 1 \; .
\ee
Then for the conditional probability (\ref{2.92}), we have
\be
\label{2.97}
 p(B_kz_k,t_0+0|A_nz_n,t_0) =  
{\rm Tr} \; \hat\rho(A_n,t_0+0) \; \hat P(B_kz_k)
\ee
and the joint probability (\ref{2.93}) becomes
\be
\label{2.98}
p(B_kz_k,t_0+0,A_nz_n,t_0) = {\rm Tr}\; 
\hat P(A_nz_n) \; \hat\rho(t_0-0) \; \hat P(A_nz_n) \; \hat P(B_kz_k) \; .
\ee
The conditional probability (\ref{2.94}) takes the form
\be
\label{2.99}
p(B_kz_k,t_0+0|A_nz_n,t_0) =  
\frac{p(B_kz_k,t_0+0,A_nz_n,t_0)}{p(A_nz_n,t_0-0)} \; ,
\ee
which can be called von Neumann-L\"{u}ders probability. The explicit expression for 
the joint probability (\ref{2.98}) turns into
\be
\label{2.100}
p(B_kz_k,t_0+0,A_nz_n,t_0) = 
|\; \lgl \; z_k B_k \; | \; A_n z_n \; \rgl \; |^2 \; p(A_nz_n,t_0-0) \; .
\ee
This transforms the conditional probability (\ref{2.99}) into the symmetric form
\be
\label{2.101}
 p(B_kz_k,t_0+0|A_nz_n,t_0) = 
|\; \lgl \; z_k B_k \; | \; A_n z_n \; \rgl \; |^2 \;  ,
\ee
where
$$
\lgl \; z_k B_k \; | \; A_n z_n \; \rgl = \sum_\mu a^*_{k\mu} a_{n\mu} \;
\lgl \; B_k \; | \; A_n \; \rgl \; .
$$

If the repeated measurement is accomplished with respect to the same observable, so 
that $B_k=A_k$, then the conditional probability (\ref{2.101}) reduces to
\be
\label{2.102}
  p(A_kz_k,t_0+0|A_nz_n,t_0) = \dlt_{nk} \; .
\ee
This is in agreement with the {\it principle of reproducibility} in quantum theory, 
according to which, when the choice, among the same set of alternatives, is made 
twice, immediately one after another, the second choice has to reproduce the first 
one \cite{Neumann_49}. This also is in agreement with decision making: when a decision 
maker accomplishes a choice from the same set of alternatives immediately after 
another choice, so that there is no time for deliberation, then this decision maker 
should repeat the previous choice \cite{Yukalov_56,Yukalov_71}.

Generally, the joint probability (\ref{2.98}) is not symmetric with respect to the 
interchange of the events $A_n$ and $B_k$. It becomes symmetric only when there is no 
noise and the corresponding operators commute with each other, hence enjoy the common 
basis of eigenvectors. At the same time, the conditional probability (\ref{2.101}) is 
always symmetric, whether for commuting or noncommuting observables, and whether in the 
presence or absence of noise. Therefore the immediate consecutive probabilities, with 
the symmetry properties
$$
 p(B_kz_k,t_0+0,A_nz_n,t_0) \neq p(A_nz_n,t_0+0,B_kz_k,t_0) \;  ,
$$
\be
\label{2.103}
 p(B_kz_k,t_0+0|A_nz_n,t_0) = p(A_nz_n,t_0+0|B_kz_k,t_0) 
\ee
cannot be accepted as a generalization of classical Kolmogorov-type probabilities, 
where the joint probability is symmetric, while the conditional one is not. This fact 
should not be of surprise as far as the definitions of quantum consecutive probability 
of two events occurring at different times and classical probability for two events 
occurring synchronously are principally different. Classical probability contains no 
mentioning of state evolution, while quantum probability connects two measurements 
realized at different times and involving the state evolution.

\subsection{Synchronous noiseless measurements}

Quantum theory, as it is usually formulated, is not directly analogous to classical 
probability theory in the sense of Kolmogorov \cite{Kolmogorov_1956}, but is much 
closer to the theory of stochastic processes \cite{Gardiner_55}. In nonrelativistic 
quantum mechanics, states at different times are related by dynamics, generally 
represented as a completely positive map. In that sense, consecutive measurements
correspond to dynamic probability with the underlying causal structure. This type of 
theory is closely analogous to a classical stochastic process, in which a state is
a probability distribution over a set of random variables representing the properties 
of a system at a given time and the states at different times are related by dynamics. 

In contrast, classical probability spaces make no assumptions about the causal structure 
of the events on which probabilities are defined. Two disjoint events might refer to 
properties of two different subsystems at a given time, or they might refer to properties 
of the same subsystem at two different times. In full generality, classical events need 
have no interpretation in terms of causal structure at all. 

A variant of quantum probability enjoying the same symmetry properties as classical
probability should be noiseless and allowing for the accomplishment of simultaneous 
measurements. This type of probability is defined as follows \cite{Yukalov_56}.

Let us consider two sets of alternatives
\be
\label{2.104}
\mathbb{A} =\{ A_n: ~ n= 1,2,\ldots \} \; , \qquad  
\mathbb{B} =\{ B_k: ~ k= 1,2,\ldots \} \;  .
\ee
A simultaneous measurement of two observables can be realized provided they pertain 
to two different Hilbert spaces, for instance they are located at two different 
spatial regions, or act in the spaces of different variables, e.g. momenta and spins. 
Speaking about simultaneous measurements at different spatial locations, we keep in 
mind a nonrelativistic situation when the notion of synchronously occurring events 
or measurements is well defined. In the relativistic case, we could use the notion 
of spacelike separated measurements or events. The corresponding Hilbert space is 
the tensor product
\be
\label{2.105}
\cH = \cH_A \; \bigotimes \; \cH_B \; .
\ee
For the moment, we do not include intrinsic noise.

Alternatives $A_n$ and $B_k$ are represented by the projectors $\hat P(A_n)$ and  
$\hat P(B_k)$ in the related spaces. The Hilbert space $\mathcal{H}$, the statistical 
operator $\hat{\rho}(t)$, and the family of the projectors
\be
\label{2.106}
\cP_{AB} = 
\{ \hat P(A_n) \otimes \hat P(B_k) : ~ n = 1,2,\ldots; ~  k = 1,2,\ldots\}
\ee
compose the quantum probability space
\be
\label{2.107}
 \{ \cH_A \otimes \cH_B , \; \hat\rho(t) , \; \cP_{AB} \} \; .
\ee
The probability of measuring the alternatives $A_n$ and $B_k$ in different spaces is
\be
\label{2.108}
p(A_n B_k,t) = 
{\rm Tr}\; \hat\rho(t) \; \hat P(A_n) \bigotimes \hat P(B_k) \;  .
\ee
      
The defined probability possesses the same properties as classical probability. Thus 
the marginal probabilities are given by the partial summation
\be
\label{2.109}
p(A_n,t) = \sum_k p(A_nB_ k,t) \; , \qquad  
p(B_k,t) = \sum_n p(A_n B_k,t) \;  ,
\ee
and they are normalized,
\be
\label{2.110}
 \sum_n p(A_n,t) = 1 \; , \qquad \sum_k p(B_k,t) = 1 \; .
\ee
If the measurements are not correlated, such that
$$
 \hat\rho(t) = \hat\rho_A(t) \; \bigotimes \; \hat\rho_B(t) \;  ,
$$
the joint probability becomes the product
$$
p(A_n B_k,t) =  p(A_n,t) p(B_k,t) \; .
$$

It is possible to introduce the conditional probability
\be
\label{2.111}
 p(B_k|A_n,t) \equiv  
\frac{p(B_k A_n,t)}{p(A_n,t)}  \; .
\ee

The defined joint and conditional quantum probabilities of synchronous events, happening 
in different Hilbert spaces, possess the same symmetry properties as the classical 
probability: the joint probability is symmetric with respect to the event interchange,
while the conditional probability is not symmetric,
\be
\label{2.112}
p(A_n B_k,t) =  p(B_k A_n,t) \;  , \qquad p(A_n|B_k,t) \neq  p(B_k|A_n,t) \; .  
\ee

Strictly speaking, the measurements of two events, simultaneously occurring at two 
different spatial locations, is possible only if the measuring device is sufficiently 
large, such that it includes several parts allowing for the synchronous measurement 
of two different events. In decision making, in order to accept that a subject is 
able to decide on two alternatives simultaneously, it is necessary to assume that 
either there are different parts of the brain thinking synchronously or what seems 
to be synchronous is actually a fast temporal reswitching from one object to another 
\cite{Pashler_117,Cowan_118,Goldstein_119}.

\subsection{Synchronous measurements under noise}

Synchronous measurements in different Hilbert spaces, e.g. at different spatial locations, 
can be straightforwardly generalized by including intrinsic noise. Then the system is 
defined in the Hilbert space
\be
\label{2.113}
 \cH = \cH_A \; \bigotimes \; \cH_E \; \bigotimes \; \cH_B \; \bigotimes \; \cH'_E \; ,
\ee
where $\cH'_E$ is a copy of $\cH_E$. The events are characterized by the family of the 
projectors
\be
\label{2.114}
 \cP_{ABZ} = \{ \hat P(A_n z_n) \otimes \hat P(B_k z_k) : ~
n= 1,2,\ldots; ~ k= 1,2,\ldots \} \; .
\ee

The quantum probability space becomes
\be
\label{2.115}
 \{ \cH, \; \hat\rho(t), \; \cP_{ABZ} \} \;  .
\ee
The probability of two synchronously occurring events in the presence of intrinsic 
noise is
\be
\label{2.116}
 p(A_n z_n B_k z_k,t) = {\rm Tr} \; \hat\rho(t) \; 
\hat P(A_n z_n) \bigotimes \hat P(B_k z_k)   
\ee
that is required to be normalized,
\be
\label{2.117}
 \sum_{nk}  p(A_n z_n B_k z_k,t) = 1 \; .
\ee
Notice that this probability is symmetric with respect to the order swap of the 
alternatives,
\be
\label{2.118}
p(A_n z_n B_k z_k,t) = p(B_k z_k A_n z_n,t) \; .
\ee
  
This probability, similarly to the probability of a single event, can be represented 
as a sum
\be
\label{2.119}
 p(A_n z_n B_k z_k,t) =  f(A_n z_n B_k z_k,t) +  q(A_n z_n B_k z_k,t) \; ,
\ee
of a diagonal part
\be
\label{2.120}
f(A_n z_n B_k z_k,t) =   \sum_{\mu \lambda} |\; a_{n \mu}\; |^2 \; |\; a_{k \lambda}\;|^2 \;  
\lgl \; e_\lambda e_\mu B_k A_n \; | \; \hat\rho(t) \; | \; A_n B_k e_\mu e_\lambda \; \rgl
\ee
and an off-diagonal part comprising all interference terms,
\be
\label{2.121}
 q(A_n z_n B_k z_k,t) =   \sum_{\mu\nu\gamma\lbd} a_{n\mu} a^*_{n\nu} 
a_{k\gamma} a^*_{k\lbd}  \lgl \; e_\lbd e_\nu B_k A_n \; | \; 
\hat\rho(t) \; | \; A_n B_k e_\mu e_\gamma \; \rgl \; ,
\ee
where
$$
 \sum_{\mu\nu\gamma\lbd} ~ \mapsto ~ 
\sum_{\mu\neq\nu} \; \sum_{\gamma\lbd} \dlt_{\gamma\lbd}  ~ + ~
\sum_{\mu\nu} \; \sum_{\gamma\neq\lbd} \dlt_{\mu\nu}  ~ + ~
\sum_{\mu\neq\nu} \; \sum_{\gamma\neq\lbd}  \; .
$$

Note that if the parts of the synchronous measurement are not correlated, so that
$$
 \hat\rho(t) = \hat\rho_{AZ}(t) \; \bigotimes \;  \hat\rho_{BZ}(t) \; ,
$$
then probability (\ref{2.116}) separates into two factors
$$
p(A_n z_n B_k z_k,t) = p(A_n z_n) p( B_k z_k) \;  .
$$

The conditional probability can be defined as 
\be
\label{2.122}
p(A_n z_n|B_k z_k, t) \equiv \frac{p(A_n z_n B_k z_k,t)}{p(B_k z_k,t)} \; .
\ee
This probability is not swap order symmetric,
$$
p(A_n z_n|B_k z_k, t) \neq p(B_k z_k|A_n z_n, t) \; .
$$
  
The synchronous joint probability (\ref{2.116}) is a natural generalization 
of classical probability to the case of quantum measurements under intrinsic noise. 
In decision theory, it plays the role of a behavioural probability of taking a
decision on two events simultaneously occurring in two different spatial locations.

\subsection{Swap order relations}

The symmetry properties of probabilities with respect to the swap of the order 
of events makes it straightforward to derive some relations that can be checked 
experimentally. However one has to be very cautious distinguishing necessary and 
sufficient conditions for such relations.  

Let us consider the alternatives $A_n$ and $B_k$, with $n,k = 1,2$, and the joint 
probability of immediate consecutive measurements (\ref{2.98}). Define the swap 
function
$$
S[\; p(A_nz_n,t_0+0,B_kz_k,t_0) \; ]  \equiv
p(A_1z_1,t_0+0,B_2z_2,t_0) - p(B_2z_2,t_0+0,A_1z_1,t_0) \; +
$$
\be
\label{2.123}
+ \;
p(A_2z_2,t_0+0,B_1z_1,t_0) - p(B_1z_1,t_0+0,A_2z_2,t_0) \;  .
\ee
Using the normalization conditions (\ref{2.73}) and (\ref{2.91}), and the 
symmetry property (\ref{2.103}) of the conditional probability, it is easy to 
get the relation
\be
\label{2.124}
 S[\; p(A_nz_n,t_0+0,B_kz_k,t_0) \; ] = 0 \;  .
\ee
Clearly, the same swap order relation is valid for the case when there is no intrinsic 
noise,
\be
\label{2.125}
 S[\; p(A_n,t_0+0,B_k,t_0) \; ] = 0 \;  .
\ee

Relation (\ref{2.125}) has been discussed by many authors, e.g. 
\cite{Busemeyer_35}. In a number of experimental studies in psychology, it has 
been found that the probability of answers to two consecutive questions depends 
on the question order, and relation (\ref{2.125}) holds true for empirical 
probabilities, because of which the following conclusion has been advocated: 
Since the validity of relation (\ref{2.125}) for a joint empirical probability 
has been confirmed in a vast number of experimental studies in psychology, and 
this relation has been derived for a quantum probability, this means that the 
empirical data prove that consciousness obeys quantum rules, that is, 
consciousness is quantum.   

This claim, though, is not correct, since here one confuses necessary and 
sufficient conditions. If some $p$ enjoys the properties of the quantum 
probability of immediate consecutive measurements, then this is sufficient for 
the relation $S[p]=0$ to be valid. However this is not a necessary condition, 
as far as, if the relation $S[p] = 0$ holds, it follows from nowhere that $p$ 
must be a particular quantum probability. 

Really, not only the probability (\ref{2.98}) of consecutive measurements 
satisfies the same relation, but also the probability (\ref{2.108}) of 
synchronous noiseless measurements,
\be
\label{2.126}
 S[\; p(A_nB_k,t) \; ] = 0  ,
\ee
as well as the probability of synchronous noisy measurements (\ref{2.118}),
\be
\label{2.127}
   S[\; p(A_nz_nB_kz_k,t) \; ] = 0 
\ee
do satisfy the same relation. The validity of the latter relations is the direct 
result of the swap order symmetry of these probabilities, as in (\ref{2.112}). 

Moreover, if we consider a classical probability $f(A_nB_k)$ that, by definition 
is swap order symmetric, $f(A_nB_k) = f(B_kA_n)$, and study the swap function
\be
\label{2.128}
S[\; f(A_nB_k) \; ] = f(A_1B_2) - f(B_2A_1) + f(A_2B_1) - f(B_1A_2) 
\ee
then, because of the swap order symmetry, the same relation immediately follows:
\be
\label{2.129}
S[\; f(A_nB_k) \; ] = 0  \; .
\ee

The swap order symmetry of quantum conditional probability of consecutive events 
is a sufficient condition for the validity of the swap order relation for the joint 
quantum probability of consecutive events. The swap order symmetry of the classical 
joint probability is also a sufficient condition for the validity of the swap order 
relation. However none of these symmetry properties separately is a necessary condition 
for the validity of the swap order relation. That is, different quantum as well as 
classical probabilities can satisfy the same swap order relation. However the validity 
of this relation tells us nothing on the nature of probability, whether it is quantum 
or classical.

\subsection{Quantum versus classical probabilities}

In the literature advocating the use of quantum techniques for describing 
consciousness, it is customary to counterpose quantum to classical approaches, 
arguing in favour of quantum theory that, supposedly, is more versatile in 
characterizing, e.g., such phenomena as non-commutativity of consecutive events. 
In doing this, one usually compares the classical Kolmogorov probability with 
the von Neumann-L\"{u}ders probability of consecutive measurements. However 
comparing these probabilities is not correct, since the classical Kolmogorov 
probability contains no dynamics, while the von Neumann-L\"{u}ders approach 
considers the dynamic evolution from one measurement to another. For the correct 
comparison of quantum and classical probabilities, it is necessary to remember 
that there are several types of the latter, so that the comparison has sense 
only for the probabilities from the same class. There are the following classes 
of probabilities. 

\vskip 2mm
(i) {\it Probability of single events}. For the quantum case, under events we mean 
quantum measurements, and in the classical case, some occurring events or the acts 
of taking decisions. The quantum probability $p(A_n,t)$ of a single event $A_n$ 
differs from the classical Kolmogorov probability $f(A_n)$ by including temporal 
evolution and by taking account of intrinsic noise. Here and below we assume that 
quantum probability includes intrinsic noise, but for the sake of compactness, we 
do not show this explicitly.

\vskip 2mm
(ii) {\it Probability of synchronous events}. Two or more synchronous events can 
be observed in the quantum case, provided they are happening in different Hilbert 
spaces, for instance in different spatial locations. In the classical Kolmogorov 
theory, the events are always synchronous. Both these probabilities, quantum as 
well as classical, enjoy the same symmetry properties.

\vskip 2mm
(iii) {\it Probability of consecutive events}. Events are happening one after 
another at different times. The times have to be treated as different even when 
one event occurs immediately after another. For quantum probability, the temporal 
evolution is incorporated in the evolution operators. Consecutive events in the 
quantum case are treated by the von Neumann-L\"{u}ders theory. In the classical 
case, the evolution can be imposed by the equations called Kolmogorov equations 
or master equations.    

\vskip 2mm
One often claims that the classical Kolmogorov probability is inferior to quantum 
von Neumann-L\"{u}ders probability because the classical joint probability does not 
depend on the order of events, being swap-order symmetric, while the quantum von 
Neumann-L\"{u}ders theory gives for the joint probability of two consecutive 
events at different times a non-symmetric order dependent probability. However this
comparison is not appropriate, since it considers the probability from different 
classes. Quantum consecutive probabilities have to be collated with classical 
consecutive probabilities that, in general, are also not swap-order symmetric. 

To illustrate the asymmetry of the classical consecutive probabilities, let us 
consider the probability of two events, one $A_n$ occurring at time $t_0$, after 
which the other event $B_k$ can happen at time $t$. The classical joint probability 
$f(B_k,t,A_n,t_0) = f(B_k,t|A_n,t_0)f(A_n,t_0-0)$ is expressed through the related 
conditional probability satisfying the master equation (or Kolmogorov forward 
equation)
\be
\label{2.130}
 \frac{d}{dt} \; f(B_k,t|A_n,t_0) = 
\sum_{l=1}^{N_B} \gm_{kl} f(B_l,t| A_n,t_0) \;  ,
\ee
in which $\gamma_{kl}$ is a transition rate matrix, or generator matrix, 
characterizing the transition rate from the event $B_l$ to $B_k$.  The transition 
rate matrix has the properties
\be
\label{2.131}
\gm_{kl} \geq 0 \qquad ( k \neq l )
\ee
and 
\be
\label{2.132}
\sum_{k=1}^{N_B} \gm_{kl} = 0 \;  .
\ee
The latter property can be rewritten as
\be
\label{2.133}
\gm_{ll} + \sum_{k(\neq l)}^{N_B} \gm_{kl} = 0 \;  ,
\ee
which allows us to represent equation (\ref{2.130}) in the equivalent form
\be
\label{2.134}
 \frac{d}{dt} \; f(B_k,t|A_n,t_0) = 
\sum_{l(\neq k)}^{N_B} [\; \gm_{kl} f(B_l,t| A_n,t_0) -
\gm_{lk} f(B_k,t| A_n,t_0) \; ] \;  .
\ee

It is instructive to observe an explicit solution, for example, considering two 
events $\{A_1, A_2\}$ and two events $\{B_1, B_2\}$ under the initial condition
\be
\label{2.135}
 f(B_k,t_0| A_n,t_0) = f_{kn} \; .
\ee
Then the solution reads as
$$
f(B_1,t| A_1,t_0) = \left( f_{11} \; - \; \frac{\gm_1}{\gm_1+\gm_2} \right)
e^{-(\gm_1+\gm_2)(t-t_0)} + \frac{\gm_1}{\gm_1+\gm_2} \; ,
$$
$$
f(B_1,t| A_2,t_0) = \left( f_{12} \; - \; \frac{\gm_1}{\gm_1+\gm_2} \right)
e^{-(\gm_1+\gm_2)(t-t_0)} + \frac{\gm_1}{\gm_1+\gm_2} \; ,
$$
$$
f(B_2,t| A_1,t_0) = \left( f_{21} \; - \; \frac{\gm_2}{\gm_1+\gm_2} \right)
e^{-(\gm_1+\gm_2)(t-t_0)} + \frac{\gm_2}{\gm_1+\gm_2} \; ,
$$
\be
\label{2.136}
f(B_2,t| A_2,t_0) = \left( f_{22} \; - \; \frac{\gm_2}{\gm_1+\gm_2} \right)
e^{-(\gm_1+\gm_2)(t-t_0)} + \frac{\gm_2}{\gm_1+\gm_2} \;   ,
\ee
where
$$
\gm_1 \equiv \gm_{12} = - \gm_{22} \; , \qquad 
\gm_2 \equiv \gm_{21} = - \gm_{11} \;   .
$$

Inverting the order of the events leads to the probability satisfying the equation
\be
\label{2.137}
\frac{d}{dt} \; f(A_n,t|B_k,t_0) = 
\sum_{m=1}^{N_A} \al_{nm} f(A_m,t|B_k,t_0) \;    .
\ee
The transition rate matrix $\alpha_{mn}$ describes the transition from an event $A_n$ 
to $A_m$ and possesses the properties
\be
\label{2.138}
\al_{mn} \geq 0 \quad ( m \neq n)
\ee
and 
\be
\label{2.139}
\sum_{m=1}^{N_A} \al_{mn} = 0 \; .
\ee
In the case of binary sets of alternatives, the solution of equation (\ref{2.137}), 
under an initial condition
\be
\label{2.140}
f(A_n,t_0|B_k,t_0) =  g_{nk} \; ,
\ee
is similar by form to solution (\ref{2.136}).   
      
From these equations, it is clearly seen that inverting the order of the events 
results in principally different expressions for the probability. This is because 
the initial conditions are different and the transition rate matrices are different. 
Generally, these matrices are different even in their size, since the size of the 
rate $\gamma_{kl}$ is $N_B \times N_B$, while the size of the rate $\alpha_{nm}$ 
is $N_A \times N_A$. The matrices of initial values, in general, are also different 
in their form, since $f_{kn}$ is a matrix of the size $N_B \times N_A$, while 
$g_{nk}$ is a matrix of the size $N_A \times N_B$.  

Thus the classical consecutive probabilities, whether conditional or joint, are not 
swap-order symmetric
\be
\label{2.141}
 f(B_k,t|A_n,t_0) \neq f(A_n,t|B_k,t_0)\; , \qquad 
 f(B_k,t,A_n,t_0) \neq f(A_n,t,B_k,t_0)
\ee
for any $t \geq t_0$. In this way, classical consecutive probabilities can 
perfectly explain the so-called order effects observed in human behaviour.

\subsection{Quantum decision theory}

The analysis of quantum probabilities described above and their interpretation in 
the language of decision theory is, strictly speaking, what composes the basis of 
the so-called Quantum Decision Theory 
\cite{Yukalov_56,Yukalov_57,Yukalov_58,Yukalov_71,Yukalov_78,Yukalov_82,Yukalov_92}.
Summarizing this part of the review, it is necessary to make several remarks.

Everywhere above the system state has been represented by a statistical operator 
$\hat{\rho}$. In particular cases, this operator could have the form of a pure state
$$
\hat\rho(t) = |\; \psi(t) \; \rgl \lgl \; \psi(t) \; | \;  ,
$$
where the wave function can be expanded over the given basis as
$$
|\; \psi(t) \; \rgl = \sum_{n\mu} b_{n\mu}(t) \; | \; A_n e_\mu \; \rgl \;  .
$$
The description by wave functions is appropriate for isolated systems. However, 
strictly speaking, quantum systems cannot be absolutely isolated, but can only be 
quasi-isolated \cite{Yukalov_120,Yukalov_121,Yukalov_122}. This means that, even 
if a system is prepared in a pure state described by a wave function, there always 
exist uncontrollable external perturbations from the surrounding that result in the 
system decoherence beyond a decoherence time, which makes the system state mixed. 
Moreover, to confirm that the considered system is to some extent isolated, it is 
necessary to check this by additional control measurements which again disturb the 
system's isolation. In that way, the system can be only quasi-isolated.

In addition, decision-makers are the members of a society, hence, they correspond 
to non-isolated open systems that have to be described by statistical operators. One 
could think that in laboratory tests, it would be admissible to treat decision-makers 
as closed systems characterized by wave functions. However, in laboratory tests, even 
when being separated from each other, decision-makers do communicate with the 
investigators performing the test. Moreover, even when being for some time locked 
in a separate room, any decision-maker possesses the memory of interactions with many 
other people before. From the physiological point of view, {\it memory is nothing but 
delayed interactions}. Therefore, no decision maker can be treated as an absolutely 
isolated system, which excludes the use of wave functions. It looks that the most 
general and correct description of any decision-maker requires to consider him/her 
as an open system, hence, characterized by a statistical operator.

When representing the considered alternatives as the vectors of a Hilbert space, we 
have assumed a nondegenerate representation with a one-to-one correspondence between 
each alternative $A_n$ and the representing it vector $|A_n\rgl$. Generally, in 
quantum theory there can occur the effect of degeneracy, when an operator eigenvalue 
can correspond to several eigenfunctions. In the present case, this would imply that 
an alternative $A_n$ would correspond to several vectors $|A_{ni}\rgl$, where 
$i=1,2,\ldots$. The existence of degeneracy in decision theory is sometimes supposed 
for removing the contradiction between the reciprocal symmetry of von Neumann-L\"uders
probability (\ref{2.101}), that is the symmetry with respect to the interchange of 
the events $A_n$ and $B_k$, and the experimentally observed absence of this reciprocal 
symmetry \cite{Boyer_44,Boyer_45}. Really, if at least one of the considered 
alternatives say $A_n$, is degenerate, such that $A_n$ is represented by a set of 
vectors $|A_{ni}\rgl$, with $i = 1,2,\ldots$, then the related projector becomes 
the sum
$$
 \hat P(A_n) = \sum_i |\; A_{ni} \; \rgl \lgl \; A_{ni} \; | \; .
$$ 
Considering, for simplicity, the noiseless case, for the von Neumann-L\"uders 
probability (\ref{2.99}) we get
$$
 p(B_k,t_0+0|A_n,t_0) = 
\frac{\sum_{ij}\lgl A_{ni}|\hat\rho(t_0-0)| A_{nj}\rgl
\lgl A_{nj}|B_k\rgl \lgl B_k|A_{ni}\rgl}
{\sum_i\lgl A_{ni}|\hat\rho(t_0-0)|A_{ni}\rgl} \; .
$$
Reversing the order of events yields a different expression
$$
 p(A_n,t_0+0|B_k,t_0) =  
\sum_i |\; \lgl \; A_{ni}\; | \; B_k \; \rgl \; |^2 \; .
$$

However, although the occurrence of degenerate operator spectra is natural for quantum 
systems, in decision theory the appearance of degenerate alternatives has no sense. If 
there occur several vectors $|A_{ni}\rangle$, it is always admissible to reclassify
the given alternatives so that each vector $|A_{ni}\rangle$ would correspond to a
single alternative $A_{ni}$ \cite{Yukalov_68,Yukalov_69}. This is equivalent to the 
breaking of symmetry in physics \cite{Neumann_49,Bogolubov_65,Bogolubov_66,Bogolubov_67}. 
    
The measurement procedure has been described by using projection-valued measurements 
of alternatives decorated by intrinsic noise. In general, it could be possible to 
invoke positive operator-valued measurements (POVM)
\cite{Williams_85,Nielsen_86,Vedral_87,Keyl_88,Mintert_89,Guhne_90,Wilde_91}. This, 
probably, could be useful for some quantum systems. However, as has been stressed 
above, we do not assume that the brain is a quantum system, but we are analyzing 
the possibility of employing quantum techniques for describing the operation of 
the decision-making process. For this purpose, there is no reason to complicate 
the consideration by invoking POVM bringing additional problems, such as the 
nonuniqueness of the post-measurement states and the absence of reproducibility 
of immediately repeated measurements.

Moreover, already on the level of projection-valued measurements, we meet a number 
of difficulties in the attempts of applying quantum theory for describing conscious 
processes, although certainly there are many similarities on the general qualitative 
level, as has been mentioned many times above. Summarizing, it is necessary to 
separate grains from tares by clearly formulating what are the useful recipes 
following from the quantum theory of measurements and what are the limitations in 
the attempts of their use for characterizing the operation of artificial intelligence. 
The main conclusions that can be derived from the analogies between quantum measurements 
and behavioural decision making, are as follows. 

\vskip 2mm
(i) First of all, the process of decision making has to be treated in a 
probabilistic way. The probabilistic description better corresponds to real life, 
where in any sufficiently large group of subjects, deciding on a choice among the 
same set of alternatives, not all prefer the same choice, but always there are those 
who choose other alternatives. Any such group separates into subgroups preferring 
different alternatives. The fractions of people preferring different alternatives 
are nothing but frequentist probabilities. Even a single person at different moments 
of time can choose different alternatives. In that case, the frequentist probability 
shows the ratio of particular choices to the total number of accomplished choices. 

\vskip 2mm
(ii) Generally, decision making is not a purely rational choice, but it is accompanied 
by intrinsic noise representing irrational subconscious sides of decision process, 
including emotions, gut feelings, and intuitive allusions. In that sense, decision 
making is characterized by cognition-emotion duality, or conscious-subconscious 
duality, or rational-irrational duality. Emotions and the related characteristics 
can be modeled by intrinsic noise in quantum measurements.  

\vskip 2mm
(iii) Quantum probability for measurements in the presence of intrinsic noise 
consists of two terms, one that can be called classical limit and the other caused 
by the interference of noise. The former can be associated with the rational choice 
and the other is induced by the existence of emotions. In decision theory, the 
occurrence of the additional interference term is especially noticeable under 
uncertainty \cite{Yukalov_83,Yukalov_123}. 

\vskip 2mm
(iv) Alternatives and noise, or alternatives and emotions, generally are entangled
being connected with each other. Measurement procedure as well as decision making
produce additional entanglement between alternatives and noise. 

\vskip 2mm
(v) It is necessary to distinguish between two types of quantum probabilities 
for two events. One type is the quantum probability of consecutive events happening 
at different times and the other type is the quantum probability of synchronous 
events occurring at different spatial locations or at different spaces of variables.

\vskip 2mm
Despite a number of hints on the general structure and main properties of 
probability, which could be used in developing decision theory, quantum theory 
provides no explicit rules allowing for the calculation of the probability for 
the purpose of decision making, thus possessing no predictive power. It is of 
course possible to fit the interference term for interpreting some particular 
events, however fitting is not explanation. In order to supply quantum decision 
theory with the ability of making quantitative estimates, it is necessary to 
invoke a number of assumptions not related to quantum theory 
\cite{Yukalov_77,Yukalov_124}. 

It is also necessary to understand whether the usage of quantum theory is 
compulsory for the development of adequate decision theory taking account of 
behavioural effects or this usage is just a much more complicated trendy way 
of describing what could be much easier described employing classical terminology.             
 
A great goal would be to develop a mathematical formulation of an approach, 
taking into account the general peculiarities of quantum measurements and the 
properties of quantum probabilities, but explicitly involving no quantum 
techniques, at the same time providing the ability of quantitative predictions. 
Such a formalized approach would be indispensable for the creation of affective 
artificial intelligence.

\section{Affective decision making}

Before formulating the general approach to decision making, combining the 
rational choice with the influence of irrational emotions, it is useful to 
remind the origin of emotions and to explain what are the problems in the 
earlier attempts to formalize the process of behavioural decision making.

\subsection{Evolutionary origin of emotions}

Emotions are common for humans as well as for animals. They evolved in the 
process of evolution and were adapted over time like other traits found in 
animals. Darwin \cite{Darwin_1872} was, probably, the first to seriously study 
the appearance and adaptation of emotions in the process of natural selection. 
He discussed not only facial expressions in animals and humans, but attempted 
to point out parallels between behaviours in humans and other animals 
\cite{Hess_2009}. According to evolutionary theory, different emotions evolved 
at different times. Primal emotions, such as fear, are associated with ancient 
parts of the brain and presumably evolved among our premammal ancestors. Filial 
emotions, such as a human mother's love for her offspring, seem to have evolved 
among early animals. Social emotions, such as guilt and pride, evolved among 
social primates. 

Since emotions evolved and adapted during the years of evolution, they appeared 
for some reason, and as other features, they should be useful for animals and 
humans. For example, they facilitate communication by sending signals to other 
members of the social group. Such an emotion as fear has helped humans to 
survive, warning about a danger and forcing to take actions before the cognitive 
logical part of the brain gives more detailed information. Having emotions may 
mean the difference between life and death. Certain emotions are universal to 
all humans, regardless of culture: anger, fear, surprise, disgust, happiness and 
sadness. Emotions can be defined as a specialized mechanism, shaped by natural 
selection, that increases fitness in specific situations. The physiological, 
psychological, and behavioural characteristics of emotions can be understood as 
possible design features that increase the ability to cope with the threats and 
opportunities present in the corresponding situations. Every emotion has been 
developed individually in the course of biological evolution, and they all have 
been evolved to maintain the survival needs. 

Emotions play an important role in decision making. It would be not an 
exaggeration to say that emotions shape decisions. As in the mentioned above 
example of fear that saves lives, the fear can also save from bankruptcy. Thus, 
compelling scientific evidence comes from emotionally impaired patients who have 
sustained injuries to the ventromedial prefrontal cortex, a key area of the 
brain for integrating emotion and cognition. Studies find that such neurological 
impairments reduce both the patients' ability to feel emotion and, as a result, 
reduce the optimality of their decisions. Participants with these injuries 
repeatedly select a riskier financial option over a safer one, even to the point 
of bankruptcy, despite their cognitive understanding of the suboptimality of 
their choices. These participants behave this way because they do not experience 
the emotional signals that lead normal decision makers to have a reasonable fear 
of high risks \cite{Lerner_11}.   

Living in a world where events cannot be predicted with certainty, agents must 
select actions based on limited information, i.e., they often must make risky 
decisions. Emotional information has a special weight in decision-making, as it 
automatically triggers adaptive behavioral modules selected during the course of 
evolution, driving agents to move toward attractive goals while avoiding threats. 
Emotional information is critical, because on the one hand it could prevent 
potential physical harm or unpleasant social interactions, on the other hand 
it could promote physical pleasure or pleasant social interactions 
\cite{Mirabella_2018}. 

Emotions are generally classified onto positive and negative \cite{Wentura_19}, 
respectively making alternatives more attractive or more repulsive. In the 
choice between alternatives, there is no one-to-one correspondence between 
alternatives and emotions, but there appears a multitude of different connected 
emotions. Researchers exploring the subjective experience of emotions have noted 
that emotions are highly intercorrelated both within and between the subjects 
reporting them. Subjects rarely describe feeling a specific positive or negative 
emotion without also claiming to feel other positive or negative emotions 
\cite{Garcia_2010}.  

In the process of decision making, emotions have great influence on multiple 
cognitive phenomena, such as attention, perception, memory encoding, storage, 
and retrieval of information, and associative learning \cite{Garcia_2020}. 
Emotions activate the motivational system of action tendencies. Recall that 
the word emotion comes from Latin ``emovere", which means to move. The origin 
of the word emotion already emphasizes its relevance to behavioural drive. 

Although emotions seem to be similar to noise, they in fact help to optimize 
decisions in two ways. First, they are faster than logical rational deliberations, 
thus being of crucial importance in the case of urgent decisions. Second, emotions 
reflect subconscious feelings based on your past experiences and beliefs. This 
might serve to protect you from danger or prevent your repeating past mistakes.  

Notice that in physical measurements noise also is not always an obstacle, but 
sometimes the detection of signals can be boosted by noise so that their detection 
can even be facilitated \cite{Buchleitner_52,Gardiner_53}. Measurements of thermal 
noise had been used to measure the Boltzmann constant and measurements of shot 
noise had been used to measure the charge on the electron \cite{Endelberg_2004}. 
Noise plays beneficial role in the functioning of neural systems in the framework 
of stochastic facilitation of signal processing \cite{Mcdonnell_2011}.

\subsection{Problems in decision making}

The predominant theory describing individual behaviour under uncertainty is nowadays 
the expected utility theory of preferences over uncertain prospects. This theory  
was axiomatized by von Neumann and Morgenstern \cite{Neumann_1953} and integrated 
with the theory of subjective probability by Savage \cite{Savage_1954}. The theory 
was shown to possess great analytical power by Arrow \cite{Arrow_1971} and Pratt 
\cite{Pratt_1964} in their work on risk aversion and by Rothschild and Stiglitz 
\cite{Rothschild_1970,Rothschild_1971} in their work on comparative risk. Friedman 
and Savage \cite{Friedman_1948} and Markowitz \cite{Markowitz_1952} demonstrated its 
tremendous flexibility in representing decision makers attitudes toward risk. It is 
fair to state that the expected utility theory has provided solid foundations to the 
theory of games, the theory of investment and capital markets, the theory of search, 
and other branches of economics, finance, and management.

However, a number of economists and psychologists have uncovered a growing body of 
evidence that individuals do not always conform to the prescriptions of the expected 
utility theory and indeed very often depart from the theory in a predictable and 
systematic way \cite{Ariely_30}. Many researchers, starting with the works by Allais 
\cite{Allais_1953}, Edwards \cite{Edwards_1955,Edwards_1962}, and Ellsberg 
\cite{Ellsberg_1961} and continuing through the present, experimentally confirmed 
pronounced and systematic deviations from the predictions of the expected utility 
theory leading to the appearance of many paradoxes. These paradoxes are often called 
behavioural, since the behaviour of subjects contradicts the prescriptions of the 
utility theory. Large literature on this topic can be found in the review articles 
\cite{Slovic_1974,Camerer_2003,Miljkovic_2005,Machina_2008}.

There were many attempts to change the expected utility approach, which were
classified as non-expected utility theories. There are a number of such non-expected 
utility theories, among which we may mention a few of the most known ones: prospect 
theory \cite{Edwards_1955,Kahneman_1979,Tversky_1992}, weighted-utility theory 
\cite{Karmarkar_1978,Karmarkar_1979,Chew_1983}, regret theory \cite{Loomes_1982}, 
optimism-pessimism theory \cite{Hey_1984}, ordinal-independence theory \cite{Green_1988}, 
quadratic-probability theory \cite{Chew_1991}, opportunity-threat theory 
\cite{Pandey_2018}, and state-dependent utility theory \cite{Hill_2019}. The general 
discussion of these theories can be found in the review by Machina \cite{Machina_2008}.

However, non-expected utility theories are descriptive requiring fitting of several 
parameters from empirical data. Moreover, as was shown by Safra and Segal 
\cite{Safra_2008}, none of the non-expected utility theories can explain all 
paradoxes permanently arising in behavioural decision making. The best that could be 
achieved is a kind of fitting for interpreting just one or, at best, a few paradoxes, 
with other paradoxes remaining unexplained. In addition, spoiling the structure of 
expected utility theory results in the appearance of several complications and 
inconsistences \cite{Birnbaum_2008}. As was concluded in the detailed analysis of 
Al-Najjar and Weinstein \cite{Al_2009a,Al_2009b}, any variation of the classical 
expected utility theory "ends up creating more paradoxes and inconsistences than 
it resolves". 
 
\begin{sloppypar}
An attempt of taking into account unconscious feelings has been undertaken by the 
approach called the dual-process theory 
\cite{Sun_2002,Paivio_2007,Evans_2007,Stanovich_2011,Kahneman_2011}. According 
to this theory, decisions arise in the human brain as a result of two different 
processes that can be distinguished by one of the following characteristic pairs 
\cite{Kahneman_2011}: slow/fast, rational/irrational, conscious/unconscious, 
logical/intuitive, reasoning/reasonless, deliberate/emotional, 
intentional/unintentional, voluntary/involuntary, explicit/implicit, 
analyzing/sensuous, controlled/uncontrolled, operated/automatic, regulated/impulsive, 
effortful/effortless, comprehensive/perceptional, precise/impressionistic, 
objective/subjective, verbal/nonverbal. Of course, not all these characteristics 
have to be present in one or another process. Some of them can be shared to some 
extent by both ways of thinking. Detailed discussions of the description of the 
two processes can be found in the literature \cite{Sun_2002,Paivio_2007,Evans_2007,
Stanovich_2011,Kahneman_2011,Zafirovski_2012,Bialek_2017,Gurcay_2017} and are well 
exposed in the review articles \cite{Vaisey_2009,Evans_2012,Evans_2013}. The 
existence of two ways of thinking finds support in neuropsychological studies 
\cite{Goel_2000,Goel_2003,Libert_2006,Milner_2008,Tsujii_2009}, although such a 
separation is not very strict \cite{Leventhal_16,Pessoa_17,Brosch_18}.
\end{sloppypar}

Thus, in the dual-process theory one accepts the existence of two ways of thinking, 
which to some extent finds support in some psychological and neurological studies. 
These ways, for brevity, can be termed one as cognitive, rational, logical and the 
second as emotional, irrational, intuitive. This separation has to be understood in 
the conditional operational sense applied to the process of decision making. The 
rational way of thinking is normative, being based on clearly prescribed logical 
rules, while the irrational way is poorly controlled, representing emotions induced 
by intuition and a kind of gut feelings. This distinction does not assume that one 
of the ways, say cognitive, is more correct. 

In psychophysical and neurophysiological studies it has been found that the influence 
of emotions leads to random choice, with the randomness caused by generic variability 
and local instability of neural networks 
\cite{Werner_1963,Arieli_1986,Gold_2001,Glimcher_2005,Schumacher_2011,Shadlen_2016,
Webb_2019,Kurtz_2019}. The choice varies not only for different individuals, but also 
for the same subject at different moments of time. Moreover, even a given subject 
when making a single decision, experiences intrinsic noise in the brain neural network, 
because of which the subject decision becomes probabilistic. Cognitive imprecision 
is due to the inevitable internal noise in the nervous system. Stochasticity is the 
unavoidable feature of the human brain functioning. As a result, the choice in 
decision making is not deterministic, being based on the comparison of utilities, 
but it is rather stochastic and based on the comparison of probabilities. 
The recent review \cite{Woodford_2020} summarizes the modern point of view 
that considers randomness as an internal feature of functioning of the human brain, 
where decisions are formed on the basis of noisy internal representation. 
  
The standard method employed in the attempts of taking into account irrational 
effects in human decision making is a modification of the utility functional 
\cite{Kurtz_2019,Yaari_1987,Reynaa_2011}. In that sense, the dual-process models 
are reduced to variants of non-expected utility theories, sharing with the latter 
the same deficiencies. 

Summarizing, in order to develop a decision theory comprising both ways of decision 
process, conditionally labeled as cognitive (rational) and emotional (irrational), 
one should respect the following premisses:

\begin{enumerate}
\item
Since the presence of emotions results in random decisions, behavioural decision 
making has to be treated as a generically probabilistic process. Therefore the main 
quantity to be determined is the behavioural probability of events. It is never
happens that among a given group of people all without exception would make the
identical choice prescribed by the standard deterministic utility theory. There 
always exist fractions of subjects preferring different alternatives. That is, 
there always exists a distribution of decisions over the set of the given 
alternatives.

\item
The behavioural probability of an event should reflect the superposition of two 
operational aspects in decision making, rational (cognitive), defined by the 
prescribed rules evaluating the utility of the event, and irrational (emotional), 
taking account of irrational effects.     

\item
As far as emotions randomly vary for different decision makers, as well as for 
the same decision maker at different moments of time, their quantitative influence 
cannot be predicted exactly for each subject and for each choice. However the 
approach should provide quantitative predictions at the aggregate level, when the 
behavioural probabilities could be compared with the empirical average fractions 
of decision makers choosing the related alternatives. 

\item
The efficiency of the approach should be proved by demonstrating the absence of 
paradoxes that should find quantitative resolution in the framework of the general 
methodology without fitting parameters. 

\item
The last, but not the least, the approach should not be overloaded by unnecessary 
complicated theorizing. Thus, borrowing from the theory of quantum measurements and 
quantum decision theory some general ideas, it is desirable to avoid the explicit use 
of quantum techniques. In that sense, an artificial intelligence could accomplish 
quantum operation without the necessity of involving quantum formulas \cite{Yukalov_125}.  

\end{enumerate}

\subsection{Behavioural probabilities of alternatives}

As is emphasized above, behavioural decision making is a principally probabilistic 
process. Hence the pivotal role is played by the notion of probability of 
alternatives. In decision theory, probabilistic approach is usually based on the 
random utility model, where the expected utility of alternatives is complimented 
by an additive random error characterized by a postulated distribution 
\cite{Fadden_1976,Hensher_2005,Train_2009}. This approach, being based on expected 
utility theory, contains the same deficiencies as the underlying utility theory: it 
does not take into account emotions and does not explain behavioural paradoxes. In 
addition, it contains several fitting parameters making the approach descriptive but
not predictive. 

The classical approach, axiomatically formulated by Kolmogorov \cite{Kolmogorov_1956}, 
defines the probabilities satisfying three axioms, non-negativity, normalization, and 
additivity. For behavioural probabilities, in general, it is sufficient to satisfy only 
two axioms, non-negativity and normalization, with the additivity property as a compulsory 
condition, being dropped out 
\cite{Schmeidler_1989,Gilboa_1993,Viscusi_2006,Yukalov_57,Yukalov_68,Yukalov_78,Yukalov_126}. 
Throughout the paper, the standard classical probabilities, sometimes called just 
probabilities, are distinguished from behavioural probabilities. To be precise, the basic 
points of the approach are formulated in axiomatic way that, although keeping in mind the 
properties of quantum behavioural probabilities, at the same time does not involve any 
quantum notions explicitly.
 
Let us consider a ring $\{A_n\}$ of alternatives $A_n$. Assume that decision making 
consists in the choice between the alternatives of a set
\be
\label{3.1} 
 \mathbb{A} = \{A_n: ~ n = 1,2,\ldots , N_A \} \;  .
\ee

\vskip 2mm

{\bf Axiom 1}. Each alternative $A_n$ is equipped with its behavioural probability 
$p(A_n)$, whose family forms a probability measure over the set $\{A_n\}$, with the 
properties of non-negativity and normalization
\be
\label{3.2}
 \sum_{n=1}^{N_A} p(A_n) =  1 \; , \qquad 0 \leq p(A_n) \leq 1 \;  .
\ee

\vskip 2mm

It is assumed that each alternative is decorated by emotions, which for the sake of 
notation compactness is not explicitly marked. 

\vskip 2mm  

{\bf Axiom 2}. The alternatives are connected by relations defined through the 
relations between their behavioural probabilities. The set of alternatives (\ref{3.1}) 
enjoys the following properties.
   
\vskip 2mm 
\begin{enumerate}
\item
{\it Ordering}: For any two alternatives $A_1$ and $A_2$, one of the relations 
necessarily holds: either $A_1 \prec A_2$, in the sense that $p(A_1) < p(A_2)$, or 
$A_1\preceq A_2$, when $p(A_1)\leq p(A_2)$, or $A_1\succ A_2$, if $p(A_1)>p(A_2)$,
or $A_1\succeq A_2$, when $p(A_1)\geq p(A_2)$, or $A_1\sim A_2$, if $p(A_1)=p(A_2)$.

\item
{\it Linearity}: The relation $A_1 \preceq A_2$ implies $A_2 \succeq A_1$. These 
and the relations below are to be understood as relations between the corresponding 
probabilities $p(A_n)$.

\item
{\it Transitivity}: For any three alternatives, such that $A_1 \preceq A_2$, with 
$p(A_1) \leq p(A_2)$, and $A_2 \preceq A_3$, when $p(A_2) \leq p(A_3)$, it follows 
that $A_1 \preceq A_3$, in the sense that $p(A_1) \leq p(A_3)$.

\item
{\it Completeness}: In the set of alternatives (\ref{3.1}), there exist a minimal 
$A_{min}$ and a maximal $A_{max}$ elements, for which $p(A_{min}) = \min_n p(A_n)$ 
and, respectively, $p(A_{max}) = \max_n p(A_n)$.
\end{enumerate}

\vskip 2mm

The ordered set of alternatives (\ref{3.1}), enjoying these properties, is called a 
complete lattice. 

\vskip 2mm

{\bf Definition 1}. An alternative $A_1$ is called stochastically preferable to $A_2$ 
if and only if
\be
\label{3.3}
 p(A_1) > p(A_2) \qquad  (A_1 \succ A_2) \; .
\ee  

\vskip 2mm

{\bf Definition 2}. Two alternatives are stochastically indifferent if and only if
\be
\label{3.4}
 p(A_1) = p(A_2) \qquad  (A_1 \sim A_2) \; .
\ee

\vskip 2mm

{\bf Definition 3}. The alternative $A_{opt}$ is called stochastically optimal if 
it corresponds to the maximal behavioural probability,
\be
\label{3.5}
p(A_{opt}) = \max_n p(A_n) \; .
\ee
 
\vskip 2mm

Behavioural decision making includes both rational reasoning, following prescribed 
logical rules, as well as irrational inclinations not rationalized by explicit 
logical argumentation, such as emotions, subconscious guesses, gut feelings, 
intuition, etc, all of which we shall, for brevity, call emotions.

\vskip 2mm

{\bf Definition 4}. Rational reasoning for an alternative $A_n$ is described by a 
rational fraction named {\it utility factor} $f(A_n)$ that is a classical probability 
of choosing an alternative $A_n$, being based on rational rules. A collection of 
rational fractions for a given set of alternatives forms a classical probability 
measure over the set $\{A_n\}$ with the properties
\be
\label{3.6}
 \sum_{n=1}^{N_A} f(A_n) =  1 \; , \qquad 0 \leq f(A_n) \leq 1 \; .
\ee

\vskip 2mm

Emotion categories are fuzzy and are labeled with words, expressions, and metaphors
\cite{Scherer_2009,Scherer_2019,Moors_2019}. When comparing the emotions induced by 
different alternatives, one cannot quantify them by exact numbers but one can only
characterize them in descriptive terms, e.g., as attractive or repulsive, pleasant 
or unpleasant, and like that. Emotion processes do not enjoy clear categorical 
boundaries \cite{Scherer_2009,Scherer_2019,Moors_2019}.   

\vskip 2mm

{\bf Definition 5}. Emotional impulses in choosing an alternative $A_n$ are 
characterized by an {\it attraction factor} $q(A_n)$ lying in the interval
\be
\label{3.7}
 -1 \leq q(A_n) \leq 1 \;  .
\ee

\vskip 2mm

{\bf Axiom 3}. Behavioural probability of choosing an alternative $A_n$, taking 
account of rational reasoning as well as emotion influence, is a functional of the 
utility factor $f(A_n)$ and attraction factor $q(A_n)$ satisfying the limiting 
condition
\be
\label{3.8}
 p(A_n) \; \mapsto \; f(A_n) \; , \qquad q(A_n) \; \mapsto \; 0 \; .
\ee

This is an analog of decoherence in quantum theory, when the quantum term, causing 
interference, vanishes and the quantum quantity tends to its classical form.   

\vskip 2mm

{\bf Axiom 4}. Behavioural probability of choosing an alternative $A_n$, taking 
account of emotions and satisfying the limiting condition (\ref{3.8}), is the sum
\be
\label{3.9}
 p(A_n)= f(A_n) + q(A_n) \;  .
\ee

\vskip 2mm

From the inequality $0 \leq p(A_n) \leq 1$, it follows that 
\be
\label{3.10}
- f(A_n) \leq q(A_n) \leq 1 - f(A_n) 
\ee 
in agreement with inequality (\ref{3.7}).   

The value of the rational fraction $f(A_n)$ shows how useful the alternative $A_n$ 
is, because of which it is called the {\it utility factor}. The magnitude and sign 
of the attraction factor $q(A_n)$ characterize how attractive the alternative $A_n$ 
is, hence $q(A_n)$ is termed the {\it attraction factor}.

\subsection{Quantification of utility factor}

The utility factor, describing the rational utility of each alternative, has to 
be defined by prescribed rules. Here we show how these fractions can be determined 
through expected utilities or other value functionals. For concreteness, we shall 
be talking about the expected utility of an alternative $U(A_n)$, although in the 
place of the expected utility one can take any value functional. 

Let the alternatives be represented by the lotteries
\be
\label{3.11}
 A_n = \{x_i , \; p_n(x_i): ~ i = 1,2,\ldots , N_n \} \;  ,
\ee
being the probability distributions over the payoff set $\{x_i\}$ with the 
properties
$$
 \sum_{i=1}^{N_n} p_n(x_i) =  1 \; , \qquad 0 \leq p_n(x_i) \leq 1 \; .
$$
The probabilities $p_n (x)$ can be either objective \cite{Neumann_1953} or 
subjective \cite{Savage_1954}. Following the classical utility theory 
\cite{Neumann_1953}, one can introduce a dimensionless utility function $u(x)$ 
and the expected utility
\be
\label{3.12}    
U(A_n) = \sum_i u(x_i) p_n(x_i) \; .
\ee

The utility factor reflects the utility of a choice, hence it has to be a functional 
of the expected utility. As a functional of the expected utility, the utility factor
has to satisfy the evident conditions
\be
\label{3.13}
  f(A_n) \ra 1 \; , \qquad U(A_n)\ra \infty 
\ee
and
\be
\label{3.14}
  f(A_n) \ra 0 \; , \qquad U(A_n)\ra - \infty  \; .
\ee

The utility factor plays the role of classical probability expressed through 
the expected utility $U(A_n)$ or another value functional. The general approach 
of defining the probability distribution describing utility factor is the 
minimization of an information functional including imposed constraints 
\cite{Yukalov_127,Yukalov_128}. The first such a natural constraint is the 
normalization condition (\ref{3.6}). Another constraint is the existence of 
a global mean
\be
\label{3.15}
 \sum_n f(A_n) U(A_n) = U \qquad ( |\; U \; | < \infty ) \; .
\ee
The explicit expression for an information functional can be taken in the 
Kullback–Leibler form \cite{Kullback_1951,Kullback_1959,Shore_1980}. The 
Shore-Jonson theorem \cite{Shore_1980} states that, given a prior (or trial) 
probability density $f_0(A_n)$ and additional constraints, there is only one 
posterior density $f(A_n)$ satisfying these constraints and the conditions of 
uniqueness, coordinate invariance, and system independence, such that this 
unique posterior can be obtained by minimizing the Kullback-Leibler information 
functional. The posterior probability $f(A_n)$ is the minimizer of the 
Kullback–Leibler functional provided the imposed constraints do not contain 
singularities, which requires a not divergent value of the global mean $U$. 

It is important to stress that the existence of the global mean $U$ does not 
impose any constraints on the expected utilities $U(A_n)$ that can be divergent,
as for instance in the St. Petersburg paradox to be considered below. The 
existence of the global $U$ is required for the uniqueness of the probability
$f(A_n)$ in the Shore-Jonson theorem \cite{Shore_1980}. 

In the present case, the information functional for the posterior probability 
distribution $f(A_n)$, under a prior distribution $f_0(A_n)$ and the given 
constraints, 
is
$$
I[\; f(A_n) \; ] = \sum_n f(A_n) \ln \; \frac{f(A_n)}{f_0(A_n)} + 
\al \left[ \; 1 - \sum_n f(A_n) \; \right] \; +
$$
\be
\label{3.16}
  + \;
\bt \left[ \; U - \sum_n f(A_n) U(A_n) \; \right] \; ,
\ee
where $\alpha$ and $\beta$ are Lagrange multipliers. The minimization of the 
information functional (\ref{3.16}) yields
\be
\label{3.17} 
 f(A_n) = 
\frac{f_0(A_n) e^{\bt U(A_n)} }{\sum_n f_0(A_n) e^{\bt U(A_n)} } \; .
\ee
  
The trial distribution $f_0(A_n)$ can be defined employing the Luce rule 
\cite{Luce_1959,Luce_1989,Gul_2014}. Let the attribute of an expected utility 
$U(A_n)$ be characterized by an attribute value $a_n$ assumed to be non-negative. 
Then, according to the Luce rule \cite{Luce_1959,Luce_1989,Gul_2014}, the trial 
utility factor can be defined  as 
\be
\label{3.18}
f_0(A_n) = \frac{a_n}{\sum_{n=1}^{N_A} a_n} \qquad ( a_n \geq 0 ) \;  .
\ee
The attribute value depends on whether the corresponding utility is positive 
(semi-positive) or negative. For a semi-positive utility the attribute value can 
be defined \cite{Yukalov_129} as
\be
\label{3.19} 
a_n = U(A_n) \; , \qquad U(A_n) \geq 0 \; ,
\ee 
while for a negative expected utility it can be given by
\be
\label{3.20}
a_n = \frac{1}{|\; U(A_n)\; |} \; , \qquad U(A_n) < 0 \; .
\ee

For example, in the case of two lotteries, we have
\be
\label{3.21}
 f_0(A_n) = \frac{U(A_n)}{U(A_1) + U(A_2)} \; , \qquad U(A_n) \geq 0 \; 
\ee
for semi-positive utilities, and
\be
\label{3.22}
f_0(A_n) = 1 - \frac{|\;U(A_n)\;|}{|\;U(A_1)\;| + |\;U(A_2)\;|} \; , \qquad 
U(A_n) < 0 \; 
\ee
for negative utilities. 

\vskip 2mm

{\bf Definition 6}. A lottery $A_1$ is more useful than $A_2$ if and only if
\be
\label{3.23}
f(A_1) > f(A_2) \; .
\ee

\vskip 2mm

{\bf Definition 7}. Two lotteries $A_1$ and $A_2$ are equally useful if and only if
\be
\label{3.24}
f(A_1) = f(A_2) \; .
\ee

\vskip 2mm

As is evident, a lottery can be more useful but not preferable, since the behavioural 
probability consists of two terms, including a utility factor and an attraction 
factor. Generally, the rational fraction could be taken in a different form. However 
the considered Luce form, probably, is the simplest containing no fitting parameters 
and sufficient for providing quite reliable estimates, as will be shown below.    

The utility factor (\ref{3.17}), with the trial distribution (\ref{3.18}) and 
attributes (\ref{3.19}) and (\ref{3.20}), for non-negative utilities, reads as
\be
\label{3.25}
f(A_n) = \frac{U(A_n) e^{\bt U(A_n)} }{\sum_n U(A_n) e^{\bt U(A_n)}} \; ;
\qquad U(A_n) \geq 0 
\ee
and for negative utilities, as
\be
\label{3.26}
f(A_n) = 
\frac{|U(A_n)|^{-1} e^{-\bt|U_n|} }{\sum_n |U(A_n)|^{-1} e^{-\bt |U_n|}} \; ;
\qquad U(A_n) < 0 \; .
\ee

The parameter $\beta$, called {\it belief parameter}, characterizes the level of 
certainty of a decision maker in the fairness of the decision task and in the subject 
confidence with respect to his/her understanding of the overall rules and conditions 
of the decision problem. The absolute certainty of a decision maker is characterized
by $\beta \ra \infty$, when
\begin{eqnarray}
\label{3.27}
f(A_n) = \left\{ \begin{array}{ll}
1 , ~ & U(A_n) = \max_n U(A_n) \\
\\
0 , ~ & U(A_n) \neq \max_n U(A_n)
\end{array} \right. \qquad (\bt \ra \infty) \; ,
\end{eqnarray}
so that we return to the standard deterministic decision theory prescribing to choose 
the alternative with the maximal expected utility. The case of absolute uncertainty
corresponds to $\beta \ra -\infty$, when
\begin{eqnarray}
\label{3.28}
f(A_n) = \left\{ \begin{array}{ll}
1 , ~ & U(A_n) = \min_n U(A_n) \\
\\
0 , ~ & U(A_n) \neq \min_n U(A_n)
\end{array} \right. \qquad (\bt \ra -\infty) \; .
\end{eqnarray}
In the intermediate case of neutral attitude of a decision maker, $\bt=0$. Then 
for non-negative expected utilities,
\be
\label{3.29}  
 f(A_n) = \frac{U(A_n)}{\sum_n U(A_n)} \; ; \qquad 
U(A_n) \geq 0 \; , \qquad \bt = 0 \; ,
\ee
while for negative expected utilities,
\be
\label{3.30}
f(A_n) = \frac{|U(A_n)|^{-1}}{\sum_n |U(A_n)|^{-1}} \; ; \qquad 
U(A_n) < 0 \; , \qquad \bt = 0 \;  .
\ee

The described way of constructing utility factors serves as a typical example. 
Other variants are admissible, as will be mentioned below. Recall that, instead 
of expected utility, other value functionals can be used, e.g. the value 
functional of prospect theory \cite{Kahneman_1979}.

\subsection{Magnitude of attraction factor}

In order to calculate the behavioural probability (\ref{3.9}), we need to know 
the value of the attraction factor $q(A_n)$. Generally, this quantity is varying, 
being distributed over interval (\ref{3.10}). It may vary for different decision 
makers, different problems, and even for the same problem and the same decision 
maker at different times. Nevertheless, even varying does not preclude the quantity 
from possessing some well defined features on average \cite{Yukalov_92,Yukalov_124}. 

\vskip 2mm

{\bf Theorem 1}. {\it The sum of attraction factors over a given set of alternatives 
$\mathbb{A}$ is zero:
\be
\label{3.31}
\sum_{n=1}^{N_A} q(A_n) = 0 \;  .
\ee
}

\vskip 2mm

{\it Proof}. This property that can be called the {\it alternation law} directly 
follows from the normalization conditions (\ref{3.2}) and (\ref{3.6}) and the form 
of the behavioural probability (\ref{3.9}). $\square$ 

\vskip 2mm

Modal emotions compose a fuzzy set on the background of an infinite variety of 
emotional processes and their qualia representations 
\cite{Scherer_2009,Scherer_2019,Moors_2019}. Therefore the attraction factor 
characterizing this fuzzy set of emotions seems to be not representable by fixed 
numbers. Being idiosyncratic and vague, emotions are elicited on the basis of 
individual subjective evaluations of alternatives. However although emotions are 
subjective and may vary from individual to individual, they must remain within 
certain limits appropriate to the objective situation  
\cite{Scherer_2009,Scherer_2019,Moors_2019}. In other words, although the 
attraction factor is rather a random quantity, it can be characterized, as any 
random variable, by its average value that can be accepted as a non-informative 
prior estimate. Below we keep in mind the simple arithmetic average. Recall that 
if a variable lies in the range $[a,b]$, then its arithmetic average is given by 
$(a+b)/2$. A useful property is the additivity of the average, according to which 
the average of a sum equals the sum of the averages. 
 
The attraction factor can be positive or negative. If it is positive, its average 
value will be denoted as $q_+ > 0$, while when it is negative, its average value 
will be denoted as $q_- < 0$. These quantities can be considered as non-informative 
priors for estimating the attraction factor. An emotionally neutral alternative
corresponds to $q = 0$. 

It is necessary to stress the importance of estimating the typical, or average 
value of the attraction factor. As is evident, this factor varies for different 
subjects, because of which it cannot be derived from theoretical considerations 
for a single decision maker. However in a probabilistic approach what we need is 
an estimate for the average, or typical value of the factor, which can be used 
as a prior approximation. The average value of the attraction factor magnitude 
is estimated in the theorem below.   

\vskip 2mm

{\bf Theorem 2}. {\it The non-informative priors for the positive and negative attraction 
factors are
\be
\label{3.32}
q_+ = \frac{1}{4} \; , \qquad q_- = -\; \frac{1}{4} \;  .
\ee
 }
\vskip 2mm

{\it Proof}. By agreement, the non-informative priors are given by the corresponding 
average values. From the definition of the behavioural probability (\ref{3.9}) and 
normalization conditions (\ref{3.2}) and (\ref{3.6}), it follows that a positive 
attraction factor of an alternative $A_n$ varies in the interval 
$0 < q(A_n)\leq 1-f(A_n)$, while the range of a negative attraction factor is 
$-f(A_n)\leq q(A_n) < 0$. By definition (\ref{3.6}), we have $0\leq f(A_n)\leq 1$, 
hence the average value of $f(A_n)$ is $1/2$. Applying the definition of arithmetic 
averages to $q(A_n)$ yields the averages (\ref{3.32}). $\square$ 

\vskip 2mm

The alternative with a better quality is more attractive. Therefore a larger 
attraction factor implies a more attractive alternative, although it is not 
necessarily more preferable. 

\vskip 2mm

{\bf Definition 8}. An alternative $A_1$ is more attractive than $A_2$ if and 
only if
\be
\label{3.33}
 q(A_1) > q(A_2) \;  .
\ee
Respectively, one can say that $A_2$ is more repulsive than $A_1$.

\vskip 2mm

{\bf Definition 9}. Two alternatives $A_1$ and $A_2$ are equally attractive if 
and only if
\be
\label{3.34}
 q(A_1) = q(A_2) \;  .
\ee

Thus, for each alternative $A_n$, there are two characteristics describing the 
alternative from two different points of view. The rational utility factor $f(A_n)$ 
describes the utility of the alternative telling us what the fraction (the probability) 
of decision makers selecting the alternative $A_n$ would be, if the choice were made 
solely on the basis of rational rules. The attraction factor $q(A_n)$, caused by the 
influence of emotional effects, shows how attractive (repulsive) the considered 
alternative is. When choosing between several alternatives, one compares the rational 
utility and emotional factors of different alternatives. An alternative can be more 
useful but less attractive, because of which it can become not preferable. The 
alternative $A_1$ is stochastically preferable to $A_2$, when $p(A_1) > p(A_2)$, 
which yields
$$
f(A_1) - f(A_2) > q(A_2) - q(A_1) \; .
$$  

When the set $\{A_n\}$ contains only two alternatives, it is possible to use the 
non-informative priors for attraction factors, estimating the related behavioural 
probabilities as
\be
\label{3.35}
p(A_n) = f(A_n) \pm 0.25 \, .
\ee

\subsection{Multiple attraction factors}

In the case where there are many alternatives in the set $\mathbb{A}$, it is also 
possible to estimate typical emotional corrections playing the role of non-informative 
priors for attraction factors. 
   
Suppose $N_A$ alternatives can be classified according to the level of their 
attractiveness, so that 
\be 
\label{3.36}
 q(A_n) > q(A_{n+1}) \qquad ( n = 1,2,\ldots, N_A-1) \;  .
\ee
Let the nearest to each other attraction factors $q(A_n)$ and $q(A_{n+1})$ be 
separated by a typical gap 
\be
\label{3.37}  
\Dlt \equiv  q(A_n) - q(A_{n+1}) \; .
\ee
Let us accept that the average, over the set $\mathbb{A}$, absolute value of the 
quality factor can be estimated by the non-informative prior $\overline{q} = 1/4$, 
in agreement with Theorem 2, so that
\be
\label{3.38}
 \overline q \equiv \frac{1}{N_A} \sum_{n=1}^{N_A} |\; q(A_n)\; | = \frac{1}{4} \; .
\ee

\vskip 2mm

{\bf Theorem 3}. {\it For a set $\mathbb{A}$ of $N_A$ alternatives, under conditions 
(\ref{3.36}) (\ref{3.37}), and (\ref{3.38}), the non-informative priors for the  
attraction factors are
\be
\nonumber
q(A_n) = \frac{N_A-2n+1}{2N_A} \qquad \qquad (N_A ~ even) \;  ,
\ee
\be
\label{3.39}
 q(A_n) = \frac{N_A(N_A-2n+1)}{2(N_A^2-1)} \qquad (N_A ~ odd) \;  ,
\ee
depending on whether $N_A$ is even or odd}.

\vskip 2mm
{\it Proof}. In accordance with conditions (\ref{3.36}), (\ref{3.37}), and 
(\ref{3.38}), we can write 
\be
\label{3.40}
q(A_n) = q(A_1) - (n-1)\Dlt \; .
\ee
From the alternation law (\ref{3.31}) it follows
\be
\label{3.41}
 q(A_1) = \frac{N_A-1}{2} \; \Dlt \;  .
\ee
Using the definition of the average $\overline{q}$ in (\ref{3.38}) we have the 
gap
\begin{eqnarray}
\label{3.42}
\Dlt = \left\{ \begin{array}{ll}
4\overline q/N_A , ~ & N_A ~ even \\
\\
4\overline q N_A/(N_A^2-1) , ~ & N_A ~ odd 
\end{array} \right.  \; ,
\end{eqnarray}
depending on whether the number of alternatives $N_A$ is even or odd. Then 
expression (\ref{3.41}) becomes
\begin{eqnarray}
\label{3.43}
q(A_1) = \left\{ \begin{array}{ll}
2\overline q(N_A-1)/N_A , ~ & N_A ~ even \\
\\
2\overline q N_A/(N_A+1) , ~ & N_A ~ odd 
\end{array} \right.  \; .
\end{eqnarray}
Using (\ref{3.40}), we get
\begin{eqnarray}
\label{3.44}
q(A_n) = \left\{ \begin{array}{ll}
2\overline q (N_A+1-2n)/N_A ,  & N_A ~ even \\
\\
2\overline q N_A(N_A+1-2n)/(N_A^2-1) , & N_A ~ odd 
\end{array} \right.  \; ,
\end{eqnarray}
Since according to (\ref{3.38}), $\overline q=1/4$, then expression (\ref{3.43}) 
leads to
\begin{eqnarray}
\label{3.45}
q(A_1) = \left\{ \begin{array}{ll}
(N_A-1)/2N_A , ~ & N_A ~ even \\
\\
N_A/2(N_A+1) , ~ & N_A ~ odd 
\end{array} \right.  \; .
\end{eqnarray}
Finally, equation (\ref{3.44}) results in the answer (\ref{3.39}). $\square$

\vskip 2mm

Thus, for a set of two alternatives, we have the already known values of the 
non-informative priors for quality factors
\be
\label{3.46}  
 \{ q(A_n) : ~ n = 1,2 \} = 
\left\{ \frac{1}{4}\;, \; - \; \frac{1}{4} \right\}\; .
\ee
For three alternatives, we find
\be
\label{3.47}
\{ q(A_n) : ~ n = 1,2,3 \} = 
\left\{ \frac{3}{8}\; , \; 0 \; , \; - \; \frac{3}{8} 
\right\}\; .
\ee
Respectively, for the set of four alternatives, we find the quality factors
\be
\label{3.48}
 \{ q(A_n) : ~ n = 1,2,3,4 \} = 
\left\{ \frac{3}{8}\; , \; \frac{1}{8}\; , \; - \; \frac{1}{8} \; , \; 
- \; \frac{3}{8} \right\} \; .
\ee
In the case of five alternatives, we obtain
\be
\label{3.49}
 \{ q(A_n) : ~ n = 1,2,3,4,5 \} = 
\left\{ \frac{5}{12}\; , \; \frac{5}{24}\; , \; 0 \; , \; - \; \frac{5}{24} \; , \; 
- \; \frac{5}{12} \right\} \; .
\ee
 
For large $N_A\gg 1$, the attraction factor varies between $1/2$ and $-1/2$ with the 
gap $1/N_A$.

\subsection{Problems in classifying attractiveness}

The typical magnitudes of attraction factors, describing to what extent the 
considered alternatives are attractive or repulsive, can be estimated by invoking 
the non-informative priors described above, provided the alternatives can be 
classified as more or less attractive. Generally, emotions can be classified as 
positive or negative \cite{Scherer_2009,Scherer_2019,Moors_2019}. Keeping this in 
mind, in many cases, the sign of the quality factor can be defined according to 
whether emotions are positive or negative. When an alternative $A_1$ is more 
attractive than $A_2$, then $q(A_1)>q(A_2)$. As a rule, an alternative looks 
more attractive, as compared to another one, when it seems to provide a better 
possibility for a gain, or is more in line with moral norms accepted in the 
society. For example, there exist general features that could be accepted, 
following Scherer and Moors \cite{Moors_2019}, for classifying emotions into 
positive (attractive) or negative (repulsive): These are the {\it uncertainty 
aversion} and the {\it propensity for cooperation}.  

\vskip 2mm
{\it Uncertainty aversion} implies that an undetermined or ambiguously determined 
alternative, missing precise information on its features, induces negative emotions 
resulting in a passive tendency and making the alternative repulsive. The 
uncertainty aversion is a fundamental and well-documented emotional characteristic 
of human beings. The typical human behaviour demonstrates the preference for known 
risks over unknown risks. As a consequence, for the majority of humans, uncertainty 
induces negative emotions, so that the attractiveness of an uncertain alternative 
is considered to be smaller than the attractiveness of a certain alternative. It 
is widely recognized that uncertainty frightens living beings, whether humans or 
animals. It is well documented that fear paralyzes, as in the cartoon where a rabbit 
stays immobile in front of an approaching boa, instead of running away. There are 
many circumstantial evidences that uncertainty frightens people as a boa frightens 
rabbits. Being afraid of uncertainty, a majority of human beings are hindered to act 
\cite{Moors_2019,Epstein_1999}.  

\vskip 2mm
{\it Propensity for cooperation} is one of the main evolutionary laws due to social 
selection, evidencing that the well-being of neighbors in the long run can be useful, 
because of which an unfair alternative, causing harm to somebody, produces negative 
emotions making the alternative repulsive. Modern theories of the evolution of human 
societies propose that humans' species are unique forms of cooperation derived from 
mutualistic collaboration with social selection against cheaters. In a first step, 
humans became obligate collaborative foragers such that individuals were interdependent 
with one another and so had a direct interest in the well-being of their partners. In 
this context, they evolved new skills and motivations for collaboration not possessed 
by other great apes, and they helped their potential partners. In a second step, these 
new collaborative skills and motivations were scaled up to group life in general, as 
humans faced competition from other groups. As part of  this new group-mindedness, 
they created cultural conventions, norms, and institutions, all characterized by 
collective intentionality, with knowledge of a specific set of these marking 
individuals as members of a particular cultural group. Human cognition and sociality 
thus became ever more collaborative and altruistic as human individuals became ever 
more interdependent \cite{Camerer_2003a,Tversky_2004,Tomasello_2009,Tomasello_2012,
Perc_2013,Perc_2017,Jusup_2021}.

In the case of two alternatives, $A_1$ and $A_2$, if one of them, say $A_1$, is 
found to be more attractive, so that $q(A_1)>q(A_2)$, then taking into account the 
alternation law $q(A_1)+q(A_2)=0$ and using the non-informative prior, we obtain 
$q(A_1)=0.25$ and $q(A_2)=- 0.25$. For the case of multiple alternatives, it is 
also possible to use the non-informative priors described in the previous section, 
provided the considered alternatives can be classified into more or less attractive.  

The classification of alternatives with respect to their attractiveness by humans 
is done on the basis of the available information. The attractiveness classification 
by an artificial intelligence also can be done on the basis of information that the 
intelligence possesses. If this information contains the main principles, such as the 
above examples of uncertainty aversion and propensity for cooperation, an artificial
intelligence will accordingly classify the given alternatives. 

When employing non-informative priors for attraction factors, it is important to 
keep in mind that we are talking about a typical or average decision maker. Of 
course, among the wide human population it is possible to find a variety of 
individuals with very different features, habits, moral and mental peculiarities. 
It seems to be evident that there exists no unique way of characterizing each of 
them. However we recall it again that the non-informative priors characterize the 
behaviour of a typical decision maker, that is the aggregate behaviour averaged over 
a large number of individuals. The predicted probabilities characterize the fraction 
of decision makers preferring the related alternatives. 
 
Thus, one way how an artificial intelligence could classify the attractiveness of 
alternatives is by loading into its memory the information on basic principles of 
behaviour. This way seems to be the sole available when the alternatives are 
described in words giving the qualitative picture not specified by numerical data.   

The other way could be tried for the cases where the alternatives are presented 
in numerical form, as it is done for describing the lotteries. Then it could be 
possible to suggest numerical recipes for estimating attraction factors. Below we 
shall consider the examples of both these approaches. 

In any case, we have to keep in mind that the notion of attractiveness is 
contextual. It can vary for different people, at different times, being dependent 
on circumstances and the particular set of alternatives. There is no unique 
global definition of attractiveness, but it is possible to prescribe the rules 
of evaluating the attraction factors for typical decision makers at some typical 
situations. Wether these rules are efficient or not is controlled by the comparison 
of theoretical predictions with empirical observations.

\subsection{Explicit attraction factors}

In some cases, where the alternatives are represented by lotteries, it is possible 
to directly verify the quarter law (\ref{3.32}) and to construct explicit expressions 
for attraction factors. Thus in Ref. \cite{Favre_130} the set of binary decision 
tasks consisting in the choice between two lotteries $L_n$, one risky,
$$
L_1 = \{\; x_1, \; 0 \; | \; x_2 , \; p \; | \; 0 , \; 1 - p \; \} \; , 
$$
with $p < 1$, and the other certain, 
$$
L_2 = \{\; x_1, \; 1 \; | \; x_2 , \; 0 \; | \; 0 , \; 0 \; \}
$$
is analyzed. The attraction factor $q(L_n) = p(L_n) - f(L_n)$ averaged over 
participants and lotteries is found to agree, within the accuracy caused by 
statistical errors, with the quarter law. 
 
A large pool of subjects making decisions with respect to a set of binary decision 
tasks \cite{Murphy_2018}, choosing between two lotteries $A$ and $B$, is analyzed 
in Refs. \cite{Vincent_131,Ferro_2021}. The attraction factor is modeled in the form
\be
\label{3.50}
 q(A) = \min \{ \vp(A) \; , \vp(B) \} \; \tanh\{ a[\; U(A) - U(B) \; ] \; ,
\ee
satisfying the alternation law (\ref{3.31})
$$
q(B) = - q(A) \; ,
$$
with $U(A)$ and $U(B)$ taken as the value functionals of cumulative prospect theory 
\cite{Kahneman_1979,Tversky_1992}, with
$$
\vp(A) = \frac{1}{Z} \; e^{\bt U(A)} \; , \qquad   
\vp(B) = \frac{1}{Z} \; e^{\bt U(B)} \; ,
$$
and the normalization constant
$$
 Z =  e^{\bt U(A)}  + e^{\bt U(B)}  \;  .
$$
The model parameters entering the attraction factor are fitted so that to optimally 
agree with the given experimental data set. It is shown that the decision theory 
with this attraction factor better describes the empirical data than the stochastic 
cumulative prospect theory \cite{Tversky_1992} and than the stochastic rank-dependent 
utility theory \cite{Quiggin_1982}.

Some authors \cite{Zhang_2021} include into the attraction factor the terms 
responsible for taking account of several subjective feelings, such as framing 
effect \cite{Tversky_1981}, stress caused by time pressure \cite{Diederich_2020}, 
and memory and aspiration effects \cite{Siegel_1957}. The approach with this 
multimodal attraction factor is shown to outperform the cumulative prospect 
theory \cite{Tversky_1992}, the method of random decision forests \cite{Ho_1998}, 
machine learning algorithm XGBoost \cite{Chen_2016}, and Deep Learning Feedforward 
neural networks prediction method \cite{Jain_1996}.

\subsection{Buridan's donkey problem}

It is necessary to keep in mind that, strictly speaking, the calibration of the 
parameters for the optimal characterization of one given set of lotteries may be 
not appropriate for another set of lotteries. In that sense, each family of 
parameters is contextual, being suitable for a particular collection of decision 
tasks, but not necessarily adequate for other types of decision problems. This 
fact can be easily understood, for instance, noticing that the attraction factor 
(\ref{3.50}) becomes zero, when the functionals $U(A)$ and $U(B)$ coincide, and 
the attraction factor becomes negligibly small when these functionals are close to 
each other. At the same time very close, or equal utilities can correspond to high 
uncertainty in the choice, producing large attraction factors. A typical example 
of this situation has been illustrated by Kahneman and Tversky \cite{Kahneman_1979} 
for a set of lotteries with close or coinciding expected utilities. The choice 
between two alternatives with equal or close utilities is a kind of the “Buridan’s 
donkey problem” \cite{Kane_2005} whose solution can be formalized by taking into 
account emotions \cite{Yukalov_125}.

The way how the Buridan's donkey problem can be resolved is based on the studies 
in experimental neuroscience, which have discovered that, when making a choice, 
the main and foremost attention of decision makers is directed towards the payoff 
probabilities \cite{Kim_2012}. This implies that subjects, when choosing, evaluate 
higher the probabilities than the related payoffs. In mathematical terms, this can 
be formulated as the existence of different types of scaling with respect to payoff 
utility $u(x_i)$ and payoff probability $p_n(x_i)$. For example, the payoff utility 
can be scaled linearly, while the payoff probability, exponentially. The exponential 
base can be defined as follows.

Suppose we compare two lotteries $L_1$ and $L_2$ defined as
$$
L_1 = \{u,p \; | \; 0, 1 - p \} \;  ,
$$
with a payoff utility $u$ and a related probability $p$, and the other lottery
$$
L_2 = 
\left\{\lbd u, \frac{p}{\lbd} \; | \; 0, 1 - \;\frac{p}{\lbd} \right\} \; ,
$$
whose payoff utility and probability are scaled in such a way that the expected 
utilities of the lotteries coincide, $U(L_1) = U(L_2) = up$. Hence the corresponding 
utility factors also coincide, $f(A_1) = f(A_2) = 1/2$, because of which, from the 
point of view of rational arguments, the lotteries are not distinguishable. This 
is a typical example of a series of lotteries considered by Kahneman and Tversky 
\cite{Kahneman_1979}. Nevertheless, empirical studies show that subjects do make 
clear preferences between the lotteries, depending on their payoffs and probabilities. 
This implies that intuitively subjects do classify the lotteries into attractive or 
repulsive \cite{Kahneman_1979}.  

Suppose the lottery $L_1$ is rather certain, which implies that the payoff 
probability is in the interval $1/2<p\leq 1$, with the average probability $p=3/4$. 
This value of probability, as follows from numerous empirical observations 
\cite{Hillson_2003,Hillson_2019}, is appreciated by people as highly certain. Let 
the scaling be such that $\lbd>1$, so that the payoff utility increases, while its 
probability diminishes. When $\lbd$ is not large, subjects do prefer the more 
certain lottery $L_1$. However strongly increasing the payoff utility attracts more 
people, despite a small payoff probability, which is confirmed by real-life lotteries 
\cite{Rabin_2000}. 

To classify the lotteries into attractive or repulsive quality classes, let us 
introduce the {\it lottery quality} $Q(L_n)$. The fact that decision makers in 
their choice pay the main and foremost attention to the payoff probabilities 
\cite{Kim_2012} is formalized by a linear dependence of the lottery quality with 
respect to the payoff utility and by an exponential dependence with respect to 
the payoff probability. For the case of the considered lotteries, this implies 
the lottery quality functionals
$$
Q(L_1) = u b^p \; , \qquad Q(L_2) = \lbd u b^{p/\lbd} \;  .
$$
When the scaling $\lambda$ is larger than unity, then $L_1$ is more certain, since 
$p$ is larger than $p/\lambda$, and subjects consider the more certain lottery as 
more attractive, which means that $Q(L_1)$ is larger than $Q(L_2)$. But if the payoff 
probability is diminished by an order, which assumes $\lbd=10$, while the payoff 
utility increases by an order, then the lottery $L_1$ becomes less attractive than 
$L_2$, so that $Q(L_1)$ becomes smaller than $Q(L_2)$. The change of attractiveness 
occurs where $Q(L_1)=Q(L_2)$. The latter equality gives the expression for the base 
$b$ that for $p=3/4$ and $\lbd=10$ yields
$$
 b = \lbd^{\lbd/(\lbd-1)p} =  30 \;  .
$$

Extending the definition of lottery quality to the general lottery defined in line 
with (\ref{3.12}), we come to the functional
\be
\label{3.51}
 Q(L_n) = \sum_i u(x_i) 30^{p_n(x_i)} \; .
\ee  
The form of $Q_n \equiv Q(L_n)$ shows that the lottery quality depends on both, 
payoff utilities as well as payoff probabilities, although this dependence is 
different due to the different attention of a decision maker with respect to utility 
and probability \cite{Kim_2012}.  

\vskip 2mm

{\bf Axiom 5}. The attractiveness of a lottery $L_n$ is quantified by the lottery 
quality (\ref{3.51}).

\vskip 2mm

The quality $Q_n$ shows which of the considered lotteries is more attractive. That 
lottery is more attractive for which the quality $Q_n$ is larger. Respectively, this 
defines the relation between the attraction factors. 

\vskip 2mm

{\bf Definition 10}. A lottery $L_1$ is more attractive than $L_2$, so that 
$q(L_1) > q(L_2)$, if the quality of the lottery $L_1$ is larger than that of $L_2$, 
that is when
\be
\label{3.52}
 Q_1 > Q_2 \; ,
\ee
where $Q_n \equiv Q(L_n)$ is the lottery quality (\ref{3.51}).

\vskip 2mm

When the choice is between two alternatives, and $Q_1>Q_2$, then the quality factors 
can be estimated by non-informative priors $q(L_1)= 1/4$ and $q(L_2)=-1/4$. Thus, if 
one makes a choice between two alternatives and one of them, say $A_1$, is more 
attractive than the other, say $A_2$, then the behavioural probabilities of selecting 
these alternatives can be estimated as
\be
\label{3.53}
 p(A_1) =f(A_1) + 0.25 \; , \qquad p(A_2) =f(A_2) - 0.25 \;  ,
\ee
where $f(A_n)$ are utility factors. Here the inequality $0\leq p(A_n)\leq 1$ is 
assumed, which can be formalized by the definition
\be
\nonumber
p(A_n) = {\rm Ret}_{[0,1]} \{ f(A_n) \pm  0.25 \} \; ,
\ee
where the retract function is defined as
\begin{eqnarray}
\nonumber
{\rm Ret}_{[0,1]}(z) = \left\{  \begin{array}{ll}
0 , ~ & ~ z < 0 \\
z , ~ & ~ 0 \leq z \leq 1 \\
1 , ~ & ~ z > 1 
\end{array} \right. .
\end{eqnarray}

It may happen that, considering the quality functionals of different lotteries, we 
meet the situation, where these functionals are equal, but the lotteries differ from 
each other by the gain-loss number difference
\be
\label{3.54}
N(A_n) = N_+(A_n) - N_-(A_n)
\ee 
between the number of admissible gains $N_+(A_n)$ and the number of possible losses 
$N_-(A_n)$. A typical example is the comparison of the lottery $A_1$ defined as 
$$
A_1 = \{ u , \; p \; | \; 0 , \; 1 - p \} 
$$
and the lottery
$$
A_3 = \{ u_1,p \; | \; u_2, p \; | \; 0 , 1 - 2p \} \;  ,
$$
in which $u_1 + u_2 = u$. Then the related quality functionals are equal
$$
 Q(A_3) = u_1 b^p + u_2 b^p = u b^p = Q(A_1)  \; ,
$$
where $b=30$. If $u_n>0$, then the lottery $A_1$ possesses only one admissible 
gain and no losses, while the lottery $A_3$, two gains and also no losses. Hence 
$N(A_1)=1$ and $N(A_3)=2$. Since $N(A_3)$ is larger than $N(A_1)$, the lottery 
$A_3$ is treated as more attractive. Similarly, for the lotteries with losses, 
where $u_n<0$, and quality functionals are equal, the gain-loss number difference 
$N(A_1)=-1$ is larger than $N(A_3)=-2$, so that the lottery $A_1$ with a smaller 
number of losses is more attractive.  

\vskip 2mm 
{\bf Definition 11}. If the lottery qualities are equal, with $Q(A_1)=Q(A_2)$, then 
a lottery $A_1$ is more attractive, than $A_2$, so that $q(A_1)>q(A_2)$, if 
\be
\label{3.55}
N(A_1) > N(A_2) \; .
\ee

The lottery $L_n$ is said to be in the positive quality class, when its attraction 
factor is positive, $q(L_n) > 0$, and the lottery is in the negative quality class, 
if the related attraction factor is negative, $q(L_n) < 0$. When some lotteries $A_1$ 
and $A_2$ cannot be classified as more or less attractive, they are said to be of 
equal quality, or equally attractive, so that $q(A_1)=q(A_2)$. If there are only 
two of these lotteries, then the alternation law $q(A_1)+q(A_2)=0$ implies 
$q(A_1)=q(A_2)=0$. In that case, the lotteries are said to be in the neutral quality 
class.  

\vskip 2mm
In applications, the probabilities are to be interpreted in the standard frequentist 
way as fractions of decision makers choosing the corresponding alternatives. So the 
utility factor $f(A_n)$ shows the fraction (frequentist probability) of decision 
makers that would choose the corresponding alternative on the basis of only rational 
rules. However, the behavioural probability $p(A_n)$ defines the real fraction of 
decision makers actually choosing $A_n$, taking into account both the rational utility 
as well as the emotional attractiveness of the alternatives.

\subsection{Kahneman-Tversky lotteries}

Considering a number of examples, illustrating the Buridan's donkey problem, 
Kahneman and Tversky \cite{Kahneman_1979} have showed that when the utility of 
lotteries are close to each other, decision makers do not obey the expected utility 
theory and often make choices contradicting this theory. Nevertheless, it is possible 
to demonstrate that taking into account emotions according to the scheme of the 
previous subsection resolves contradictions and allows one to make clear choices 
based on the influence of emotions. 

The method described above correctly predicts the aggregate choice, giving good 
quantitative estimates for behavioural probabilities. The estimates can be made 
without involving fitting parameters. Utility factors are defined according to the 
Luce rule described in Sec. 3.4. For simplicity, the linear utility function $u(x)=x$ 
is accepted. The payoffs are measured in monetary units, whose measures are of no 
importance when using dimensionless utility factors. The utility factor is calculated 
by the formula
$$
f(L_n) = \frac{U(L_n)}{\sum_n U(L_n)} \; , \qquad  U(L_n) \geq 0 \; ,
$$ 
for semi-positive expected utilities and as
$$
f(L_n) = \frac{|\;U(L_n)\;|^{-1}}{\sum_n|\; U(L_n)\;|^{-1}} \; , \qquad 
 U(L_n) < 0 \; ,
$$
for negative expected utilities, where the expected utilities are given by the 
expression
$$    
U(L_n) = \sum_i x_i p_n(x_i) \; .
$$
The quality functional $Q_n \equiv Q(L_n)$ is defined in (\ref{3.51}). The attraction 
factors are given by their non-informative priors, with the sign prescribed by the 
lottery quality functionals. For brevity, the notation $Q(A_n) \equiv Q_n$ is used.   

The typical situation in the considered Buridan's donkey problem is as follows. One 
needs to choose between the very close lotteries, for instance 
$$
L_1 = \{ 2.5, \; 0.33 \; | \; 2.4, \; 0.66 \; | \; 0, \; 0.01 \} \; , \qquad
L_2 = \{2.4 , \; 1 \} \; ,
$$
with the almost coinciding utility factors  $f(L_1)=0.501$ and $f(L_2)=0.499$. 
Formally, the first lottery should be chosen on the rational grounds, since 
$f(L_1)>f(L_2)$. However, the lottery quality functionals $Q_1=30.3$ and $Q_2=72$ 
show that the second lottery, being more certain, is more attractive, because 
$Q_2>Q_1$. This implies that $q(L_2)>q(L_1)$, which, involving the non-informative 
prior, yields $q(L_1) = - 0.25$, while $q(L_2) = 0.25$. This gives the behavioural 
probabilities
$$
p(L_1) = 0.25 \; , \qquad p(L_2) = 0.75 \; ,
$$
according to which the second lottery is preferable. This is in agreement with the 
empirical results
$$
p_{exp}(L_1) = 0.18 \; , \qquad p_{exp}(L_2) = 0.82 \; .
$$
The more certain, but less useful lottery is chosen. The typical statistical error 
in the experiments was close to $\pm 0.1$.

The total of $18$ choices between the pairs of lotteries with close or exactly equal 
utility factors were studied. The detailed description of calculations is presented 
in \cite{Yukalov_125}. Here we summarize the results in Table 1 showing the optimal 
lottery having the largest predicted behavioural probability
\be
\label{3.56}
 p(L_{opt}) \equiv \max_n p(L_n)  
\ee
over the given lattice of alternatives, the utility factor $f(L_{opt})$ of the 
optimal lottery, experimental probabilities of the optimal lottery $p_{exp}(L_{opt})$, 
defined as the fractions of decision makers choosing the optimal lottery, and the 
related empirical attraction factors 
\be
\label{3.57}
q_{exp}(L_{opt}) = p_{exp}(L_{opt}) - f(L_{opt}) \; .
\ee
The data, corresponding to the non-optimal lotteries, can be easily found from the 
normalization conditions
\be
\label{3.58}
p(L_1) + p(L_2) = 1 \; , \qquad f(L_1) + f(L_2) = 1 \; , \qquad  
q(L_1) + q(L_2) = 0 \;  .
\ee
At the bottom of Table 1, the average values over all $18$ cases are given for 
the utility factors $\overline f(L_{opt})=0.5$, predicted behavioural probability
$\overline p(L_{opt})=0.75$, experimentally observed probability 
$\overline p_{exp}(L_{opt})=0.77$, and the experimentally observed average absolute 
value of the attraction factor
\be
\label{3.59} 
\overline q_{exp} = \overline p_{exp}(L_{opt}) - \overline f(L_{opt}) = 0.27 \;  .
\ee
Within the accuracy of the experiment, the predicted average behavioural probability 
of choosing an optimal lottery, $0.75$, equals the empirical average fraction of 
decision makers $0.77$, and the average attraction factor $0.27$ practically coincides 
with the theoretical estimate of $0.25$.

\begin{table}[hp]
\caption{\small Optimal lotteries $L_{opt}$ from the Kahneman-Tversky set of binary 
choices, the utility factors  $f(L_{opt})$ of the optimal lotteries, predicted 
behavioural probabilities $p(L_{opt})$, experimentally observed probabilities 
$p_{exp}(L_{opt})$, defined by the fractions of the participants choosing the optimal 
lottery $L_{opt}$, and the empirical attraction factors $q_{exp}(L_{opt})$ corresponding 
to the optimal lotteries. The bottom line shows the related average values.}
\vskip 2mm
\centering
\renewcommand{\arraystretch}{1.2}
\begin{tabular}{|c|c|c|c|c|c|} \hline
   & $L_{opt}$ & $f(L_{opt})$  & $p(L_{opt})$ & $p_{exp}(L_{opt})$ &  $q_{exp}(L_{opt})$ \\ \hline
1  &   $L_2$   & 0.50  & 0.75 & 0.82   & 0.32 \\ 
2  &   $L_1$   & 0.50  & 0.75 & 0.83   & 0.33 \\ 
3  &   $L_2$   & 0.48  & 0.73 & 0.80   & 0.32 \\ 
4  &   $L_1$   & 0.52  & 0.77 & 0.65   & 0.13 \\
5  &   $L_2$   & 0.50  & 0.75 & 0.86   & 0.36  \\ 
6  &   $L_1$   & 0.50  & 0.75 & 0.73   & 0.23 \\
7  &   $L_2$   & 0.50  & 0.75 & 0.82   & 0.32 \\
8  &   $L_1$   & 0.50  & 0.75 & 0.72   & 0.22 \\ 
9  &   $L_2$   & 0.50  & 0.75 & 0.84   & 0.34 \\ 
10 &   $L_2$   & 0.50  & 0.75 & 0.80   & 0.30 \\ 
11 &   $L_1$   & 0.48  & 0.73 & 0.92   & 0.44 \\
12 &   $L_2$   & 0.52  & 0.77 & 0.58   & 0.06 \\
13 &   $L_2$   & 0.50  & 0.75 & 0.92   & 0.42 \\
14 &   $L_1$   & 0.50  & 0.75 & 0.70   & 0.20 \\ 
15 &   $L_1$   & 0.50  & 0.75 & 0.69   & 0.19 \\
16 &   $L_1$   & 0.50  & 0.75 & 0.70   & 0.20 \\
17 &   $L_2$   & 0.50  & 0.75 & 0.83   & 0.33 \\
18 &   $L_1$   & 0.50  & 0.75 & 0.69   & 0.19 \\  \hline
   &           & 0.50  & 0.75 & 0.77   & 0.27 \\ \hline
\end{tabular}
\end{table}

As the analysis of this set of Buridan's donkey problems demonstrates, it is not 
possible to predict the behavioural decision making of humans by considering 
separately either lottery utilities, lottery payoffs, or payoff probabilities. But 
reliable predictions can be made by defining behavioural probabilities, including 
the estimates of both, rational utility factors as well as attraction factors 
characterizing emotions. On the aggregate level, these predictions provide good 
quantitative agreement with empirical data, involving no fitting parameters.      

At the same time, the expected utility theory is not applicable to the Kahneman-Tversky 
lotteries, since the lottery with a higher utility is preferred only twice among $18$ 
lotteries. It is important to stress that the formula (\ref{3.50}), with the expected 
utilities $U(L_n)$ is not valid here, since for the coinciding utilities it gives zero 
attraction factor, while the aggregate experimental data give for the attraction factor 
$0.27$.

\subsection{Verification of quarter law}

As is clear from the analysis of the Kahneman-Tversky lotteries illustrating the 
Buridan's donkey problem, the magnitude of the attraction factor is well described by 
the non-informative prior $\overline{q}=0.25$, called quarter law. Close estimates for 
the attraction factor have been found for experimental observations of Ref. \cite{Favre_130}. 

In order to show that the quarter law is not typical merely for the pairs of lotteries 
with close or coinciding utility factors, but has a larger range of applicability, it 
is possible to verify this law on the basis of a large set of binary lotteries studied 
recently \cite{Murphy_2018}. In the studied laboratory experiments, a group of $142$ 
subjects were suggested a set of binary decision tasks (lotteries), with repeating 
the same study after two weeks, after randomly changing the order of the pairs of 
lotteries. The experiments at these two different sessions are referred to as session 
$1$ and session $2$. There have been considered three types of lotteries: lotteries 
containing only gains, with all payoffs being positive, lotteries with only losses, 
with all payoffs being negative, and mixed lotteries containing gains as well as 
losses. A loss implies the necessity to pay the designed amount of money. The 
evaluation of attraction factors is accomplished in the positive and negative quality 
classes of the lotteries $L_+$ and $L_-$, where the attraction factors of the related 
lotteries are positive, $q(L_+)>0$, or negative, $q(L_-)<0$, respectively. The neutral 
quality class, has to be excluded. The latter is defined as the set of lotteries, where 
the difference between the utility factors and the empirical choice probabilities, in 
both sessions, is smaller than the value of the typical statistical error of $0.1$ 
corresponding to random noise. 

Table 2 presents the results for the optimal lotteries with only gains and Table 3 
shows the results for the optimal lotteries with only losses. Recall that a lottery 
$L_1$ is called optimal, as compared to a lottery $L_2$ if and only if the corresponding 
probability $p(L_1)$ is larger than $p(L_2)$. In both the cases, of either the lotteries 
with only gains or the lotteries with only losses, an optimal lottery is always a lottery
from the positive quality class, in which $q(L_{opt})>0$. The situation can be different
for the mixed lotteries, containing gains as well as losses. In these cases, an optimal 
lottery can occasionally pertain to a negative quality class. Table 4 summarizes the 
results for the mixed lotteries containing both gains and losses. Among these lotteries,
the first sixteen examples in Table 4 are the lotteries from the positive quality class, 
which at the same time are the optimal lotteries. The last five cases are the lotteries
that are not optimal, however being from the positive quality class. 

\begin{table}[hp]
\caption{\small Optimal lotteries with only gains. The utility factors $f(L_{opt})$ 
of the optimal lotteries, fractions of subjects $p_i(L_{opt})$ choosing the optimal 
lottery in the session $i=1,2$, and the attraction factors $q_i(L_{opt})$ of the 
optimal lotteries in the session $i=1,2$. The bottom line shows the average values 
of the related quantities.}
\vskip 2mm
\centering
\renewcommand{\arraystretch}{1.2}
\begin{tabular}{|c|c|c|c|c|c|} \hline
   & $f(L_{opt})$ & $p_1(L_{opt})$  & $p_2(L_{opt})$ & $q_1(L_{opt})$ & $q_2(L_{opt})$ \\ \hline
1  &   0.55       & 0.86 & 0.89 & 0.31   & 0.34 \\ 
2  &   0.48       & 0.66 & 0.69 & 0.18   & 0.21 \\ 
3  &   0.51       & 0.68 & 0.62 & 0.17   & 0.11 \\ 
4  &   0.59       & 0.80 & 0.75 & 0.22   & 0.17 \\
5  &   0.63       & 0.89 & 0.90 & 0.26   & 0.27  \\ 
6  &   0.66       & 0.96 & 0.95 & 0.30   & 0.29 \\
7  &   0.51       & 0.79 & 0.81 & 0.28   & 0.30 \\
8  &   0.48       & 0.60 & 0.63 & 0.12   & 0.15 \\ 
9  &   0.63       & 0.88 & 0.92 & 0.26   & 0.30 \\ 
10 &   0.56       & 0.89 & 0.82 & 0.33   & 0.26 \\ 
11 &   0.63       & 0.77 & 0.73 & 0.14   & 0.10 \\
12 &   0.51       & 0.72 & 0.73 & 0.21   & 0.21 \\
13 &   0.61       & 0.87 & 0.85 & 0.26   & 0.24 \\
14 &   0.63       & 0.93 & 0.93 & 0.30   & 0.30 \\ 
15 &   0.64       & 0.85 & 0.87 & 0.21   & 0.23 \\
16 &   0.64       & 0.80 & 0.80 & 0.16   & 0.16 \\
17 &   0.64       & 0.89 & 0.89 & 0.25   & 0.25 \\
18 &   0.48       & 0.65 & 0.70 & 0.17   & 0.22 \\  
19 &   0.65       & 0.87 & 0.93 & 0.22   & 0.28 \\
20 &   0.66       & 0.86 & 0.82 & 0.20   & 0.16 \\
21 &   0.58       & 0.84 & 0.80 & 0.26   & 0.22 \\
22 &   0.52       & 0.75 & 0.74 & 0.23   & 0.22 \\
23 &   0.48       & 0.64 & 0.65 & 0.16   & 0.17 \\
24 &   0.44       & 0.60 & 0.53 & 0.16   & 0.10 \\
25 &   0.62       & 0.73 & 0.79 & 0.11   & 0.17 \\
26 &   0.64       & 0.81 & 0.90 & 0.17   & 0.26 \\
27 &   0.66       & 0.93 & 0.96 & 0.27   & 0.30 \\ \hline
   &   0.58       & 0.80 & 0.80 & 0.22   & 0.22 \\ \hline
\end{tabular}
\end{table}

\begin{table}[hp]
\caption{\small Optimal lotteries with only losses. The utility factors 
$f(L_{opt})$ of the optimal lotteries, fractions of subjects (frequentist 
probabilities) $p_i(L_{opt})$ choosing the optimal lottery in the sessions $i=1,2$, 
and the attraction factors $q_i(L_{opt})$ of the optimal lotteries in the session 
$i=1,2$. The bottom line shows the average values of the related quantities.}
\vskip 2mm
\centering
\renewcommand{\arraystretch}{1.2}
\begin{tabular}{|c|c|c|c|c|c|} \hline
   & $f(L_{opt})$ & $p_1(L_{opt})$  & $p_2(L_{opt})$ & $q_1(L_{opt})$ & $q_2(L_{opt})$ \\ \hline
1  &   0.52       & 0.77 & 0.75 & 0.25   & 0.23 \\ 
2  &   0.60       & 0.85 & 0.83 & 0.25   & 0.23 \\ 
3  &   0.53       & 0.72 & 0.71 & 0.19   & 0.18 \\ 
4  &   0.64       & 0.96 & 0.92 & 0.32   & 0.28 \\
5  &   0.55       & 0.70 & 0.68 & 0.15   & 0.13  \\ 
6  &   0.54       & 0.73 & 0.72 & 0.20   & 0.19 \\
7  &   0.63       & 0.79 & 0.84 & 0.16   & 0.21 \\
8  &   0.54       & 0.66 & 0.63 & 0.12   & 0.09 \\ 
9  &   0.56       & 0.80 & 0.89 & 0.24   & 0.33 \\ 
10 &   0.58       & 0.89 & 0.92 & 0.31   & 0.34 \\ 
11 &   0.49       & 0.66 & 0.71 & 0.17   & 0.22 \\
12 &   0.62       & 0.87 & 0.93 & 0.25   & 0.31 \\
13 &   0.55       & 0.79 & 0.74 & 0.24   & 0.19 \\
14 &   0.54       & 0.82 & 0.77 & 0.29   & 0.24 \\ 
15 &   0.53       & 0.65 & 0.70 & 0.12   & 0.17 \\
16 &   0.51       & 0.59 & 0.62 & 0.08   & 0.11 \\
17 &   0.56       & 0.79 & 0.86 & 0.23   & 0.30 \\
18 &   0.58       & 0.89 & 0.90 & 0.31   & 0.32 \\  
19 &   0.61       & 0.76 & 0.74 & 0.15   & 0.13 \\ \hline
   &   0.56       & 0.77 & 0.78 & 0.21   & 0.22 \\ \hline
\end{tabular}
\end{table}

\begin{table}[hp]
\caption{\small Mixed lotteries, containing both gains and losses, and pertaining 
to the positive quality class, where the attraction factors are positive. The 
utility factors $f(L_+)$, fractions of subjects $p_i(L_+)$ choosing the corresponding 
lotteries in the session $i=1,2$, and the attraction factors $q_i(L_+)$ of the 
lottery in that sessions $i=1,2$. The bottom lone shows the average values of 
the related quantities.}
\vskip 2mm
\centering
\renewcommand{\arraystretch}{1.2}
\begin{tabular}{|c|c|c|c|c|c|} \hline
   & $f(L_+)$     & $p_1(L_+)$  & $p_2(L_+)$ & $q_1(L_+)$ & $q_2(L_+)$ \\ \hline
1  &   0.40       & 0.69 & 0.66 & 0.29   & 0.26 \\ 
2  &   0.62       & 0.85 & 0.85 & 0.23   & 0.23 \\ 
3  &   0.67       & 0.87 & 0.82 & 0.20   & 0.15 \\ 
4  &   0.44       & 0.62 & 0.61 & 0.18   & 0.17 \\
5  &   0.50       & 0.64 & 0.54 & 0.15   & 0.05  \\ 
6  &   0.59       & 0.71 & 0.65 & 0.12   & 0.06 \\
7  &   0.54       & 0.69 & 0.63 & 0.16   & 0.10 \\
8  &   0.49       & 0.66 & 0.60 & 0.18   & 0.16 \\ 
9  &   0.57       & 0.87 & 0.85 & 0.30   & 0.28 \\ 
10 &   0.65       & 0.75 & 0.77 & 0.10   & 0.12 \\ 
11 &   0.52       & 0.77 & 0.70 & 0.26   & 0.19 \\
12 &   0.49       & 0.58 & 0.63 & 0.09   & 0.14 \\
13 &   0.55       & 0.87 & 0.92 & 0.32   & 0.37 \\
14 &   0.52       & 0.61 & 0.67 & 0.09   & 0.15 \\ 
15 &   0.53       & 0.80 & 0.83 & 0.27   & 0.30 \\
16 &   0.56       & 0.67 & 0.63 & 0.11   & 0.07 \\
17 &   0.00       & 0.27 & 0.27 & 0.27   & 0.27 \\
18 &   0.00       & 0.29 & 0.36 & 0.29   & 0.36 \\  
19 &   0.00       & 0.30 & 0.45 & 0.30   & 0.45 \\
20 &   0.00       & 0.39 & 0.38 & 0.39   & 0.38 \\
21 &   0.00       & 0.37 & 0.35 & 0.37   & 0.35 \\ \hline
   &   0.41       & 0.63 & 0.63 & 0.22   & 0.22 \\ \hline
\end{tabular}
\end{table}

As follows from these tables, the value of the attraction factor in the positive 
or negative quality classes is $\pm 0.22$, respectively, which is in very good 
agreement with the predicted non-informative priors $\pm 0.25$. Therefore the 
quarter law provides a rather accurate estimate of the attraction factor at the 
aggregate level.

\subsection{Contextuality of attraction factors}

Analyzing separate cases of particular games or lotteries, it becomes evident 
that the values of attraction factors can rather strongly vary. They, of course, 
also vary for different decision makers and even can vary for the same person 
at different times. In that sense, attraction factors are contextual and random. 
From the examples considered above, it is easy to separate some groups of 
lotteries that demonstrate the occurrence of the attraction factors essentially 
differing from the quarter law. In the broad sense, attraction factors are 
contextual, depending on particular subjects, concrete problems, types of 
lotteries, and even time. It is necessary to keep in mind that the modeling 
of attraction factors by explicit expressions or methods has limits of their 
application. For example, characterizing the attraction factors for the lotteries 
with close utility factors, by calculating the lottery quality, which works so 
well for the Buridan's donkey problem, might be not so good for the lotteries 
with strongly differing utility factors. The age and gender of decision makers 
can also influence the values of the attraction factors \cite{Favre_130}. Such 
variations reflect the contextual and random nature of emotions, hence their 
interference defining the attraction factor. 

An attraction factor for a particular decision problem and for a particular person 
can significantly deviate from the quarter law. However it is necessary to remember 
that this law is the signature of typical, that is average, behaviour of decision 
makers subject to emotions. Therefore to check the validity of the quarter law, one 
has to average the behavioral data for a number of subjects and decision tasks. It 
is possible to define the typical attraction factor as the average over a 
distribution, for instance over the beta distribution \cite{Yukalov_124}.
  
One more, and a very important, way of evidencing the validity of the quarter law, 
hence of the feasibility to use non-informative priors for demonstrating the role 
of emotions on the aggregate level, is the application of this law for explaining 
the main paradoxes in behavioural decision making, which is demonstrated below in 
Sec. 4.

\subsection{Choice between bundled alternatives}

In some cases, one has to make a choice not between separate alternatives but 
between so-called bundled alternatives, when each alternative comprises several 
objects or possibilities. In other words, each alternative is a package embodying 
several parts. Often these parts are heterogeneous and require multiattribute 
choice. A very widespread example is the choice a customer has to make when 
deciding between different packages each containing more than one goods or 
services. The attribute-based approach to study customer choices cannot deal 
with bundles of heterogeneous components, which are usually drawn from different 
product categories. Usually, the choice between bundled alternatives is emotionally 
difficult \cite{Cohen_1983,Payne_1993}.

A choice between two packages of services can be an example. Say, one package 
includes standard telephone service, television programming, and cellular telephone 
service. The other package suggests standard telephone service, heating and cooling 
maintenance service, and electricity and gas service. The other example is the 
package of hamburger, eggs and bacon, or a bundle consisting of hamburger, chips 
and cola. The choice between bundled alternatives can be clearly formalized for 
the case of lotteries or alternatives that can be quantified by utility. When one 
chooses which package to buy, one can compare the package prices. 

Let us consider a bundle 
\be
\label{3.60}
 \mathcal{B} = \{ \mathbb{A}_j : ~ j = 1,2,\ldots N_\mathcal{B} \} \;  ,
\ee
whose elements are the sets of alternatives
\be
\label{3.61}
  \mathbb{A}_j = \{ A_{jn} : ~ n = 1,2,\ldots N_A \} \;  .
\ee
If each alternative $A_{jn}$ can be quantified by a value, say, by the expected 
utility $U(A_{jn})$, then we can define the utility factors $f(A_{jn})$ and the 
attraction factors $q(A_{jn})$ giving the behavioral probability
\be
\label{3.62}
p(A_{jn}) = f(A_{jn}) + q(A_{jn}) \;   .
\ee
Then the bundle (\ref{3.60}) induces a superlottery 
$$
 \{ U(A_{jn}), \; p(A_{jn}) : ~ j = 1,2,\ldots N_\mathcal{B} \} \;  ,
$$
for which one can define the expected utility
\be
\label{3.63}
  U(\mathbb{A}_j) = \sum_{n=1}^{N_A}  U(A_{jn})\; p(A_{jn})
\ee
and the corresponding behavioural probability 
\be
\label{3.64}
 p(\mathbb{A}_j)  = f(\mathbb{A}_j) + q(\mathbb{A}_j) \; .
\ee
The package, whose probability is the largest, is stochastically optimal,
\be
\label{3.65}
  p(\mathbb{A}_{opt})  = \max_j  p(\mathbb{A}_j) \;  .
\ee

\subsection{Quantum versus classical consciousness}

The material of the present Sec. 3 and the following Sec. 4 show that taking into 
account emotions does not necessarily need the use of quantum techniques. Although 
some structures and notions borrow the ideas of the theory of quantum measurements 
under intrinsic noise, but affective decision theory can be formulated as a 
self-contained theory without any resort to quantum formulas. This, e.g., concerns 
the main notion of behavioural probability consisting of two terms, utility factor 
and attraction factor. The latter in the theory of quantum measurements under 
intrinsic noise appears due to the interference of noise modes, which can be 
associated with the interference of emotions. However, there is no need to repeat 
each time its derivation employing the techniques of quantum theory. It is 
sufficient to postulate the structure of behavioural probability as of a two-term 
quantity. Yes, many structural features of the behavioural probability, hence of 
the affective decision making, are analogous to the features of quantum theory. 
However analogy does not mean identity. In this respect, there emerges the 
principal question: What is the nature of consciousness and the brain, whether 
it is quantum or classical? There can happen three possibilities.

\vskip 2mm
(i) Both the brain and consciousness are quantum.

\vskip 2mm
(ii) The brain is classical, but consciousness is quantum. 

\vskip 2mm
(iii) Both the brain and consciousness are classical. 

\vskip 2mm
The case of a quantum brain, but classical consciousness evidently has to be rejected
at once, as far as a quantum object cannot be characterized in classical terms. The 
case (i) is rather suspicious, since at the present time there are no empirical 
indications on the brain quantum nature, and its functioning is well described in the 
frame of classical language. Of course, on the microscopic level, there exist numerous
quantum effects with molecules and atoms composing the brain, as is reviewed in Ref. 
\cite{Jedlicka_2017}. There is nothing unusual in that, as far as any macroscopic object 
is composed of atoms and molecules that need for their description quantum theory. However, 
from this it does not follow that the macroscopic object as a whole requires for its 
description quantum techniques. A truck is made of atoms and molecules that are quantum 
particles. But none would insist that the truck itself needs quantum theory for describing 
its work.    

The widely spread claim is that, well, the brain is classical, but consciousness is
quantum or quantum-like in the sense that it can be described only using quantum theory.
This claim, however, confronts clear logical contradictions. It is obvious that a 
classical object and its functioning do not need for their description quantum 
terminology, but have to allow for a classical language of their functioning. Suppose 
the brain is classical, although consciousness, being the brain basic feature, cannot 
be characterized in classical terms, but necessarily requires the use of quantum theory. 
Then this implies that the brain has to be quantum, which contradicts the assumption 
that it is classical. It is logically impossible that a classical object would not 
allow for a classical description, but would necessarily require quantum representation.

If we have to accept that consciousness is classical, implying that it has to be described
in classical language, then why do we need to invoke quantum theory for its description?            
The answer is very simple: For the description of consciousness, we do not need quantum 
theory, since consciousness, being a classical feature of the classical brain, can be 
described in classical terms. Plunging into complicated formulas of quantum theory, 
instead of applying much simpler classical language, is merely a matter of fashion. 

Here it is possible to remark that the statement that a classical system can be 
described in classical terms does not contradict to the fact that a classical system
could be described as well in quantum language. The point is that among several 
admissible descriptions one usually chooses the simplest. 

Quantum theory can be useful in the process of formulating the theory of consciousness 
operation, providing many ideas on the structure of the feasible approach. It is 
therefore useful to study the theory of quantum measurements under intrinsic noise
for understanding the similarities of the quantum measurements with decision making and
for catching hints of the possible mathematical representation of consciousness 
functioning. In that sense, quantum operation of the brain or of an artificial 
intelligence implies the adjusting of some formal mathematical analogies, borrowed from
quantum theory, for their use in the elaboration of decision making, after translating
these ideas into classical language.            

Quantum notions can serve as a powerful inspiring tool for understanding the functioning
of cognition and for creating practical algorithms for artificial intelligence, the 
algorithms that, anyway, can be formulated in classical terms.

\section{Resolution of behavioural paradoxes}  

There can occur three types of paradoxes plaguing the classical utility theory. 
In one case, when emotions do not play important role, to avoid the paradox, it 
is sufficient to resort to the general form of the utility factor derived in Sec. 
$3.3$, without involving attraction factors. The other situation is when the 
paradox cannot be resolved without taking into account emotions and the related 
attraction factors, and the considered problem can be formulated in explicit 
mathematical terms involving a comparison of lotteries. In that case, it is 
straightforward to suggest concrete rules allowing for the classification of 
attraction factors, describing emotions, into positive or negative, as has been 
done above for the Buridan's donkey problem. The third situation happens, when 
the attractiveness of alternatives is based on several fuzzy feelings that are 
more difficult, although not impossible, to formalize, and, being taken into 
account, emotions allow for rather good quantitative resolution of paradoxes. 
Below all these types of paradoxes are illustrated by particular examples.

\subsection{St. Petersburg paradox}  

The St. Petersburg paradox is, probably, the oldest paradox in decision theory, 
and can be said to have promoted the birth of modern decision theory itself. It 
was invented by Nicolas Bernoulli in $1713$ and discussed in his private letters. 
The paradox was formally stated by his cousin Bernoulli \cite{Bernoulli_1738}, 
who worked in St. Petersburg. The paradox was published in the Proceedings of 
the Imperial Academy of Sciences of St. Petersburg \cite{Bernoulli_1738}. The 
history and attempts of the paradox solution have been recently reviewed 
in \cite{Martin_2008,Yukalov_132}.

In order to understand the paradox, it is necessary to distinguish between the 
Bernoulli game and the paradox as such. Originally, the game is formulated as follows. 
Paying a fixed fee to enter the game, a player tosses a fair coin repeatedly until 
tails first appear, ending the game. If tails appear for the first time at the $n$-th 
toss, the player wins $x_n=2^n$ monetary units. Thus, the player wins $2$ monetary 
units if tails appear on the first toss, $2^2$ units, if on the second, $2^3$ units, 
if on the third, and so on. The gain of $2^n$ units occurs therefore with the 
probability $1/2^n$. The cost of the ticket to enter the game is proportional to 
the number of allowed tosses. The main question is: How much should a player pay for 
the ticket to enter the game? In other words: How many tosses should a player buy? 

In mathematical terms, the game is defined as follows 
\cite{Samuelson_1960,Samuelson_1977}. There is a set of lotteries 
$\{L_n:\; n=1,2,\ldots,\infty\}$ labeled according to the number of allowed tosses, 
so that the first lottery is
$$
 L_1 = \left\{ 2 , \; \frac{1}{2} \; | \; 0 , \; \frac{1}{2} \right\} \; ,
$$
the second is
$$
 L_2 = \left\{ 2 , \; \frac{1}{2} \; | \; 2^2 , \; \frac{1}{2^2}\; | \; 
0 , \; \frac{1}{2^2} \right\} \;  ,
$$
and so on, with the $n$-th lottery
$$
 L_n = 
\left\{ 2 , \; \frac{1}{2} \; | \; 2^2 , \; \frac{1}{2^2}\; | \; \ldots \; | \;
2^n , \; \frac{1}{2^n} \; | \; 0 , \; \frac{1}{2^n} \right\} \; .
$$
The lottery $L_\infty$ implies the limit of the series $\{L_n\}$ for $n\ra\infty$.

Then the main question ``for how many tosses the player should buy a ticket?" is 
equivalent to the question ``which lottery has to be treated as optimal in the sense 
of being preferred by the players?" This game is a well posed mathematical problem 
and does not contain any paradox in its formulation.

The paradox appears in the attempt to suggest a normative mathematically justified 
way of choosing an optimal lottery from the given set of lotteries in the Bernoulli 
game, so that the prescribed choice would be in agreement with the choice of real 
players. In the original setup, one compares the lottery expected utilities. The 
lottery enjoying the largest expected utility is nominated optimal. One considers 
a linear utility function $u(x) = x$ of payoffs $x$. 

Using the standard rules of expected utility theory \cite{Neumann_1953,Savage_1954}, 
one comes to the conclusion that the expected utility of the $n$-th lottery is 
$$
U(L_n) = \sum_{m=1}^n u\left( 2^m\right) \; \frac{1}{2^m} = n \; .
$$
Therefore the largest expected utility corresponds to the lottery with the infinite 
number of tosses:
$$
\max_n U(L_n) \ra \infty \;  .
$$
Since the largest expected utility, with infinite payoff, assuming the infinite 
number of tosses, by assumption is optimal, one is led to conclude that any rational 
player should decide to always enter the game for any arbitrarily large price. 
However, in real life, human players would not play for more than a few tosses. 
The contradiction between the infinite lottery, assumed to be optimal, and the 
behaviour of real human players, accepting just a few tosses, is the essence of 
the paradox \cite{Bernoulli_1738}. 

First suggestions for avoiding the problem consisted in the introduction of concave 
utility functions, embodying risk aversion and decreasing marginal utility of gains, 
as has been suggested in the letter to Nicolas Bernoulli by Cramer (1728), advising 
to accept the square-root utility function, and by Bernoulli \cite{Bernoulli_1738} 
proposing the logarithmic utility function leading to finite expected utilities. 
However, these ways do not eliminate the paradox, as far as a slight variation of 
payoffs makes the paradox reappear, as is evident from the theorem below.   

\vskip 2mm

{\bf Theorem 4}. {\it For each nonincreasing probability distribution $p_m$, there 
exist nondecreasing utility functions $u(x_m)$ such that the expected utility $U(L_n)$ 
diverges as $n \ra \infty$}. 

\vskip 2mm
{\bf Proof}. For each probability distribution $p_m$, it is possible to define such 
utility functions $u(x_m)$ that for all $m > m^*$ one has the inequality
$$
 \frac{u(x_{m+1})p_{m+1}}{u(x_m)p_m} \geq 1 \qquad ( m \geq m^* ) \; .
$$
Then by the d'Alembert ratio test the sequence $\{U_n\}$ diverges as $n\ra\infty$. 
$\square$

\vskip 2mm

There have been suggested a large number of proposals attempting to avoid the paradox,
as is summarized in \cite{Yukalov_132}, such as the recommendations of assuming the 
boundness of utility functions, changing the expected utility definition, setting a 
finite limit to prizes, setting a finite number of tosses, replacing the single game 
by a set of repeated games, and like that. However all these proposals change the 
formulation of the Bernoulli game, but do not solve the paradox. Replacing one game 
by another is not a solution. 

The resolution of the paradox requires that, for the given game, the definition of 
the optimal lottery be in agreement with the actual choice of decision makers. 
According to the probabilistic approach expounded in this review, the optimal lottery 
is not that enjoying the maximal expected utility, but a lottery characterized by the 
largest probability that is nominated as {\it stochastically optimal}. In the case 
where the expected utility diverges, as in the Bernoulli game, the general form of the 
utility factor $f(L_n)$ imposes a restriction on the parameter $\beta$.

\vskip 2mm
{\bf Theorem 5}. {\it If the maximal expected utility diverges, so that 
${\rm max}_n U(L_n)\ra\infty$, then the normalization condition (\ref{3.15}) requires 
that the belief parameter be negative, $\beta < 0$}. 

\vskip 2mm
{\bf Proof}. This results directly from the requirement of the finite global mean 
(\ref{3.15}), which is necessary for the uniqueness of the probability distribution
defined by the Shore-Jonson theorem \cite{Shore_1980}. Details are given in 
\cite{Yukalov_132}. $\square$ 

\vskip 2mm
Then the following statement holds true \cite{Yukalov_132}. 
 
\vskip 2mm
{\bf Theorem 6}. {\it Assume that in the Bernoulli game the maximal expected 
utility diverges, so that} ${\rm max}_n U(L_n)\ra\infty$. {\it Then, under the 
probability distribution (\ref{3.25}), the stochastically optimal lottery has 
a finite expected utility corresponding to a finite number of tosses}. 

\vskip 2mm 
{\bf Proof}. In the Bernoulli game, the expected utility of a lottery $L_n$ with 
$n$ tosses is positive, $U(L_n) > 0$. By definition, the largest expected utility 
diverges, ${\rm max}_n U(L_n)\ra\infty$, hence according to Theorem 5, the belief 
parameter has to be negative, $\beta < 0$. The stochastically optimal lottery is 
defined as that one providing the maximum of $f(L_n)$, which gives  
$$
U(L_{opt}) = \frac{1}{|\;\bt\;|} \;  .
$$
This utility is finite for any negative value of the belief parameter, therefore there 
is no paradox. $\square$

\vskip 2mm

In the case of the St. Petersburg paradox, the belief parameter is negative because 
of the attitude of decision makers to the game. The formulation of the Bernoulli gamble 
suggests a seeming possibility of gaining infinite, or at least enormous, amount of 
money. For usual people, such a possibility sounds as highly unrealistic, because of 
which they do not give much trust to this promise. That is, a normal person does not 
believe in the fairness of the gamble. This is why the belief parameter is negative,
which implies disbelief.

The parameter $\beta$ is negative because of high uncertainty of the gamble. From the 
other side, the uncertainty of a game can be quantified by the corresponding standard 
deviation. This suggests the way of estimating the value of $\beta$, assuming that its
magnitude is proportional to the standard deviation, so that
$$     
 |\;\bt\;| \sim \sqrt{{\rm var}[\; U(L_n)\;] } \;  ,
$$
where the variance is
$$
{\rm var}[\; U(L_n)\;] = \lgl \; [\; U(L_n)\;]^2\; \rgl -
\lgl \; U(L_n) \; \rgl^2    
$$ 
and 
$$
\lgl \; U(L_n) \; \rgl  = \sum_{n=1}^\infty U(L_n) f(L_n) \; , 
\qquad
\lgl \;[\; U(L_n)\;]^2\; \rgl = 
\sum_{n=1}^\infty [\; U(L_n)\;]^2 f(L_n) \; .
$$
Here the distribution $f(A_n)$ is given by expression (\ref{3.25}).

In the Bernoulli game, $U(L_n) \sim n$, with $n$ being the number of the tosses. 
Then, calculating the variance, the estimate for the absolute value of $\beta$ 
can be found from the equation
$$
|\; \bt \; | \sim 
\frac{1}{\sqrt{2}\;\sinh\left(\frac{|\bt|}{2}\right)} \; .
$$ 
Since the number of the optimal tosses, preferred by a player, is 
$n_{opt}\sim 1/|\bt|$, we find that the estimate for the optimal number of the 
tosses preferred by a typical player satisfies the equation
$$
 n_{opt} \sim  \sqrt{2}\;\sinh\left(\frac{1}{2n_{opt}} \right) .
$$
This gives the estimates for $\beta$ and $n_{opt}$ of order one. Hence the optimal 
number of tosses one would prefer to buy is just a few tosses, in agreement with 
experimental studies 
\cite{Cox_2009,Hayden_2009,Neugebauer_2010,Cox_2018,Nobandegani_2020}. More 
details can be found in Ref. \cite{Yukalov_132}.

\subsection{Martingale illusion}

There exists a situation that sometimes is called the inverse St. Petersburg 
paradox, when one plays a game followed by permanent losses, as a result of which 
the expected utilities are negative and tend to $-\infty$. Nevertheless, some 
people, although of course not all, still continue playing. A typical example 
concerns a roulette. 

There is a well-known martingale strategy, or martingale illusion, used by gamblers 
which appears to guarantee a win. Every time a player loses, he/she doubles his/her 
bet on the next spin of the wheel and continuing to do so on each successive loss, 
when eventually he/she does win. If so, he/she will win back the amount he/she lost 
in addition to a profit equal to the original stake. However, this strategy leads 
to consecutive losses rising to infinity. Then the question is: How would it be 
possible to explain this paradox, when some people keep gambling and losing, instead 
of stopping? \cite{Aloysuis_2003}. 

The general answer is as follows. The case of gambling is different from the choice 
between the lotteries in the St. Petersburg paradox. In the latter, when hesitating 
what ticket to buy, one considers a set of possible lotteries $L_n$, choosing the 
preferable among them. In the martingale game, one never considers all possibilities 
at once, imagining numerous forthcoming losses, but at each step the gambler compares 
just two alternatives: to stop or keep playing. This choice repeats at each next step. 
And at each step, there exists the probability of stopping or playing, which means 
that at each step there are some people deciding to stop, as well as there is a 
fraction of subjects deciding to keep gambling. The details are described 
in \cite{Yukalov_132}.

\subsection{Allais paradox}

One of the oldest paradoxes in decision making is the Allais paradox 
\cite{Allais_1953}. In this paradox, subjects make choices that are incompatible 
with each other, as well as with expected utility theory, because of which this 
effect is called the compatibility violation. 

When choosing between the lotteries
\be
\nonumber
L_1 = \{ 1, \;  1 \} \; , \quad 
L_2 = \{ 1, \; 0.89 \; | \; 5 , \; 0.10 \; | \; 0, \; 0.01 \} \;   ,
\ee
the majority chooses the first lottery, which, according to the expected utility 
theory, implies that the expected utility $U(L_1)$ should be larger than $U(L_2)$. 
When deciding on the lotteries
\be
\nonumber
L_3 = \{ 5, \; 0.1 \; | \; 0, \; 0.9 \} \; , \quad 
L_4 = \{ 1, \; 0.11 \; | \; 0, \; 0.89 \} \;   ,
\ee
the majority prefers the lottery $L_3$, which means that $U(L_3)$ has to be larger 
than $U(L_4)$.  

The expected utilities of the lotteries are
$$
U(L_1) = u(1) \; , 
$$
\be
\nonumber
  U(L_2) = 0.01 u(0) + 0.89 u(1) + 0.1 u(5) \; , 
\ee
$$
 U(L_3) = 0.9 u(0) + 0.1 u(5) \; , 
$$
\be
\nonumber
 U(L_4) = 0.89 u(0) + 0.11 u(1) \; , 
\ee
where $u(x)$ is an arbitrary utility function. From $U(L_1) > U(L_2)$ one has
\be
\nonumber
0.11 u(1) >  0.01 u(0) + 0.1 u(5) \;  ,
\ee
while from $U(L_3) > U(L_4)$ it follows that
\be
\nonumber
0.11 u(1) <  0.01 u(0) + 0.1 u(5) \;   .
\ee
As is evident, the last two inequalities are incompatible, hence utility theory is not 
obeyed. This contradiction does not depend on the form of the utility function. 

The resolution of the Allais paradox is straightforward in the frame of the affective 
decision theory. Taking for simplicity the linear utility function, we have the utility 
factors
\be
\nonumber
 f(L_1) = 0.418 \; , \qquad f(L_2) = 0.582 \;  ,
\ee
showing that the second lottery is more useful. However the lottery qualities
\be
\nonumber
Q_1 = 30 \; , \qquad Q_2 = 27.7
\ee
show that the first lottery is more attractive, hence for the non-informative priors 
we have the attraction factors $q(L_1) = 0.25$, while $q(L_2) = - 0.25$. This gives 
the behavioural probabilities $p(L_1) = 0.67$ and $p(L_2) = 0.33$. That is, the first 
lottery is preferable. 

Considering the choice between $L_3$ and $L_4$, we get the rational fractions
\be
\nonumber
f(L_3) = 0.82 \; , \qquad f(L_4) = 0.18 \; 
\ee
and the lottery qualities
\be
\nonumber
Q_3 = 7.03 \; , \qquad Q_4 = 1.45 \; .
\ee
This tells us that $L_3$ is more useful as well as more attractive than $L_4$. So 
the absolute majority of people will prefer $L_3$. In that way, there is no any 
contradiction between the theory and the actual behaviour of decision 
makers \cite{Yukalov_78}.

\subsection{Independence paradox}

The Allais paradox in decision making contradicts the independence axiom of expected 
utility theory \cite{Allais_1953,Machina_1987}. This is easy to show taking the 
lotteries from the previous subsection and composing the mixed lotteries
\be
\nonumber
L_5 = \al L_1 + ( 1 - \al ) L_3 \; , \qquad L_6 = \al L_2 + ( 1 - \al ) L_4 \;  .
\ee
The related expected utilities are
$$
U(L_5) = \al U(L_1) + ( 1 - \al ) U(L_3) \; ,
$$
\be
\nonumber
U(L_6) = \al U(L_2) + ( 1 - \al ) U(L_4) \;.
\ee
The independence axiom in the expected utility theory requires that if $L_1$ is 
preferred to $L_2$ in the sense that $U(L_1) > U(L_2)$, and $L_3$ is preferred to 
$L_4$, so that $U(L_3) > U(L_4)$, then $L_5$ is to be preferred to $L_6$, with 
$U(L_5) > U(L_6)$. However, for the given lotteries $L_1$, $L_2$, $L_3$, and $L_4$, 
it follows that $U(L_5) = U(L_6)$.

Contrary to this, in the approach taking account of emotions, preferences are understood 
as the corresponding relations between the behavioural probabilities $p(L_n)$ for which 
there is no independence axiom. Mixed lotteries $L_5$ and $L_6$ are new lotteries to be 
treated separately from the terms they are composed of. From $p(L_1)\geq p(L_2)$ and 
$p(L_3)\geq p(L_4)$ it does not follow that $p(L_5)$ is larger than $p(L_6)$. If it has 
happened that $L_5$ is identical to $L_6$, this just means that $p(L_5) = p(L_6)$ and 
there is no paradox \cite{Yukalov_78}.

\subsection{Ellsberg paradox}

There exist different variants of the Ellsberg paradox 
\cite{Ellsberg_1961,Machina_2009,Machina_2014}. Here, we consider the original 
version of Ellsberg \cite{Ellsberg_1961} with two urns setup. 

In one of the urns, there are $50$ red and $50$ black balls, while in the other urn, 
there are $100$ red and black balls with an unknown portion $p$ of red balls. People 
are offered to draw a ball from one of the urns getting a payoff for the red ball and 
no payoff for the black ball. Which of the urns to choose? In the formal language, 
this is the choice between the lotteries 
\be
\nonumber
L_1 = \{ 1, \; 0.5 \; | \; 0, \; 0.5 \} \; , \qquad 
L_2 = \{ 1, \; p \; | \; 0, \; 1-p \} \;   .
\ee
Here payoffs are measured in some monetary units. Since the lottery $L_2$ is 
uncertain, people, following the principle of uncertainty aversion, choose $L_1$ 
over $L_2$, which in the expected utility theory implies that $U(L_1)$ is larger 
than $U(L_2)$.

Then subjects are suggested to draw a black ball from one of the urns, which assumes 
the choice between the lotteries
\be
\nonumber
L_3 = \{ 0, \; 0.5 \; | \; 1, \; 0.5 \} \; , \qquad 
L_4 = \{ 0, \; p \; | \; 1, \; 1 - p \} \;   .
\ee
Again, people avoid uncertainty preferring a more certain urn choosing $L_3$ over 
$L_4$, which means that $U(L_3)$ should be larger than $U(L_4)$. 

From the inequality $U(L_1) > U(L_2)$ one has
\be
\nonumber
\left( \frac{1}{2} \; - \; p \right) u(1) > \left( \frac{1}{2} \; - \; 
p \right) u(0) \;   .
\ee
However from $U(L_3) > U(L_4)$, it follows that
\be
\nonumber
\left( \frac{1}{2} \; - \; p \right) u(1) < \left( \frac{1}{2} \; - \; 
p \right) u(0) \; ,
\ee
which is in evident contradiction with the previous inequality for any $p\in[0,1]$
and for any utility function.  

The resolution of the Ellsberg paradox in the affective decision theory is as 
follows. For the first choice between $L_1$ and $L_2$, keeping in mind, for 
simplicity, a linear utility function, we find the utility factors
\be
\nonumber
 f(L_1) = \frac{1}{1+2p} \; , \qquad f(L_2) = \frac{2p}{1+2p} \;  .
\ee
The second lottery $L_2$, being uncertain, is less attractive, hence the attraction 
factors are such that $q(L_1) > q(L_2)$. This is prescribed by the criterion of 
uncertainty aversion.

Therefore, using the non-informative priors, we have $q(L_1)=0.25$ and $q(L_2)=-0.25$. 
This gives the behavioural probabilities 
$$
 p(L_1) = f(L_1) + 0.25 = \frac{5+2p}{4(1+2p)} \; , 
$$
\be
\nonumber
 p(L_2) = f(L_2) - 0.25 = \frac{6p-1}{4(1+2p)} \; .
\ee
Due to the positive difference 
\be
\nonumber
p(L_1) - p(L_2) =  \frac{3-2p}{2(1+2p)} > 0 \;  ,
\ee
it follows that the lottery $L_1$ is stochastically preferable over $L_2$. 

In the case of the second choice between $L_3$ and $L_4$, the utility factors are
\be
\nonumber
 f(L_3) = \frac{1}{3-2p} \; , \qquad f(L_4) = \frac{2(1-p)}{3-2p} \;  .
\ee
Since $L_4$ is uncertain, for the attraction factors we have $q(L_3) > q(L_4)$. With
the use of the non-informative priors, this means that $q(L_3) = 0.25$ and 
$q(L_4) = - 0.25$. Thus, the behavioural probabilities are
$$
p(L_3) = f(L_3) + 0.25 = \frac{7-2p}{4(3-2p)} \; , 
$$
\be
\nonumber
 p(L_4) = f(L_4) - 0.25 = \frac{5-6p}{4(3-2p)} \; .
\ee
Again, the positive difference
\be
\nonumber
p(L_3) - p(L_4) =  \frac{1+2p}{2(3-2p)} > 0 
\ee
tells us that the lottery $L_3$ is stochastically preferable over $L_4$. And there 
is no any paradox \cite{Yukalov_78}.

\subsection{Prisoner dilemma}

A great number of gambles has the structure of the prisoner dilemma game 
\cite{Poundstone_1992,Kaminski_2004}. It is therefore instructive to consider how 
this game is treated in the decision theory taking into account emotions.  

One considers two players that occurred to be imprisoned and each of them is given 
the choice of either betray the other or cooperate with the other, not betraying 
him/her. Let us denote the action of cooperating by the first prisoner by $A_1$ 
and the action of betraying by $A_2$. Similarly, the action of cooperation by the 
second prisoner is denoted by $B_1$, while the action of betraying, by $B_2$.  

Thus there are the following possible cases: both cooperate (event $A_1 B_1$), the 
first cooperates, while the other betrays (event $A_1 B_2$), the first betrays, while 
the other cooperates (event $A_2 B_1$), both betray each other (event $A_2 B_2$). Two 
setups can be considered. In one setup a prisoner has to decide knowing the action of 
the other prisoner. The other setup is when a prisoner has to decide being not aware 
of the other prisoner action (event $B = B_1 \bigcup B_2$). Conditions are arranged 
in such a way that, according to the rational rules, it is always more profitable to 
betray, since betraying is the Nash equilibrium in that game. Hence, being given the 
choice, each prisoner should betray the other with the $100\%$ probability. In other 
words, when several prisoners are in the game, all of them should betray the others. 
Surprisingly, this does not happen in empirical tests, but an essential number of 
decision makers chooses not to betray, thus contradicting the expected utility 
theory, which is especially noticeable when decisions are taken under uncertainty. 
In the present case, the maximal deviation from the prescription of the expected 
decision theory happens when a prisoner decides not knowing the action of the other 
prisoner. 

In the probabilistic approach, each of the decision makers confronting the choice 
between two alternatives to betray or not to betray and being not aware of the 
choice of the opponent, estimates two behavioural probabilities. Since the situation 
is symmetric with respect to the prisoners, it is sufficient to consider only one of 
them, say, the first prisoner who decides on either to cooperate, not knowing the 
action of the other prisoner, or to betray, also not being aware of the action of 
the second prisoner. Thus, the first prisoner needs to estimate the probabilities
$$
p(A_1 B) = f(A_1 B) + q(A_1 B) \; , 
$$
\be
\nonumber
  p(A_2 B) = f(A_2 B) + q(A_2 B) \;  .
\ee
The utility factors, being the standard classical probabilities, can be expressed 
through conditional probabilities
\be
\nonumber
f(A_1 B) = f(A_1|B_1) f(B_1) + f(A_1|B_2) f(B_2) \; ,
\ee
\be 
\nonumber
 f(A_2 B) = f(A_2|B_1) f(B_1) + f(A_2|B_2) f(B_2) \;  .
\ee
The a priori probabilities $f(B_\al)$ can be assumed to be equal, hence 
$f(B_\al)=1/2$. The conditional probabilities satisfy the standard normalization 
$f(A_1|B_\al) + f(A_2|B_\al) = 1$ with $\al = 1,2$.

There has been a number of empirical tests studying the prisoner dilemma, with the 
results very close to each other. We consider here the results of the classical 
investigations by Tversky \cite{Tversky_2004} and Tversky and Shafir 
\cite{Tversky_Shafir_1992}. In these experiments, eighty subjects played a series 
of prisoner dilemma games, without feedback, each against a different unknown 
opponent supposedly selected at random from among the participants. Three types 
of setups were used: (i) when the subjects knew that the opponent had defected, 
(ii) when they knew that the opponent had cooperated, and (iii) when the subjects did 
not know whether their opponent had cooperated or defected. The typical statistical 
error of the experiments was about $20 \%$. It was found that the fractions of 
decision makers choosing to betray or to cooperate when knowing with certainty that 
the other side betrayed or, respectively, cooperated were
$$
f(A_1|B_1) = 0.16 \; , \qquad f(A_2|B_1) = 0.84 \; , 
$$
\be
\nonumber
f(A_1|B_2) = 0.03 \; , \qquad f(A_2|B_2) = 0.97 \; . 
\ee
From here, we find the utility factors $f(A_1 B)=0.095$ and $f(A_2 B)=0.905$.

At the same time, as has been formulated in the principle of propensity for 
cooperation, that is a well documented evolutionary law 
\cite{Camerer_2003,Tversky_2004,Tomasello_2009,Tomasello_2012,Jusup_2021}, people 
have general understanding that betraying is not an attractive action 
\cite{Moors_2019,Poundstone_1992,Kaminski_2004}. Therefore, cooperation looks 
to be more attractive, which, according to the non-informative prior estimates, 
means that $q(A_1 B) = 0.25$, while $q(A_2 B) = - 0.25$. Substituting the values 
of the utility factors and attraction factors into the behavioural probabilities, 
corresponding to the case when participants do not know whether their opponent 
betrayed or cooperated, we get the predicted probabilities
\be
\nonumber
 p(A_1 B) = 0.35 \; ,  \qquad p(A_2 B) = 0.65 \; . 
\ee
These values practically coincide, within the accuracy of the experiments, 
with the aggregate results of Tversky and Shafir \cite{Tversky_Shafir_1992}
\be
\nonumber
  p_{exp}(A_1 B) = 0.37 \; , \qquad p_{exp}(A_2 B) = 0.63 \; .
\ee
The use of the non-informative prior for the attraction factor allows us to 
make a very accurate prediction without any fitting parameters \cite{Yukalov_79}.

\subsection{Disjunction effect}

Disjunction effect \cite{Tversky_Shafir_1992} demonstrates that people do not 
follow the Savage sure-thing principle, although the latter is a direct consequence 
of the classical probability theory \cite{Savage_1954}.

One considers a two-step gamble. In the first gamble, one can either win (event 
$B_1$) or lose (event $B_2$). Then one has to decide on the second gamble, either 
accepting it (event $A_1$) or refusing from gambling (event $A_2$). 

In one setup, the participants know the outcome of the first gamble. In that 
case, the majority of people accepts the second gamble, independently of the known 
outcome of the first gamble. That is, more people accept the second gamble after 
winning the first, as well as more people accept the second gamble after losing 
the first. In the language of the classical probability theory, this means that 
$f(A_1 B_\al) > f(A_2 B_\al)$, where $\al = 1,2$.

In the other setup, the results of the first gamble are not announced, so that 
although the gamble has happened, its outcome is not known (event 
$B\equiv B_1\bigcup B_2$). In that case, when the results of the first gamble 
are not known, the majority restrains from the second gamble, which could imply 
that $f(A_1 B)$ is smaller than $f(A_2 B)$. However, for the usual probabilities 
one has
$$
f(A_1B) = f(A_1B_1) + f(A_1B_2) \; , 
$$
\be
\nonumber
f(A_2B) = f(A_2B_1) + f(A_2B_2) \;   .
\ee
Since $f(A_1B_\al)>f(A_2B_\al)$, then it should be that $f(A_1B)>f(A_2B)$, which 
is the Savage sure-thing principle \cite{Savage_1954}: If one prefers an action 
$A_1$ to action $A_2$, when event $B_1$ happens, and one prefers $A_1$ still, 
when another event $B_2$ happens, then one should prefer the action $A_1$ despite 
having no knowledge of whether event $B_1$ or $B_2$ has happened. Thus, the actual 
decisions of humans contradict the sure-thing principle. 

Tversky and Shafir \cite{Tversky_Shafir_1992} accomplished several experiments 
confirming the disjunction effect. 

\vskip 2mm
(i) {\it To gamble or not to gamble}. One of the experiments employed the setup 
exactly as described above. When the results of the first gamble were known, the 
corresponding fractions of subjects were
$$
f(A_1|B_1) = 0.69 \; , \qquad f(A_2|B_1) = 0.31 \; ,
$$
\be
\nonumber
f(A_1|B_2) = 0.59 \; , \qquad f(A_2|B_2) = 0.41 \;  .
\ee
Proceeding analogously to the prisoner dilemma, we find the rational utility factors
\be
\nonumber
 f(A_1B) = 0.64 \; , \qquad  f(A_2B) = 0.36 \;  ,
\ee
in agreement with the Savage sure-thing principle. However, in reality decision 
makers acted in the opposite way. 

In the approach suggested in this paper, the behavioural probabilities, in addition 
to utility factors include attraction factors. As has been explained above, in 
line with the principle of uncertainty aversion, decisions requiring activity in the 
presence of uncertainty are less attractive than passive waiting \cite{Moors_2019}. 
This corresponds to the attraction factors $q(A_1 B) < q(A_2 B)$. Using the 
non-informative priors, this gives $q(A_1 B) = - 0.25$ and $q(A_2 B) = 0.25$. Thus, 
the behavioural probabilities become
\be
\nonumber
p(A_1B) = 0.39 \; , \qquad  p(A_2B) = 0.61 \;    ,
\ee
which practically coincides with the empirical observations of Tversky and Shafir 
\cite{Tversky_Shafir_1992}, giving
\be
\nonumber
p_{exp}(A_1B) = 0.36 \; , \qquad  p_{exp}(A_2B) = 0.64 \;   .
\ee

\vskip 2mm

(ii) {\it To go or not to go}. The other setup concerned students undergoing 
examinations. At the first stage, there are two alternatives: exam passed (event  
$B_1$) or exam failed (event $B_2$). Then the students are suggested to decide on 
going to vacations to a resort. Here two possible actions are available: vacation 
accepted (event $A_1$) or vacation refused (event $A_2$). It turned out that the 
majority of students decide to go to vacations regardless to the results, whether 
they passed the exam or failed. This means that similarly to the previous example 
of gambling, $f(A_1 B_\al) > f(A_2 B_\al)$, where $\al = 1,2$. According to Tversky 
and Shafir \cite{Tversky_Shafir_1992}, 
$$
f(A_1|B_1) = 0.54 \; , \qquad f(A_2|B_1) = 0.46 \; , 
$$
\be
\nonumber
f(A_1|B_2) = 0.57 \; , \qquad f(A_2|B_2) = 0.43 \;   .
\ee
In the other setup, students need to decide on going to vacations when they have 
undergone examinations but the outcomes are not announced. So that students do 
not know whether they have passed the exam or failed. In that case, it is 
straightforward to find the rational utility factors in the same way as above by 
obtaining $f(A_1B)=0.555$ and $f(A_2B)=0.445$, which agrees with the sure-thing 
principle. However, when the results are not known, the majority of students 
forgo vacations, which contradicts the sure-thing principle. 

Similarly to the previous example, since the activity under uncertainty is less 
attractive, due to uncertainty aversion, the attraction factors are $q(A_1 B)=-0.25$ 
and $q(A_2 B) =  0.25$. Hence the behavioural probabilities are
\be
\nonumber
p(A_1B) = 0.31 \; , \qquad  p(A_2B) = 0.69 \;   ,
\ee
in perfect agreement with the empirical data
\be
\nonumber
p_{exp}(A_1B) = 0.32 \; , \qquad  p_{exp}(A_2B) = 0.68 \;   .
\ee
 
Again, good agreement of predictions with experimental results are obtained without 
using any fitting parameters \cite{Yukalov_78,Yukalov_92}.

\subsection{Conjunction fallacy}

The conjunction fallacy constitutes another example illustrating that human beings 
do not conform to the standard probability theory. This effect was first described 
by Tversky and Kahneman \cite{Tversky_1983}. There have been numerous studies 
observing the effect. Here we shall consider the results of the experiments by 
Shafir, Smith, and Osherson \cite{Shafir_1990}. These experiments up-to-date, 
probably, are the most detailed. 

Shafir, Smith, and Osherson \cite{Shafir_1990} questioned large groups of students 
who were provided with booklets each containing a brief description of a person. 
It was stated that the described person could have a primary characteristic and also 
another characteristic. In total, there were around $100$ students. The subjects had 
to decide on the existence of specific features of persons, whose description had 
been apriori given. The prominent example is the following description: "Linda is 
$31$ years old, single, outspoken, and very bright. She majored in philosophy. As 
a student, she was deeply concerned with issues of discrimination and social justice, 
and also participated in anti-nuclear demonstrations. Which is more likely? Linda is 
a bank teller? Linda is a bank teller and is active in the feminist movement?" Most 
people answer that it is more probable that Linda is a bank teller and feminist. 

This answer contradicts conjunction rule, according to which the probability of two 
events occurring in conjunction (that is together) is always less than or equal to the 
probability of either one occurring alone.
    
Features were grouped in pairs. In each pair there could be compatible or 
incompatible characteristics. The characteristics were treated as compatible, 
when they were felt as closely related according to some traditional wisdom, for 
instance, "woman teacher" and "feminist". Those characteristics that were not 
related by direct logical connections were considered as incompatible, such as 
"bird watcher" and "truck driver" or "bicycle racer" and "nurse". For compatible 
characteristics, as is expected, the probabilities of a conjunction category were 
very close to those of its constituents. Thus, the aggregate probability of a 
conjunction characteristic was $0.567$, while that of its constituents was $0.537$. 
Within the typical statistical error of $20\%$, these values can be treated as 
equal, which means that, within this accuracy, the effect is too weak to be considered 
seriously. So, although the conjunction effect can, in principle, be considered for 
compatible characteristics, it is an order smaller of magnitude than the conjunction 
fallacy for incompatible characteristics. Therefore, below we shall consider the more 
clear case of incompatible characteristics.  

Shafir, Smith, and Osherson \cite{Shafir_1990} also studied the difference between 
the notions of typicality and probability. Since normal people usually understand 
typicality just as a synonym of probability, the results in the case of typicality 
were indistinguishable from those corresponding to probabilities. Because of this, 
it is sufficient to limit ourselves to studying the results dealing with probability.  

It is also important to specify how the questions are asked. There exist two ways 
of posing questions. One way would be by suggesting to compare two simultaneously 
defined alternatives, a conjunction characteristic and its constituent, as has been
done by Tversky and Kahneman \cite{Tversky_1983}, say, "bank teller and feminist" 
and just "bank teller". The other way is asking to make decisions independently, 
on a conjunction characteristic and separately on its constituent, as is considered
by Shafir, Smith, and Osherson \cite{Shafir_1990}. The setup of Tversky and Kahneman 
\cite{Tversky_1983} has been considered in \cite{Yukalov_92}, the setup of Shafir, 
Smith, and Osherson \cite{Shafir_1990}, in \cite{Kovalenko_2022}. Below we consider 
the independent way of making decisions as in the experiments of Shafir, Smith, and 
Osherson \cite{Shafir_1990}, but by using methodology different from that employed 
by \cite{Kovalenko_2022}, and involving no quantum techniques.  

The resolution of conjunction fallacy is based on taking account of the 
attractiveness of different alternatives that are classified according to their 
certainty, respectively, with respect to the uncertainty aversion.

Let us denote the decision that the person possesses a primary feature by $A_1$ 
and that the person does not have this feature, by $A_2$. The decision that the 
person has a secondary feature will be denote by $B_1$ and that he/she does not 
have it, by $B_2$. The detailed illustration is given below for the case of Linda. 
Other cases are treated similarly. Then the primary feature $A_1$ implies 
being a bank teller, and $A_2$ means not being a bank teller.   

When subjects are inquired on the existence of just the primary feature, they meet 
the problem of choosing between $A_1$ (bank teller) and $A_2$ (not bank teller). 
This means that one has to evaluate the behavioural probabilities
$$
p(A_1) = f(A_1) + q(A_1) \; , \qquad p(A_2) = f(A_2) + q(A_2) \;   .
$$
Deciding on utility grounds, the existence or absence of the considered feature 
seem to be equally probable, hence $f(A_n) = 1/2$. However presented description 
hints on the existence or absence of that feature. Thus, the characteristic ``bank 
teller" does not follow from the description of Linda. Deciding whether Linda is 
a bank teller involves quite a strong uncertainty with respect to why she would 
be exactly a bank teller but not somebody else. Therefore, the precise statement 
that she is a bank teller is highly uncertain, hence not attractive, which suggests 
that $q(A_1) = - 0.25$ and $q(A_2) = 0.25$. Then the behavioural probabilities 
become $p(A_1) = 0.25$ and $p(A_2) = 0.75$.

In the case of deciding on the conjunction of two characteristics, say, bank teller 
and feminist, there are four alternatives: $A_1 B_1$ (bank teller and feminist), 
$A_1 B_2$ (bank teller but not feminist), $A_2 B_1$ (not bank teller but feminist), 
$A_2 B_2$ (not bank teller and not feminist). Respectively, there are four behavioural
probabilities
$$
p(A_1B_1) = f(A_1B_1) + q(A_1B_1) \; , 
$$
\be
\nonumber
 p(A_1B_2) = f(A_1B_2) + q(A_1B_2) \; ,
\ee
$$
p(A_2B_1) = f(A_2B_1) + q(A_2B_1) \; , 
$$
\be
\nonumber
 p(A_2B_2) = f(A_2B_2) + q(A_2B_2) \; .
\ee
The normalization conditions are valid:
\be
\nonumber
\sum_{ij} p(A_iB_j) = 1 \; , \qquad  \sum_{ij} f(A_iB_j) = 1 \; ,
\ee
\be
\nonumber
\sum_{ij} q(A_iB_j) = 0 \qquad ( i,j = 1,2 ) \; .
\ee

Since the considered pairs of features are not compatible (bank teller has no direct 
relation to feminist), utility factors seem to be equal, which, taking account of the 
above normalizations, leads to the equality $f(A_i B_j) = 1/4$.

To estimate attraction factors, we can use the results of Sec. $3.5$ considering 
multiple alternatives. For this purpose, we need to arrange the four given 
alternatives in the order of their attractiveness based on the level of their 
certainty (uncertainty). For concreteness, let us again consider the case of Linda 
mentioned above. The other cases are treated in the same way. From the suggested 
description of Linda, it is almost evident that she is a feminist. Whether she is a 
bank teller or not is uncertain, since this does not follow from the description. 
The probability that Linda is a bank teller, that is assuming one of the professions 
from thousands of others equally possible, is rather uncertain. In that way, comparing 
the alternatives $A_2 B_1$ (not bank teller but feminist) and $A_1B_1$ (bank teller 
and feminist), the alternative $A_2 B_1$, according to the provided description, 
looks more certain, hence more attractive. Comparing the alternative $A_1B_1$ (bank 
teller and feminist) with the alternative $A_2B_2$ (not bank teller and not feminist), 
the first alternative $A_1B_1$ (bank teller and feminist) sounds more certain, since 
the given description clearly hints on the feminist feature. Finally, in the comparison 
of the alternative $A_2B_2$ (not bank teller and not feminist) and the alternative 
$A_1B_2$ (bank teller but not feminist) the latter is evidently more uncertain, hence 
less attractive, since from the description it does not follow that Linda has anything 
to do with banks. 
  
Therefore the order of attraction factors characterizing the attractiveness of the 
four alternatives, arranged according to their certainty, is 
$$ 
 q(A_2B_1) > q(A_1B_1) > q(A_2B_2) > q(A_1B_2) \; .
$$
Then from Sec. $3.5$, we have the non-informative prior estimates:
$$
q(A_2B_1) = \frac{3}{8} \; , \qquad q(A_1B_1) = \frac{1}{8} \; , 
$$
\be
\nonumber 
q(A_2B_2) = -\; \frac{1}{8} \; , \qquad q(A_1B_2) = -\; \frac{3}{8} \; .
\ee
Thus we find the behavioural probability that Linda is a bank teller and feminist 
as $p(A_1 B_1) = 1/4 + 1/8 = 0.375$. 

In this way, we have 
\be
\nonumber
p(A_1) = 0.25 \; , \qquad  p(A_1 B_1) = 0.375 \; ,
\ee
which practically coincides with the average empirical data in the experiments 
of Shafir, Smith, and Osherson \cite{Shafir_1990}, being 
\be
\nonumber
p_{exp}(A_1) = 0.22 \; ,  \qquad p_{exp}(A_1 B_1) = 0.346 \; .
\ee
The conjunction error is defined as 
$$
  \ep(A_1B_1) \equiv p(A_1B_1) - p(A_1) \; .
$$
The predicted and average experimental conjunction errors are practically equal:
\be
\nonumber
 \ep(A_1B_1) = 0.125 \; , \qquad  \ep_{exp}(A_1B_1) = 0.126 \;  .
\ee

Of course, in each particular case the conjunction errors vary, as is seen in 
Table 5, where the corresponding experimental probabilities and errors are shown 
for different pairs of characteristics \cite{Shafir_1990}. At the bottom of Table 5, 
the average values are given.

\begin{table}[hp]
\caption{Conjunction fallacy. Pairs of characteristics, probability of the 
conjunction $p(A_1B_1)$, probability of the primary feature $p(A_1)$, and 
conjunction error $\ep(A_1B_1)$. The bottom line presents the related average 
values.}
\vskip 2mm
\centering
\renewcommand{\arraystretch}{1.2}
\begin{tabular}{|l|c|c|c|} \hline
characteristics            &  $p(A_1B_1)$  &  $p(A_1)$  &  $\ep(A_1B_1)$   \\ \hline
$A_1$ bank teller          & 0.401         & 0.241      &    0.160  \\ 
$B_1$ feminist             &               &            &           \\ 
$A_1$ bird watcher         & 0.274         & 0.173      &    0.101   \\ 
$B_1$ truck driver         &               &            &            \\
$A_1$ bicycle racer        & 0.226         & 0.160      &    0.066     \\ 
$B_1$ nurse                &               &            &            \\
$A_1$ drum player          & 0.367         & 0.266      &    0.101     \\
$B_1$ professor            &               &            &              \\ 
$A_1$ boxer                & 0.269         & 0.202      &    0.067   \\ 
$B_1$ chef                 &               &            &          \\  
$A_1$ volleyboller         & 0.282         & 0.194      &    0.088  \\ 
$B_1$ engineer             &               &            &           \\ 
$A_1$ librarian            & 0.377         & 0.152      &    0.225   \\ 
$B_1$ aerobic trainer      &               &            &            \\
$A_1$ hair dresser         & 0.252         & 0.188      &    0.064     \\ 
$B_1$ writer               &               &            &            \\
$A_1$ floriculturist       & 0.471         & 0.310      &    0.161     \\
$B_1$ state worker         &               &            &              \\ 
$A_1$ bus driver           & 0.314         & 0.172      &    0.142   \\ 
$B_1$ painter              &               &            &          \\ 
$A_1$ knitter              & 0.580         & 0.315      &    0.265  \\ 
$B_1$ correspondent        &               &            &           \\ 
$A_1$ construction worker  & 0.249         & 0.131      &    0.118   \\ 
$B_1$ labor-union president&               &            &            \\
$A_1$ flute player         & 0.339         & 0.180      &    0.159   \\ 
$B_1$ car mechanic         &               &            &            \\
$A_1$ student              & 0.439         & 0.392      &    0.047   \\
$B_1$ fashion-monger       &               &            &        \\  \hline
                           & 0.346         & 0.220      &    0.126 \\ \hline
\end{tabular}
\end{table}

\subsection{Disposition effect}

This effect demonstrates the asymmetry of subjects' behaviour with respect to 
gains and losses \cite{Shefrin_1985}. It also relates to the tendency of investors 
to sell assets that have increased in value, while keeping assets that have dropped 
in value \cite{Weber_1995}. It can be illustrated by the following example 
\cite{Kahneman_1979}.

Consider the lotteries with gains 
$$
L_1 = \{ x , 0.5 \; | \; 0 , 0.5 \} \; , \qquad
L_2 = \left\{ \frac{x}{2} , 1 \; | \; 0 , 0 \right\} \;  ,
$$
whose expected utilities, under linear utility function, are equal
$$
U(L_1) = U(L_2) = x \;  .
$$
Despite that the expected utilities are equal, people prefer more certain lottery 
$L_2$.

Now let us turn the gains into losses by replacing in the above lotteries $x$ by 
$-x$. Then we get the lotteries with losses
$$
L_3 = \{ -x , 0.5 \; | \; 0 , 0.5 \} \; , \qquad
L_4 = \left\{ -\;\frac{x}{2} , 1 \; | \; 0 , 0 \right\} \; ,
$$
whose expected utilities are again equal,
$$
 U(L_3) = U(L_4) = - x \;  .
$$
Thereat people prefer less certain lottery $L_3$. 

That is, there is the asymmetry between gains and losses, so that in the case 
of gains, one prefers the more certain lottery, while in the case of losses, one 
chooses the less certain lottery. It is, of course, possible to invent special 
forms of utility functions, replace in the lotteries the payoff probabilities by 
some specially constructed weights \cite{Kahneman_1979}, and to fit the parameters 
so that to conform the empirically observed asymmetry. However fitting is not 
explanation. 

The explanation of the disposition effect is easy, following the approach 
described in the present paper. Calculating for the lotteries $L_1$ and $L_2$ 
the quality functionals gives
$$
Q(L_1) = x \cdot 30^{0.5} = 5.5 x  \; , \qquad
Q(L_2) = \frac{x}{2} \; \cdot \; 30 = 15 x  \;  ,
$$
which shows that $Q(L_1) < Q(L_2)$, hence the second lottery is more 
attractive. 

For the case of losses, we have the quality functionals
$$
Q(L_3) = - Q(L_1) =  -5.5 x \; , \qquad
Q(L_4) = - Q(L_2) =  -15 x \; ,
$$
which tells us that $Q(L_3) > Q(L_4)$ and the third lottery is more attractive. 

As is straightforward to see, the utility factors for all lotteries are equal
$$
 f(L_1) = f(L_2) = f(L_3) = f(L_4) = \frac{1}{2} \; .
$$
Estimating the attraction factors by non-informative priors, we obtain the 
probabilities of the optimal lotteries
$$        
 p(L_2) = 0.75 \; , \qquad p(L_3) = 0.75 \;  .
$$
In experiments \cite{Kahneman_1979}, the probabilities of optimal lotteries vary 
depending on the payoff $x$. The average over the lotteries \cite{Yukalov_125} 
gives for the optimal lottery $p_{exp}(L_{opt}) = 0.77$, practically coinciding 
with the predicted probability $p(L_{opt}) = 0.75$. This explains why the more 
certain lottery $L_2$ is preferred to $L_1$, while the less certain lottery $L_3$ 
is preferred to $L_4$. The asymmetry between gains and losses is due to different 
attractiveness of the related lotteries.

\subsection{Ariely paradox}

Ariely \cite{Ariely_30} describes a series of experiments stressing the drastic 
change in the choice of customers buying goods when one of them becomes more 
attractive, being suggested for free. The structure of this effect is as follows. 
Two types of goods, e.g., the Lindt truffle and Kiss chocolate, are sold for the 
reasonable prices, which in the formal language amounts to choosing between two 
lotteries, $L_1$ and $L_2$ whose expected utilities are practically the same, 
$U(L_1) = U(L_2)$. The equality of the expected utilities can be understood as 
due the compensation of quality by price, because the Lindt truffle, being more 
expensive, is also of better quality, while the Kiss chocolate, being cheaper,
is of lower quality. However, the difference is in the emotions caused by the 
pleasure of eating more tasty food, which makes the objects more or less attractive,
so that the expected pleasure, hence the attractiveness, of the first item (Lindt 
truffle) in the lottery $L_1$ is higher than that of the second item (Kiss chocolate) 
in the lottery $L_2$. Not surprisingly, the majority of people prefer $L_1$. 

Then the price of the goods is reduced to such a level that the first item (Lindt 
truffle) is sold slightly cheaper (lottery $L_3$), while the second item (Kiss 
chocolate) becomes free (lottery $L_4$), although the utilities of the new lotteries 
are assumed to remain equal. Now people have to choose between the lotteries $L_3$ 
and $L_4$. It looks natural that if the majority were choosing the first item (Lindt 
truffle) that was of better quality, the same choice should persist. However, people 
prefer the lottery with a free item (Kiss chocolate), that is lottery $L_4$.       
 
The resolution of this paradox is straightforward. As far as the utilities of 
the lotteries $L_1$ and $L_2$ are the same, the utility factors are also equal, 
$f(L_1) = f(L_2) = 0.5$. Due to the higher quality, the expected pleasure, hence 
the attractiveness, of $L_1$, the attraction factors are $q(L_1) = 0.25$ and 
$q(L_2) = - 0.25$. Then the behavioural probabilities are 
\be
\nonumber
 p(L_1) = 0.75 \; , \qquad p(L_2) = 0.25 \;  .
\ee
This is in beautiful agreement with the experiments \cite{Ariely_30}, where the 
Lindt truffle and Kiss chocolate were the items of sale, 
\be
\nonumber
p_{exp}(L_1) = 0.73 \; , \qquad p_{exp}(L_2) = 0.27 \;   .
\ee

In the case of the choice between the lotteries $L_3$ and $L_4$, where the Kiss 
chocolate is suggested for free but the utilities of the lotteries again are equal, 
the rational fractions are $f(L_3) = f(L_4) = 0.5$. However getting a free item 
looks much more attractive, hence $q(L_3) = - 0.25$ and $q(L_4) = 0.25$. This gives 
the behavioural probabilities
\be
\nonumber
p(L_3) = 0.25 \; , \qquad p(L_4) = 0.75 \;   ,
\ee
again close to the experimental results
\be
\nonumber
p_{exp}(L_3) = 0.31 \; , \qquad p_{exp}(L_4) = 0.69 \;    .
\ee
In this way, taking into account emotions distinguishing the alternatives as more or 
less attractive, it is easy to predict the people behaviour.

\subsection{Decoy effect}

The decoy effect was first studied by Huber, Payne and Puto \cite{Huber_1982}, who 
called it the effect of asymmetrically dominated alternatives. Later this effect has 
been confirmed in a number of experimental studies \cite{Simonson_1989,Wedell_1991,
Tversky_Simonson_1993,Ariely_1995}. The meaning of the decoy effect can be illustrated 
by the following example. Suppose a buyer is choosing between two objects, $A$ and $B$. 
The object $A$ is of better quality, but of higher price, while the object $B$ is of 
slightly lower quality, while less expensive. As far as the functional properties of 
both objects are not drastically different, but $B$ is cheaper, the majority of buyers 
value the object $B$ higher. At this moment, the salesperson mentions that there is 
a third object $C$, which is of about the same quality as $A$, but of a much higher 
price than $A$. This causes the buyer to reconsider the choice between the objects 
$A$ and $B$, while the object $C$, having practically the same quality as $A$ but being 
much more expensive, is of no interest. Choosing now between $A$ and $B$, the majority 
of buyers prefer the higher quality but more expensive object $A$. The object $C$, being 
not a choice alternative, plays the role of a decoy. Experimental studies confirmed the 
decoy effect for a variety of objects: cars, microwave ovens, shoes, computers, 
bicycles, beer, apartments, mouthwash, etc. The same decoy effect also exists in 
the choice of human mates distinguished by attractiveness and sense of humor 
\cite{Bateson_2006}. It is common as well for animals, for instance, in the choice 
of female frogs of males with different attraction calls characterized either by 
low-frequency and longer duration or by faster call rates \cite{Lea_2015}.

The decoy effect contradicts the regularity axiom in decision making telling 
that if $B$ is preferred to $A$ in the absence of $C$, then this preference 
has to remain in the presence of $C$. In the frame of affective decision theory, 
the decoy effect is explained as follows. Assume buyers consider an object $A$, 
which is of higher quality but more expensive, and an object $B$, which is of 
moderate quality but cheaper. Since the quality of the objects is close, both 
$A$ and $B$ look equally attractive from that point of view, so that the buyers 
evaluate only the object utilities, that is their prices. Suppose the buyers 
have evaluated these objects, $A$ and $B$, being based on the initial values of 
the objects described by the utility factors $f(A)$ and $f(B)$. In experiments, 
the latter correspond to the fractions of buyers evaluating higher the related 
object. When the decoy $C$, of high quality but essentially more expensive, is 
presented, it attracts the attention of buyers to the quality characteristic, 
thus classifying the objects by their attractiveness based on quality. The role 
of the decoy is well understood as attracting the attention of buyers to a 
particular feature of the considered objects, because of which the decoy effect 
is sometimes named the attraction effect \cite{Simonson_1989}. In the present case, 
the decoy attracts the buyer attention to quality. The attraction, induced by the 
decoy, is described by the attraction factors $q(A)$ and $q(B)$. Hence the 
probabilities of the related choices are now $p(A)=f(A)+q(A)$ and $p(B)=f(B)+q(B)$. 
Since the quality feature becomes more attractive, we have $q(A) > q(B)$. According 
to the non-informative prior, we can estimate the attraction factors as $q(A)=0.25$ 
and $q(B)=-0.25$.

To be more precise, let us take numerical values from the experiment of Ariely 
and Wallsten \cite{Ariely_1995}, where the objects under sale are microwave ovens. 
The evaluation without a decoy results in $f(A)=0.4$ and $f(B)=0.6$. In the presence 
of the decoy, the choice probabilities can be evaluated as $p(A)=f(A)+0.25$ and 
$p(B) = f(B)-0.25$. This gives $p(A) = 0.65$ and $p(B) = 0.35$. The experimental 
values for the choice between $A$ and $B$, in the presence of (but excluding) $C$, 
correspond to the fractions $p_{exp}(A) = 0.61$ and $p_{exp}(B) = 0.39$, which is 
close to the predicted probabilities.

Another example can be taken from the studies of the frog mate choice \cite{Lea_2015},
where frog males have attraction calls differing in either low-frequency sound or 
call rate. The males with lower frequency calls are denoted as $A$, while those with 
high call rate, as $B$. In an experiment with $80$ frog females, without a decoy, it 
was found that females evaluate higher the fastest call rate, so that $f(A) = 0.35$ 
and $f(B) = 0.65$. In the presence of an inferior decoy, attracting attention to the 
low-frequency characteristic, the non-informative prior predicts the probabilities
$p(A) = 0.35 + 0.25 = 0.6$  and $p(B) = 0.65 - 0.25 = 0.4$. The empirically observed 
fractions are found to be $p_{exp}(A) = 0.6$ and $p_{exp}(B) = 0.4$, in remarkable
agreement with our predictions \cite{Yukalov_56,Yukalov_123}.

\subsection{Planning paradox}

In classical decision theory, there is the principle of dynamic consistency 
according to which a decision taken at one moment of time should be invariant in 
time, provided no new information has become available and all other conditions 
are not changed. Then a decision maker, preferring an alternative at time $t_1$ 
should retain the choice at a later time $t_2 > t_1$. However this principle is 
often neglected, which is called the effect of dynamic inconsistency.

A stylized example of dynamic inconsistency is the planning paradox, when a 
subject makes a plan for the future, while behaving contrary to the plan as 
soon as time comes to accomplish the latter. A typical case of this inconsistency 
is the stop-smoking paradox \cite{Benfari_1982,Westmass_2010}. A smoker, well 
understanding damage to health caused from smoking, plans to stop smoking in the 
near future, but time goes, future comes, however the person does not stop smoking. 
Numerous observations \cite{Benfari_1982,Westmass_2010} show that $85\%$ of smokers 
do plan to stop smoking in the near future, however only $36\%$ really stop smoking 
during the next year after making the plan. 

Taking into account emotions makes the paradox easily resolved by including into 
the consideration the classification of alternative attractiveness. The point is 
that there are actually two types of the attractiveness classifications. One kind 
of attractiveness is based on the advantage for health, while the other kind 
evaluates the attractiveness according to the suffering when quitting the smoking 
addiction \cite{Yukalov_71,Yukalov_139}.    

Let us denote by $A_1$ the alternative to stop smoking in the near future, and by 
$A_2$, the alternative not to stop smoking. And let us denote by $B_1$ the decision 
to really stop smoking, while by $B_2$, the decision of refusing to really stop 
smoking. According to the theory, the corresponding probabilities are
$$
p(A_1) = f(A_1) + q(A_1) \; , \qquad  p(A_2) = f(A_2) + q(A_2) \; ,
$$
$$
p(B_1) = f(B_1) + q(B_1) \; , \qquad  p(B_2) = f(B_2) + q(B_2) \; .
$$
The utility factors do not change with time, so that the utility of planning to 
stop in the near future is the same as the utility to stop in reality and the 
utility not to stop in the near future is the same as that of not stopping in 
reality, $f(A_1) = f(B_1)$ and $f(A_2) = f(B_2)$.

Planning to stop smoking, subjects understand the attractiveness of this due to 
health benefits. Hence the average attraction factors are $q(A_1) = 0.25$ and 
$q(A_2)=-0.25$. But as soon as one has to stop smoking in reality, one feels 
uneasy from the necessity to forgo the pleasure of smoking, suffering instead 
from the smoking addiction. Because of this, the attraction factors are now 
$q(B_1)=-0.25$ and $q(B_2)=0.25$. This is to be complemented by the normalization 
conditions
$$
p(A_1) + p(A_2) = 1 \; , \qquad p(B_1) + p(B_2) = 1 \; ,
$$
$$
f(A_1) + f(A_2) = 1 \; , \qquad f(B_1) + f(B_2) = 1 \; .
$$

Since $85\%$ of subjects plan to stop smoking, we have $p(A_1)=0.85$ and 
$p(A_2)=0.15$. Solving the above equations, the fraction of subjects who will 
really stop smoking is predicted to be $p(B_1)=0.35$ and $p(B_2)=0.65$. This 
is in beautiful agreement with the observed fraction of smokers really stopping 
smoking during the next year after taking the decision, $p_{exp}(B_1)=0.36$ and 
$p_{exp}(B_2)=0.64$ \cite{Benfari_1982,Westmass_2010}. Thus, knowing only the 
percentage of subjects planning to stop smoking, it is straightforward to predict 
the fraction of those who will stop smoking in reality. This case is also of 
interest because it gives an example of preference reversal: When planning to 
stop smoking, the relation between the probabilities is reversed as compared 
to the relation between the fractions of those who have really stopped smoking,
$p(A_1) > p(A_2)$, while $p(B_1) < p(B_2)$.

\subsection{Preference reversal}

Preference reversal happens, when one compares two alternatives that are treated 
as attractive with respect to two different characteristics. For example, let us 
consider two lotteries $L_1$ and $L_2$ evaluating them from the point of view of 
choosing the stochastically optimal one. Suppose the optimal lottery is $L_1$, 
such that $p(L_1)>p(L_2)$. Then let us compare the same lotteries, but keeping 
in mind pricing them in order to sell. Not to confuse the tasks, let us rename 
the lotteries to be priced as $L_3\sim L_1$ and $L_4\sim L_2$. The utility factors 
for the lotteries do not change, so that $f(L_1)=f(L_3)$ and $f(L_2)=f(L_4)$. What 
changes is the decision maker classification of the lottery attractiveness, as a 
result of which the inequality $p(L_1)>p(L_2)$ reverses to $p(L_3)<p(L_4)$. 

To illustrate the reversal effect, let us consider the example given by Tversky 
and Thaler \cite{Tversky_Thaler_1990}, considering two lotteries
$$
  L_1 = \left\{ 4 , \; \frac{8}{9}\; | \; 0 , \; \frac{1}{9} \right\} \; ,
\qquad 
 L_2 = \left\{ 40 , \; \frac{1}{9}\; | \; 0 , \; \frac{8}{9} \right\} \; ,
$$
and interpreting the case using affective decision theory \cite{Yukalov_129}. 
The related utility factors are $f(L_1)=0.444$ and $f(L_2)=0.556$, so that the 
first lottery is less useful. However the quality functionals (\ref{3.51}) are 
$Q(L_1)=82.2$ and $Q(L_2)=58.4$. Therefore, when choosing, the first lottery is 
classified as more attractive, which implies the non-informative priors for the 
attraction factors $q(L_1)=0.25$ and $q(L_2)=-0.25$. Then the predicted behavioural 
probabilities are $p(L_1)=0.694$ and $p(L_2)=0.306$, which is close to the empirical 
data $p_{exp}(L_1)=0.71$ and $p_{exp}(L_2)=0.29$ found in \cite{Tversky_Thaler_1990}.

At the same time, when pricing these lotteries, one evaluates them more objectively,
without involving much emotions, which assumes $p(L_3)\approx f(L_3)$ and 
$p(L_3)\approx f(L_3)$, which results in the inequality $p(L_3)<p(L_4)$. In that 
way, when choosing, one has the inequality $p(L_1)>p(L_2)$ that reverses when these 
lotteries are priced yielding $p(L_3)<p(L_4)$. Many other similar examples are 
listed in \cite{Tversky_Thaler_1990} and analyzed in \cite{Yukalov_129}.

\subsection{Preference intransitivity}

Transitivity is of central importance to both psychology and economics. It is 
the cornerstone of normative and descriptive decision theories \cite{Neumann_1953}. 
Individuals, however, are not perfectly consistent in their choices. When faced 
with repeated choices between alternatives $A$ and $B$, people often choose in 
some instances $A$ and $B$ in others. Such inconsistencies are observed even in 
the absence of systematic changes in the decision maker's taste, which might be 
due to learning or sequential effects. The observed inconsistencies of this type 
reflect inherent variability or momentary fluctuations in the evaluative process. 

There also occur several choice situations where transitivity may be violated 
because of the inconsistent account of emotions, which is called preference 
intransitivity \cite{Tversky_1969,Fishburn_1988,Fishburn_1991,Panda_2018}. 
The occurrence of preference intransitivity depends on the considered decision model 
and on the accepted definition of transitivity. Its general meaning is as follows. 
Suppose one evaluates three alternatives $A$, $B$, and $C$, considering them in turn 
by pairs. However, in the process of comparison, one does not follow a particular
utility model, but rather applies behavioural reasoning including emotions. Thus one 
compares $A$ and $B$ comparing their utilities and concludes that, say, $A$ is 
preferred over $B$, which can be denoted as $A\succ B$. Then one compares $B$ and 
$C$, again using the comparison of the related utilities, finding that $B\succ C$. 
Finally, comparing $C$ and $A$, in addition to utility, one takes into account their 
attractiveness and discovers that $C\succ A$. This seems to result in the preference 
loop $A \succ B \succ C \succ A$ signifying the intransitivity effect.

As an illustration of the intransitivity effect, we can adduce the Fishburn example 
\cite{Fishburn_1991}, where a person is about to change jobs. When selecting a job, 
the person evaluates the suggested salary, which defines the job utility. In 
addition, the jobs differ by their attractiveness connected with the prestige of 
the position. There are three choices: job $A$, with the salary $65,000\$ $, but 
low prestige, job $B$, with the salary $58,000\$ $, and medium prestige, and job 
$C$, with the salary $50,000\$ $ and very high prestige. The person chooses $A$ 
over $B$ because of the better salary, and with a small difference in prestige, 
$B$ over $C$ because of the same reason, and comparing $C$ and $A$, the person 
prefers $C$ being attracted by an essentially higher prestige, although a lower 
salary. Thus, one comes to the preference loop $A \succ B \succ C \succ A$ considered 
as a paradox. 

However, we do not meet any paradox taking into account both utility and 
attractiveness. The utility factors can be calculated as described in Sec. 3. 
Thus, at the first step, considering the pair $A$ and $B$ in the Fishburn 
example, we have the utility factors $f_1(A) = 0.528$ and $f_1(B) = 0.472$. 
Due to the close level of prestige of the both jobs, their attraction factors 
coincide, which, as follows from the alternation law, gives $q_1(A)=q_1(B)=0$. 
This leads to the behavioral probabilities $p_1(A)=0.528$ and $p_1(B)=0.472$. 
Since $p_1(A) > p_1(B)$, the job $A$, at this stage, is stochastically preferred 
over $B$.

Then, at the second step, comparing the jobs $B$ and $C$, we get the utility 
factors $f_2(B)=0.537$ and $f_2(C)=0.463$, and the attraction factors 
$q_2(B)=q_2(C)=0$. Hence, the probabilities are $p_2(B)=0.537$ and $p_2(C)=0.463$, 
which implies that $B$, at that stage, is stochastically preferred over $C$.

In the comparison of the jobs $C$ and $A$, at the third step, we find the utility 
factors $f_3(C) = 0.435$ and $f_3(A) = 0.565$. Now the positions are of a very 
different quality, with the job $C$ being much more attractive, so that the 
non-informative priors for the attraction factors are $q_3(C) = 0.25$ and 
$q_3(A) = -0.25$. Therefore, the probabilities become $p_3(C) = 0.685$ and 
$p_3(A) = 0.315$. Hence, at this stage, the job $C$ is stochastically preferred 
over $A$.

Stochastic preference implies an inequality between behavioural probabilities.
At the first step, $A$ is preferred over $B$, since $p_1(A) > p_1(B)$. At the 
second step, $B$ is preferred to $C$, because $p_2(B)>p_2(C)$. At the third step, 
$C$ is preferred over $A$, as far as $p_3(C)>p_3(A)$. However, it follows from 
nowhere that $p_3(C)$ is prohibited from being larger than $p_3(A)$. 

Thus, when selecting a job, the person evaluates the alternatives taking 
account of both, utility suggested by the salary, and the attractiveness, due 
to the prestige of the position. Therefore, there is nothing extraordinary that 
at different steps, differently defined probabilities can be intransitive. Even 
more, there are arguments \cite{Tsetsos_2016} that such intransitivities can be 
advantageous for alive beings in the presence of irreducible noise during neural 
information processing. In any case, there is no any paradox, when all 
alternatives are judged in the frame of the approach taking into consideration 
their utility as well as attractiveness.

\subsection{Order effects}

Order effects refer to differences in research participants' responses that result 
from the order in which the experimental materials or questions are presented to 
them. Order effects can occur in any kind of research. In survey research, for 
example, people may answer questions differently depending on the order in which 
the questions are asked \cite{Shaughnessy_2006}.

Sometimes, one speculates that the order dependence is the result of some kind 
of quantumness in the brain functioning, so that the order dependence is somehow 
connected with the noncommutativity of operators representing events. However, 
there is nothing quantum in the order effects. It is only sufficient to remember 
that different consecutive questions are asked at different times, that is under 
different conditions. The simplest reason of the differences, e.g., could be just 
because participants could perform differently at the end of an experiment or 
survey, being bored or tired. Considering the symmetry properties of behavioural 
probabilities, it is also useful to remember that, strictly speaking, emotions 
are contextual and can vary in time. 

Speculating about quantum conscious effects, one usually confuses, which of the 
quantum probabilities one keeps in mind and why. In order to recollect the 
differences between the different types of quantum probabilities, let us consider 
two sets of events, $\{A_n\}$ and $\{B_k\}$. These events can be accompanied by 
intrinsic noise in the case of physics experiments, or by emotions in the case of 
human decision making. Assuming the presence of noise or emotions, we will not expose 
them explicitly to make the formulas simpler.   

In Sec. 2, it is explained that we can define a kind of a joint quantum probability 
$p(B_k,t,A_n,t_0)$ of an event, say $A_n$, happening at the moment of time $t_0$, after 
which an event $B_k$ occurs at time $t$. The related conditional probability 
$p(B_k,t|A_n,t_0)$ describes the occurrence of an event $B_k$ at time $t$ under the 
condition that the event$A_n$ has certainly happened at time $t_0$. Both these 
probabilities are not symmetric with respect to the interchange of the events, 
\be
\label{4.1}
p(B_k,t,A_n,t_0) \neq p(A_n,t,B_k,t_0) \; , \qquad 
p(B_k,t|A_n,t_0) \neq p(A_n,t,B_k,t_0) \qquad ( t > t_0 ) \;   ,
\ee
where $t > t_0$.

The joint probabilities of immediate consecutive events are not order-symmetric, 
while the conditional probabilities for these events are symmetric,
\be
\label{4.2}
 p(B_k,t_0+0,A_n,t_0) \neq p(A_n,t_0+0,B_k,t_0) \; , \qquad 
p(B_k,t_0+0|A_n,t_0) = p(A_n,t_0+0|B_k,t_0) \; .
\ee
  
It is also possible to define the joint quantum probabilities of events occurring 
in different spaces, for instance related to different spatial locations. These 
probabilities enjoy the same symmetry property as classical probability, being order
symmetric,
\be
\label{4.3}
p(A_n B_k,t) = p(B_k A_n,t) \; .
\ee

Strictly speaking, the usual process of taking decisions is not momentary, but 
requires finite time. The modern point of view accepted in neurobiology and 
psychology is that the cognition process, through which decisions are generated, 
involves three stages: the process of stimulus encoding through which the internal 
representation is generated, followed by the evaluation of the stimulus signal and 
then by decoding of the internal representation to draw a conclusion about the 
stimulus that can be consciously reported \cite{Woodford_2020,Libert_2006}. It has 
been experimentally demonstrated that awareness of a sensory event does not appear 
until the delay time up to $0.5$ s after the initial response of the sensory cortex 
to the arrival of the fastest projection to the cerebral cortex 
\cite{Libert_2006,Teichert_2016}. About the same time is necessary for the process 
of the internal representation decoding. So, the delay time of about $1$ s is the 
minimal time for the simplest physiological processes involved in decision making. 
Sometimes the evaluation of the stimulus signal constitutes the total response time, 
necessary for formulating a decision, of about $10$ s \cite{Hochman_2010}. In any 
case, the delay time of order $1$ s seems to be the minimal period of time required 
for formulating a decision. This assumes that in order to consider a sequential 
choice as following immediately after the first one, as is necessary for the quantum 
immediate consecutive probabilities, the second decision has to follow in about $1$ s 
after the first choice. However, to formulate the second task needs time, as well as 
some time is required for the understanding of the second choice problem. This 
process demands at least several minutes. 

In this way, the typical situation in the sequential choices is when the minimal 
temporal interval between the decisions is of the order of minutes, which is 
much longer than the time of $1$ s necessary for taking a decision. Therefore 
the second choice cannot be treated as following immediately after the first one, 
hence the form of the immediate conditional probability is hardly applicable to 
such a situation. For that case, one has to use the different-time probabilities 
which are not order symmetric, in agreement with empirical observations.   

Thus the decisions can be considered as following immediately one after the other 
provided the temporal interval between them is of the order of $1$ s. Such a short 
interval between subsequent measurements could be realized in quantum experiments, 
but it is not realizable in human decision making, where the interval between 
subsequent decisions is usually not less than several minutes. Mathematically, 
the different-time probabilities are not order-symmetric, whether they are quantum 
or classical. The explanation of order effects in human decision making does not 
require any use of quantum probabilities. 
 
The temptation to resort to quantum theory comes from the inappropriate 
comparison of classical probability theory in the Kolmogorov picture, where 
there is no time, with quantum theory, where there exists dynamics prescribed 
by the evolution operators. However, it is not correct to compare quantum 
probabilities of consecutive events with classical probabilities of synchronous 
events. One has to compare the probabilities of the same class, either from 
the class of consecutive probabilities of from the class of synchronous 
probabilities.  

Order effects in real life arise not because of noncommutativity of some 
operators, but because two different consecutive events occur at different 
times. Mathematically, this implies that the evolution equations for swapped 
events can be different and are characterized by different initial conditions. 
Then it is of no surprise that different equations and initial conditions 
result in different solutions, whether in quantum or classical picture. 
  
Explicit illustration that classical theory does contain order effects, and 
for their explanation no quantum theory is required, has been done in Sec. 2.17 
considering two sets of events, $\{A_n\}$ and $\{B_k\}$. In classical picture, 
the conditional probability $f(B_k,t|A_n,t_0)$ of an event $B_k$ occurring at 
a time $t$, after an event $A_n$ has happened at the time $t_0$, can be 
characterized by the Kolmogorov forward equation that yields a different solution 
for the probability $f(A_n,t|B_k,t_0)$ with the inverted order of the events.

In that way, both the joint probabilities of consecutive events, quantum as well 
as classical, are not symmetric with respect to the event swapping. The classical
probability of consecutive events does contain order effects, so that there is no 
any advantage of using quantum probabilities for explaining order effects in human 
decision making.

\section{Networks of intelligent agents}

Intelligent Agent is a system that, evaluating the available information, is 
able to take autonomous actions and decisions directed to the achievement of 
the desired goals and may improve its performance by using obtained knowledge 
\cite{Nilsson_1,Luger_2,Rich_3,Poole_4,Neapolitan_5,Russell_6}. The intelligent 
agent notion is employed with respect to the agents possessing artificial 
intelligence as well as to the agents in economics, in cognitive science, ethics, 
philosophy, and in many interdisciplinary socio-cognitive modeling and simulations. 
Generally, from the technical and mathematical point of view, the notion of intelligent 
agent can be associated with either real or artificial intelligence. An intelligent 
agent could be anything that makes decisions, as a person, firm, machine, or software. 
In addition, we keep in mind intelligent agents subject to emotions. Several 
intelligent agents compose a society or a network of intelligent 
agents \cite{Perc_2013,Perc_2017,Jusup_2021}. In a particular case, representing 
each neuron as an intelligent agent gives the model of the brain as a network of 
intelligent agents. In this section, networks of intelligent agents are considered,
which make multiple repeated decisions and interact with each other by exchanging 
information.

\subsection{Multistep decision making}

The main distinguishing feature of an intelligent agent is the ability to make 
decisions employing the available information. The society of intelligent agents 
interacting with each other through information exchange forms a network. Let us 
consider a family $\{A_n\}$ of $N_A$ alternative actions enumerated by the index 
$n=1,2,\ldots, N_A$. A society of intelligent agents consists of $N$ agents counted 
by the index $j=1,2,\ldots,N$. Each agent has to make a decision, choosing one of 
the alternatives from the given set $\{A_n\}$. The probability of choosing an 
alternative $A_n$ by a $j$-th agent at the moment of time $t$ is denoted as $p_j(A_n,t)$. 
We keep in mind behavioural probabilities taking into account the utility of 
alternatives as well as emotions accompanying the choice, although, for brevity, this 
is not explicitly shown. The probabilities are normalized,
\be
\label{5.1}
 \sum_{n=1}^{N_A} p_j(A_n,t) = 1 \; , \qquad 0 \leq p_j(A_n,t) \leq 1 \; .
\ee

The utility of alternatives for a $j$-th agent is described by the utility factors 
$f_j(A_n,t)$ satisfying the normalization
\be
\label{5.2}
\sum_{n=1}^{N_A} f_j(A_n,t) = 1 \; , \qquad 0 \leq f_j(A_n,t) \leq 1 \; .
\ee
Emotions are characterized by a separate term that can be split into two parts 
responsible for different types of emotions. One part of emotions describes the 
attractiveness of each alternative by the attraction factor $q_j(A_n,t)$ that can 
be positive or negative, being in the interval
\be
\label{5.3}
-1 \leq q_j(A_n,t) \leq 1 \; .
\ee
The other part of emotions is related to the well known human feature of mimicking 
the behaviour of others. This part is represented by the herding factor $h_j(A_n,t)$ 
varying in the interval
\be
\label{5.4}
-1 \leq h_j(A_n,t) \leq 1 \;   .
\ee
     
The structure of the network can be understood by generalizing the behavioural 
probabilities of separate agents to the collection of intelligent agents 
\cite{Yukalov_133,Yukalov_134,Yukalov_135}. Similarly to a single decision maker, 
we do not mean that the network has to be composed of quantum objects. As for a 
single decision maker, the model can be formulated in purely classical terms 
\cite{Yukalov_135}. 

As has been mentioned earlier, the process of decision making needs time that can 
be denoted by $\tau$. Thus the probability of a $j$-th agent to choose an alternative 
$A_n$ at time $t+\tau$, after the information on utility, attractiveness, and 
herding, that has been received at time $t$, has the form
\be
\label{5.5}
 p_j(A_n,t+\tau) = f_j(A_n,t) + q_j(A_n,t) + h_j(A_n,t) \; .
\ee

The normalization conditions (\ref{5.1}) and (\ref{5.2}) lead to the condition
\be
\label{5.6}
 \sum_{n=1}^{N_A} [\; q_j(A_n,t) + h_j(A_n,t) \; ] = 0 \; .
\ee
Assuming that the properties of attractiveness and herding are independent, we 
can write
\be
\label{5.7}
 \sum_{n=1}^{N_A}  q_j(A_n,t) = 0 \; , \qquad 
\sum_{n=1}^{N_A}  h_j(A_n,t) = 0 \; .
\ee
These equations are called the alternation law. 

At rather long time scales, the utility factor may have temporal discounting 
\cite{Frederick_2002}. But at time scale shorter than the discounting scale, 
the utility factor can be treated as constant,
\be
\label{5.8}
  f_j(A_n,t) =  f_j(A_n) \; .
\ee
 
On the contrary, the information exchange between the agents is a fast process 
depending on the amount of information obtained by each agent. The temporal 
behaviour of the attraction factor imitates the decoherence term in the observable 
quantities caused by nondestructive repeated measurements 
\cite{Yukalov_71,Yukalov_2011,Yukalov_136,Yukalov_137} and can be modeled 
\cite{Yukalov_128} by the expression
\be
\label{5.9}
  q_j(A_n,t) = q_j(A_n) \exp\{ - M_j(t) \} \; ,
\ee
where $q_j(A_n)$ is the value of the attraction factor of an agent $j$ who possesses 
no information from other agents, and $M_j(t)$ is {\it memory}. 
The amount of information obtained by the $j$-th agent by the time $t$ is 
\be
\label{5.10}
 M_j(t) = \sum_{t'=0}^t \; \sum_{i=1}^N J_{ji}(t,t') \mu_{ji}(t') \; ,
\ee
with $J_{ji}(t,t')$ being the intensity of the information transfer from an 
$i$-th to a $j$-th agent in the period of time from $t'$ to $t$. The information 
received by a $j$-th agent from an $i$-th agent is described by the Kullback-Leibler 
information gain
\be
\label{5.11}
 \mu_{ji}(t) = \sum_{n=1}^{N_A} p_j(A_n,t) 
\ln \; \frac{p_j(A_n,t)}{p_i(A_n,t)} \;  .
\ee

To proceed further, one needs to specify the interaction intensity and the type 
of the agent's memory.

\subsection{Types of interactions and memory}

The interaction intensity $J_{ji}(t,t')$ depends on the distance between the agents. 
For short-range interactions, the spatial network topology is important. However, if 
one considers a modern human society, then the location of agents is not fixed and can 
vary with time, since the agents can freely move. They also can interact at any distance 
through the contemporary means of information exchange such as phones, internet, mass 
media, etc. This society is characterized by the long-range distance-independent 
interactions
\be
\label{5.12}
 J_{ij}(t,t') = \frac{1}{N-1} \; J(t,t') \qquad (long-range) \; .
\ee
In that case, the agent's memory takes the form
\be
\label{5.13}
 M_j(t) = \sum_{t'=0}^t \frac{J(t,t')}{N-1} 
\sum_{i=1}^N \mu_{ji}(t')  \qquad (long-range)  \; .
\ee

The period of time during which the memory is kept by an agent describes the type 
of the memory. In two opposite situations, there can be long-term memory, when  
\be
\label{5.14}
 J(t,t') = J \qquad (long-term) \; ,
\ee
so that the amount of the received information kept in the memory is 
\be
\label{5.15}
 M_j(t) = \frac{J}{N-1} \sum_{t'=0}^t \sum_{i=1}^N \mu_{ji}(t') \qquad (long-term) \; ,
\ee  
and short-term memory, when only the local information is appreciated,
\be
\label{5.16}
 J(t,t') = J \dlt_{tt'} \qquad (short-term) \;  .
\ee
In the latter case, the available information in the memory becomes
\be
\label{5.17}
  M_j(t) = \frac{J}{N-1} \sum_{i=1}^N \mu_{ji}(t) \qquad (short-term) \; .
\ee
In numerical calculations, we set $J=1$.
    
In societies, there exists a collective effect, when the agents are prone to 
imitate the actions of others, which is called herding effect. Herd behaviour 
occurs in animal herds, packs, bird flocks, fish schools and so on, and it is very 
widely spread in humans. The herding effect is well known and studied for many years 
\cite{Martin_1920,Sherif_1936,Smelser_1963,Merton_1968,Turner_1993,Hatfield_1994,
Brunnermeir_2001}. The mathematical description of herding in evolution equations 
of social and biological systems is represented by the replication 
term \cite{Sandholm_2010}, which we take in the form
\be
\label{5.18}
h_j(A_n,t) = \ep_j\left\{ \frac{1}{N-1} 
\sum_{i(\neq j)}^N [\; f_i(A_n) + q_i(A_n,t) \; ] -
 [\; f_j(A_n) + q_j(A_n,t) \; ] \right\} \; .
\ee
The strength of the herding behaviour is given by the parameters $\ep_j$. Because of 
the normalization conditions (\ref{5.1}), (\ref{5.2}), (\ref{5.3}), and (\ref{5.4}), 
these parameters are in the interval
\be
\label{5.19}
0 \leq \ep_j \leq 1 \qquad ( j = 1,2,\ldots,N) \;  .
\ee

Substituting the herding term into equation (\ref{5.5}) and measuring time in 
units of $\tau$ yields the delayed evolution equations
\be
\label{5.20}
 p_j(A_n,t+1) = ( 1 - \ep_j)  [\; f_j(A_n) + q_j(A_n,t) \; ] + 
\frac{\ep_j}{N-1}  \sum_{i(\neq j)}^N [\; f_i(A_n) + q_i(A_n,t) \; ] 
\ee
defining the probability that the $j$-th agent at time $t+1$ chooses the 
alternative $A_n$. 

For negative times, the utility factor is fixed, $f_j(A_n,t<0) = f_j(A_n)$, 
there is no yet any memory, $M_j(t<0) = 0$, hence the attraction factor is 
$q_j(A_n,t<0) = q_j(A_n)$. Therefore, at zero time, the initial condition for 
the probability is
\be
\label{5.21}
  p_j(A_n,0) = ( 1 - \ep_j)  [\; f_j(A_n) + q_j(A_n) \; ] + 
\frac{\ep_j}{N-1}  \sum_{i(\neq j)}^N [\; f_i(A_n) + q_i(A_n) \; ] \; .
\ee
At the first step, equation (\ref{5.20}) gives $p_j(A_n,1)$, in which 
$q_j(A_n,0) = q_j(A_n) \exp\{- M_j(0) \}$, etc.  

The evolution equations describe the dynamics of collective decision making 
developing in time due to the interaction of agents through information exchange. 
The dynamics of opinions in choosing among alternatives depends on the utility 
of alternatives, their emotional attractiveness, and herding tendency of the agents.

\subsection{Networks with uniform memory}

Under the fixed type of interactions, the society agents can differ in their 
initial conditions $f_j(A_n)$ and $q_j(A_n)$, in the strength of the herding effect, 
and in the kind of the agent memory, whether long-term or short-term. There can exist 
networks, where all agents enjoy the same type of memory, for instance all agents 
have long-term memory or all agents possess short-term memory, but the agents at the 
initial moment of time are characterized by different initial conditions. This case 
of agents with the same type of memory but different initial conditions is studied 
in \cite{Yukalov_133,Yukalov_134}. A network, where all agents have the same type of 
memory, can be called uniform-memory network. Despite that all agents in the 
uniform-memory society have the same type of memory, they can display nontrivial 
dynamics due to different initial conditions. The case of two groups of agents making 
a choice between two alternatives has been investigated \cite{Yukalov_133,Yukalov_134}. 
The groups differ by their initial conditions. Two situations have been studied. 

\vskip 2mm
(i) {\it Network with uniform long-term memory}. This network always displays the 
convergence of behavioural probabilities to fixed points. Depending on the initial 
conditions and herding strength, the dynamics can be monotonic or with intermediate 
oscillations, the fixed points can coincide or not, the probability trajectories can 
intersect, but always the probabilities tend to their fixed points. Increasing the 
herding parameters $\varepsilon_j$ makes the probability trajectories closer to each 
other, and sufficiently strong herding forces the probabilities to converge to a 
consensual limit.  

\vskip 2mm
(ii) {\it Network with uniform short-term memory}. This network displays two kinds of 
behaviour. One kind is the convergence of the behavioural probabilities to fixed points 
that, under week herding effect, do no coincide, but start coinciding for sufficiently 
strong herding parameters. The other kind of behaviour of such short-term memory networks 
is the existence of permanent oscillations of both probabilities. These oscillations become
suppressed when the herding parameters are increased.

\subsection{Network with mixed memory}
 
The most interesting and rich case is when the society contains the groups of agents 
with different types of memory and where all three features, utility, attractiveness, 
and herding are taken into account \cite{Yukalov_135}. This case also looks as more 
realistic, since in any society there always exist agents with different types of 
memory. A network, where a part of agents has long-term memory, while the other 
part, short-term memory, can be named {\it mixed-memory network}.  

For concreteness, let us consider the often met situation of choice between two 
alternatives, $A_1$ and $A_2$. The overall society is composed of a group of $N_1$ 
agents with long-term memory and $N_2$ agents with short-term memory. The total 
number of agents is the sum of the agents in each of the groups, $N=N_1+N_2$. The 
number of agents in a $j$-th group preferring an alternative $A_n$ is $N_j(A_n,t)$ 
depending on time and being normalized,
\be
\label{5.22}   
 N_j(A_1,t) + N_j(A_2,t) = N_j \qquad ( j = 1,2) \;  .
\ee
Then 
\be
\label{5.23}
p_j(A_n,t) = \frac{N_j(A_n,t)}{N_j} \qquad ( j = 1,2)
\ee
is the frequentist probability of choosing an alternative $A_n$ by a member of 
the $j$-th group at time $t$.

For the sake of convenience, it is possible to use the simplified notations for the 
probabilities
\be
\label{5.24}
 p_j(A_1,t) \equiv p_j(t) \; , \qquad p_j(A_2,t) \equiv 1 - p_j(t) \;  ,
\ee
utility factors
\be
\label{5.25}
f_j(A_1) \equiv f_j \; , \qquad f_j(A_2) \equiv 1 - f_j \;   ,
\ee
attraction factors
\be
\label{5.26}
q_j(A_1,t) \equiv q_j(t) \; , \qquad q_j(A_2,t) \equiv - q_j(t) \; ,
\ee
and herding factors
\be
\label{5.27}
 h_j(A_1,t) = h_j(t) \; , \qquad h_j(A_2,t) \equiv - h_j(t) \;  .
\ee
The evolution equations become
\be
\label{5.28}
 p_j(t+1) =f_j + q_j(t) + h_j(t) \qquad ( j = 1,2) \;  ,
\ee
with the attraction factors 
\be
\label{5.29}
q_j(t) = q_j \exp\{ - M_j(t) \} \qquad ( j =1,2 )
\ee
and herding factors
\be
\label{5.30}
 h_j(t) = \ep_j [\; f_i + q_i(t) - f_j - q_j(t) \; ] \qquad 
( i \neq j) \; ,
\ee
where $q_j$ is the attraction factor before receiving any information.

The first group, by assumption, consists of agents with long-term memory of the 
type
\be
\label{5.31}
 M_1(t) = \sum_{t'=0}^t \mu_{12}(t') \; ,
\ee
while the second group has short-term memory with
\be
\label{5.32}
 M_2(t) = \mu_{21}(t) \; .
\ee
The information gain reads as
\be
\label{5.33}
 \mu_{ij}(t) = p_i(t)  \ln\; \frac{p_i(t)}{p_j(t)}  +
[\; 1 - p_i(t)\;]  \ln\; \frac{1-p_i(t)}{1-p_j(t)} \; ,
\ee
where $i\neq j$.

The evolution equations take the form 
$$
p_1(t+1) = ( 1 - \ep_1) [\; f_1 + q_1(t) \;] + \ep_1 [\; f_2 + q_2(t) \;] \; ,
$$
\be
\label{5.34} 
p_2(t+1) = ( 1 - \ep_2) [\; f_2 + q_2(t) \;] + \ep_2 [\; f_1 + q_1(t) \;] \; ,
\ee
with the initial conditions
$$
p_1(0) = ( 1 - \ep_1) (\; f_1 + q_1 \;) + \ep_1 (\; f_2 + q_2 \;) \; ,
$$
\be
\label{5.35} 
p_2(0) = ( 1 - \ep_2) (\; f_2 + q_2 \;) + \ep_2 (\; f_1 + q_1 \;) \; .
\ee
Recall that $p_1(t)$ describes the fraction of agents with long-term memory 
choosing the alternative $A_1$ at time $t$, while $p_2(t)$ characterizes the 
fraction of agents with short-term memory preferring, at time $t$, the same 
alternative $A_1$.

\subsection{Dynamic regimes of preferences}

A detailed analysis of equations (\ref{5.34}) has been done in \cite{Yukalov_135}. 
Depending on the values of the initial conditions and herding parameters, the 
following dynamic regimes are found for the probabilities.

\vskip 2mm
1. Both probabilities tend to fixed points as in Fig. 1. The tendency can be smooth 
or with intermediate oscillations. The fixed points can coincide or not.

\begin{figure}[ht]
\centerline{
\includegraphics[width=8cm]{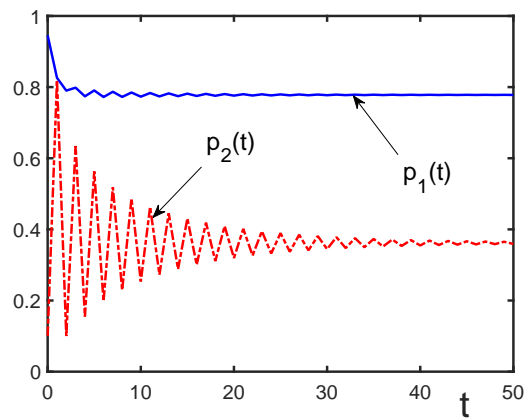}  }
\caption{\small Time dependence of the probabilities $p_1(t)$ (solid line) and 
$p_2(t)$ (dash-dotted line), for the initial conditions $f_1=0.8$, $q_1=0.19$ 
and $f_2=0.9$, $q_2=-0.8$, under weak herding effect, with the herding parameters 
$\ep_1=0.05$ and $\ep_2=0$. The probabilities tend to the fixed points 
$p_1^*=0.778$ and $p_2^*=0.362$.
} 
\label{fig:Fig.1}
\end{figure}

\vskip 2mm
2. Under sufficiently strong herding effect the dynamics is smooth and the 
probabilities are getting closer to each other, as in Fig. 2, where the herding 
parameters are larger, under the same initial conditions as in Fig. 1.

\begin{figure}[ht]
\centerline{
\includegraphics[width=8cm]{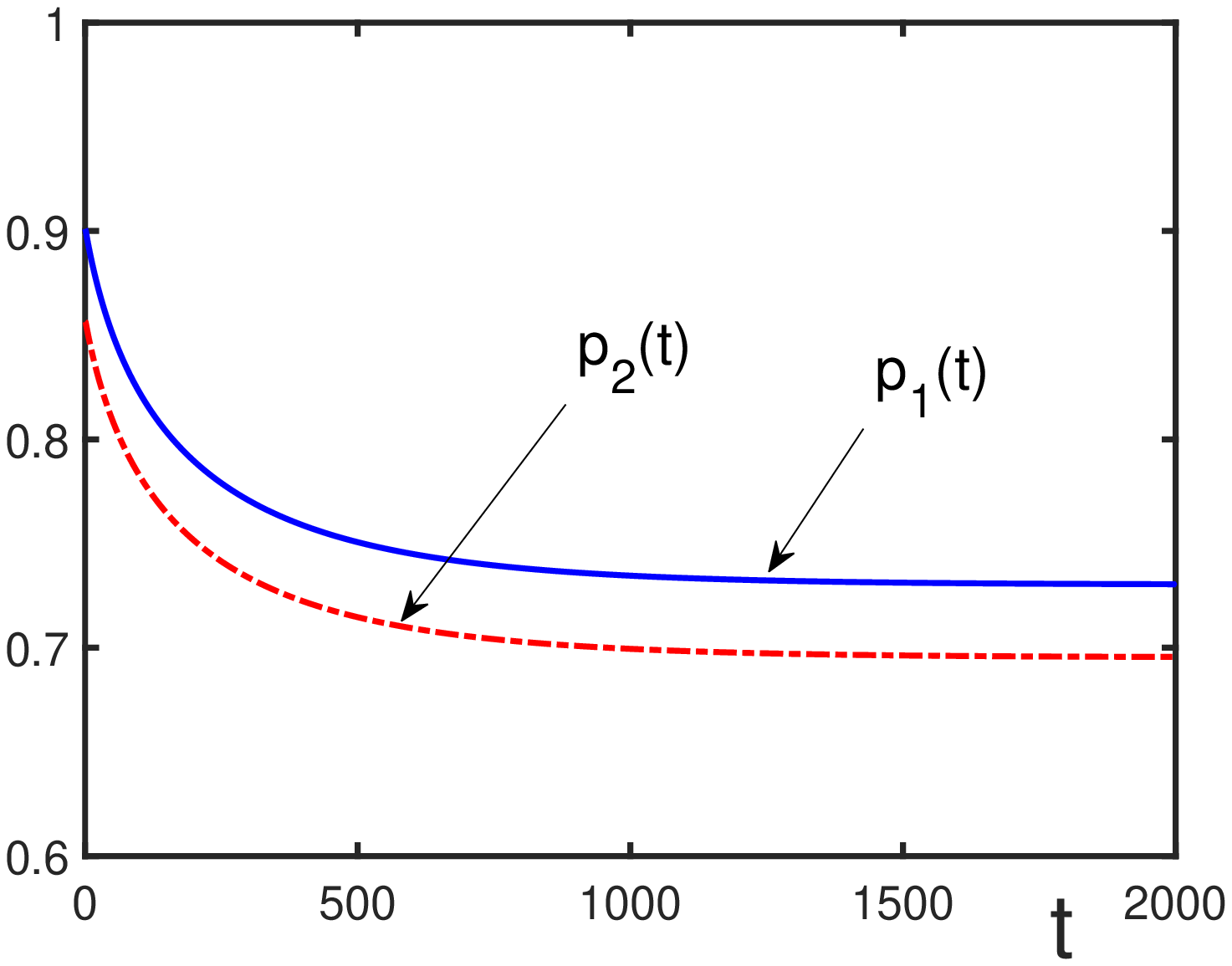}  }
\caption{\small Time dependence of the probabilities $p_1(t)$ (solid line) and 
$p_2(t)$ (dash-dotted line), for the same initial conditions, as in Fig.1, with 
$f_1=0.8$, $q_1=0.19$ and $f_2=0.9$, $q_2=-0.8$, but under stronger herding effect, 
with the herding parameters $\ep_1=0.1$ and $\ep_2=0.85$. The probabilities tend 
to the fixed points $p_1^*=0.730$ and $p_2^*=0.695$.
}
\label{fig:Fig.2}
\end{figure}

\vskip 2mm
3. One of the probabilities tends to a fixed point, while the other at some moment 
of time starts periodically oscillating and continues oscillating forever, as in 
Fig. 3.

\begin{figure}[ht]
\centerline{
\includegraphics[width=8cm]{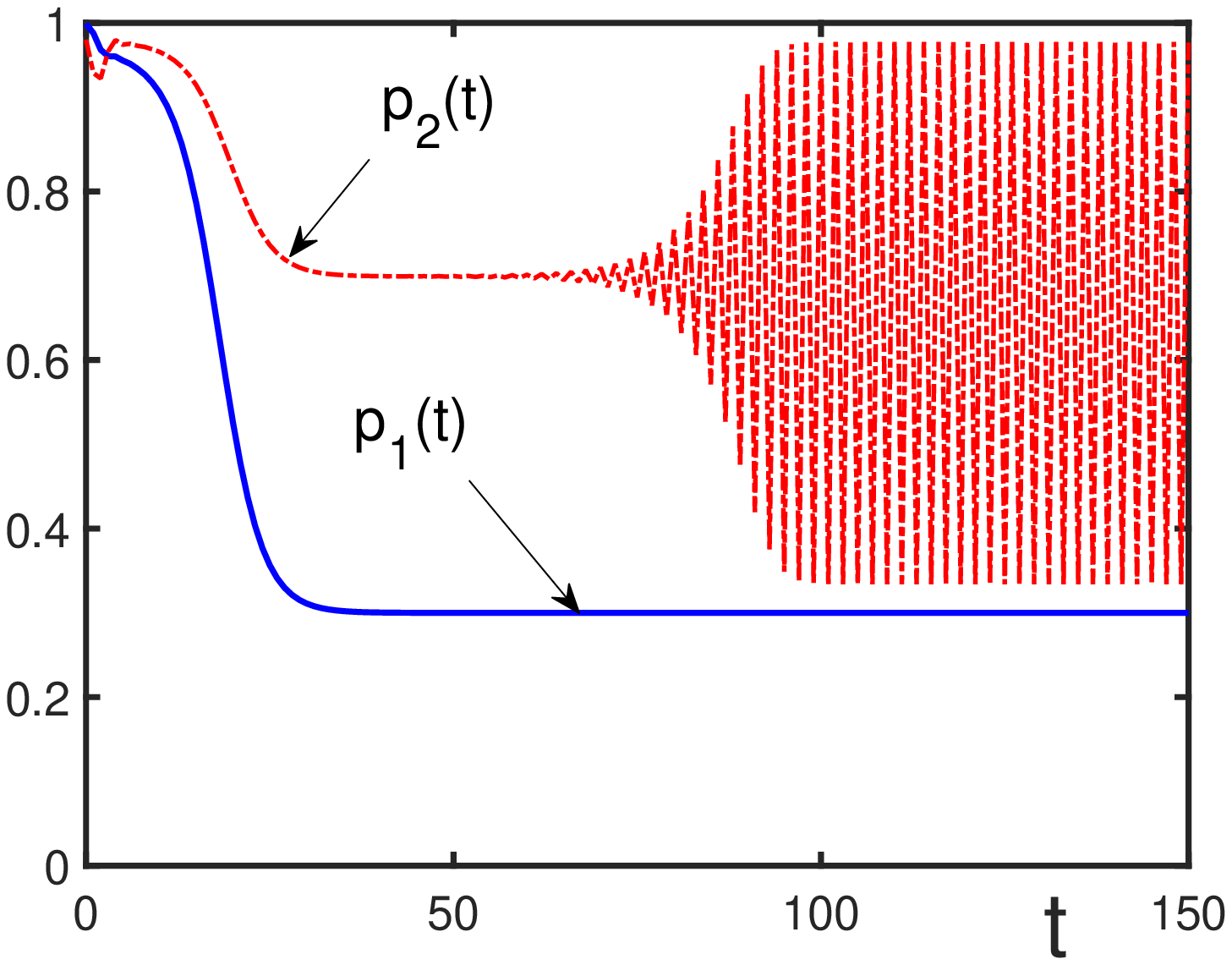}  }
\caption{\small Time dependence of the probabilities $p_1(t)$ (solid line) and 
$p_2(t)$ (dash-dotted line), for the initial conditions $f_1=0.3$, $q_1=0.699$ 
and $f_2=0$, $q_2=0.98$, in the absence of herding, with  $\ep_1=\ep_2=0$. The 
probability $p_1(t)$ tends to the fixed points $p_1^*=0.3$ while $p_2(t)$, after 
a period of smooth behaviour, starts everlasting oscillations. 
}
\label{fig:Fig.3}
\end{figure}

\vskip 2mm
4. Both probabilities exhibit everlasting periodic oscillations, as in Fig. 4.  

\begin{figure}[ht]
\centerline{
\includegraphics[width=8cm]{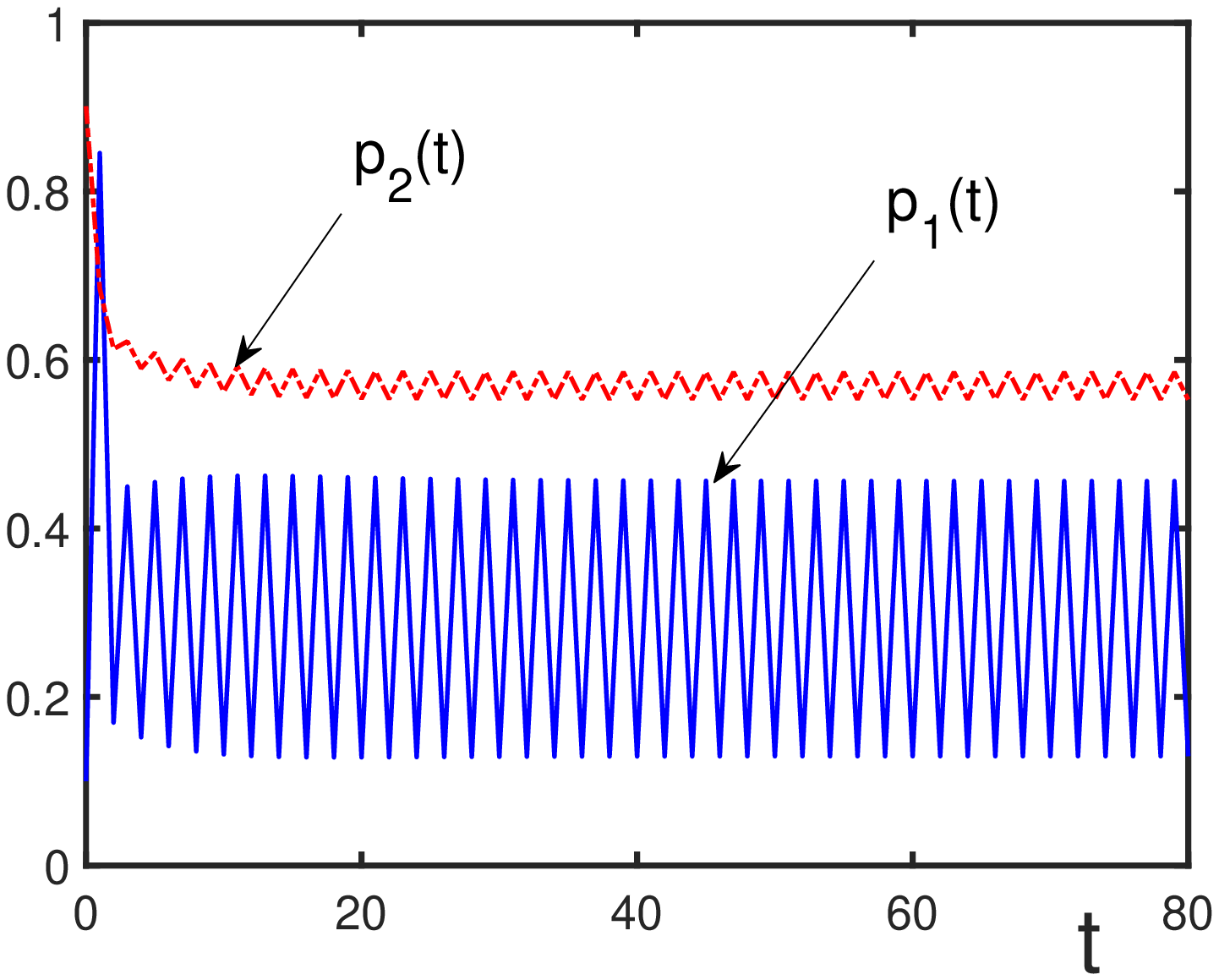}  }
\caption{\small Time dependence of the probabilities $p_1(t)$ (solid line) and 
$p_2(t)$ (dash-dotted line), for the initial conditions $f_1=0.6$, $q_1=0.39$ 
and $f_2=1$, $q_2=-0.9$, under strong herding, with the parameters $\ep_1=1$ and 
$\ep_2=0.9$. Both probabilities perpetually oscillate. 
} 
\label{fig:Fig.4}
\end{figure}

\vskip 2mm
5. One of the probabilities tends to a fixed point, but the other starts 
fluctuating chaotically and continues fluctuating for ever, as in Fig. 5. 

\begin{figure}[ht]
\centerline{
\includegraphics[width=8cm]{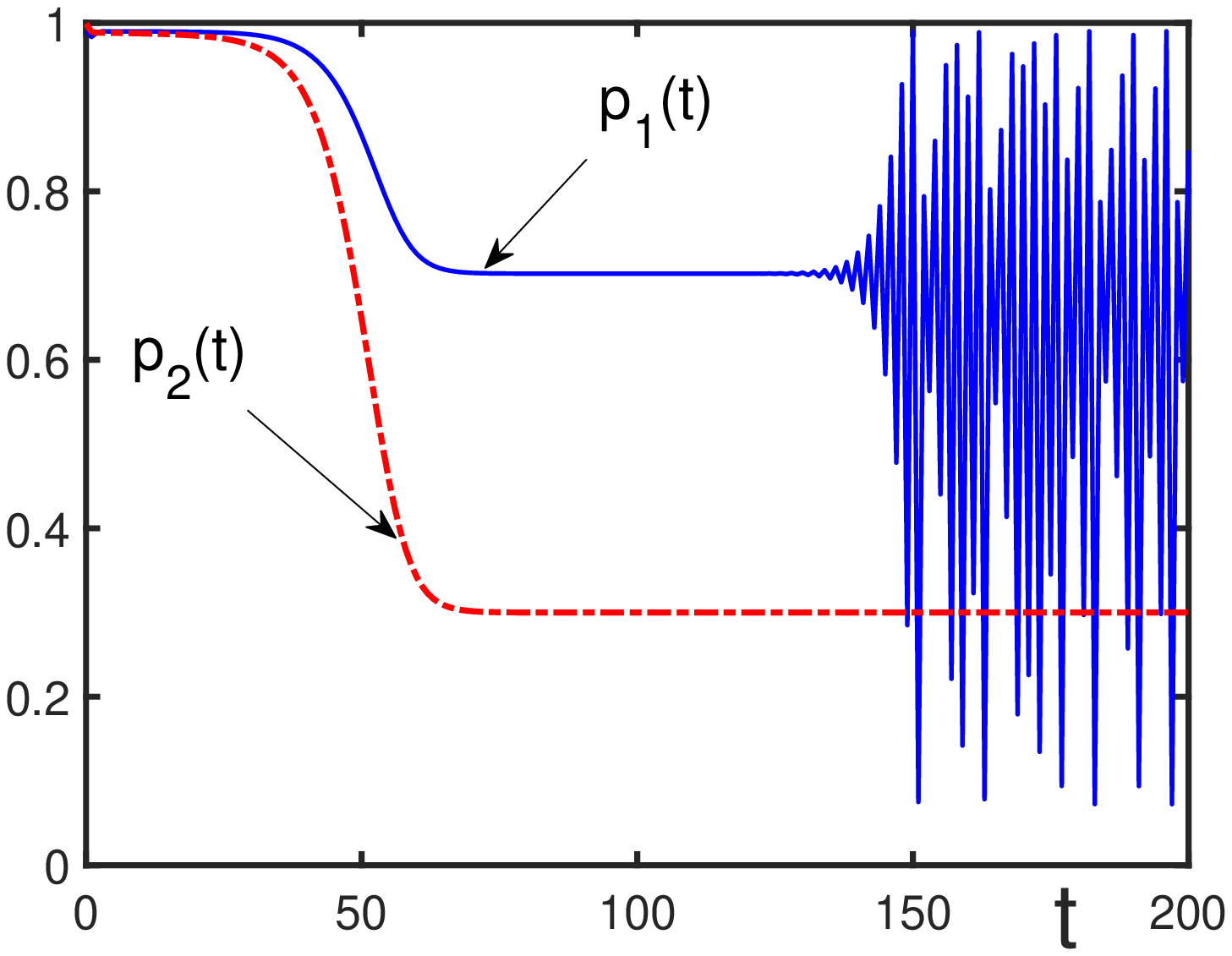}  }
\caption{\small Time dependence of the probabilities $p_1(t)$ (solid line) and 
$p_2(t)$ (dash-dotted line), for the initial conditions $f_1=0.3$, $q_1=0.699$ 
and $f_2=0$, $q_2=0.99$, with the herding parameters $\ep_1=\ep_2=1$. The 
probability $p_1(t)$, after a period of smooth behaviour, suddenly begins 
chaotically oscillate, while $p_2(t)$ monotonically tends to the fixed point 
$p_2^*=0.3$. 
} 
\label{fig:Fig.5}
\end{figure}

\vskip 2mm
6. Both probabilities at some moment of time suddenly begin fluctuating 
chaotically, and keep fluctuating for ever, as in Fig. 6.

\begin{figure}[ht]
\centerline{
\includegraphics[width=8cm]{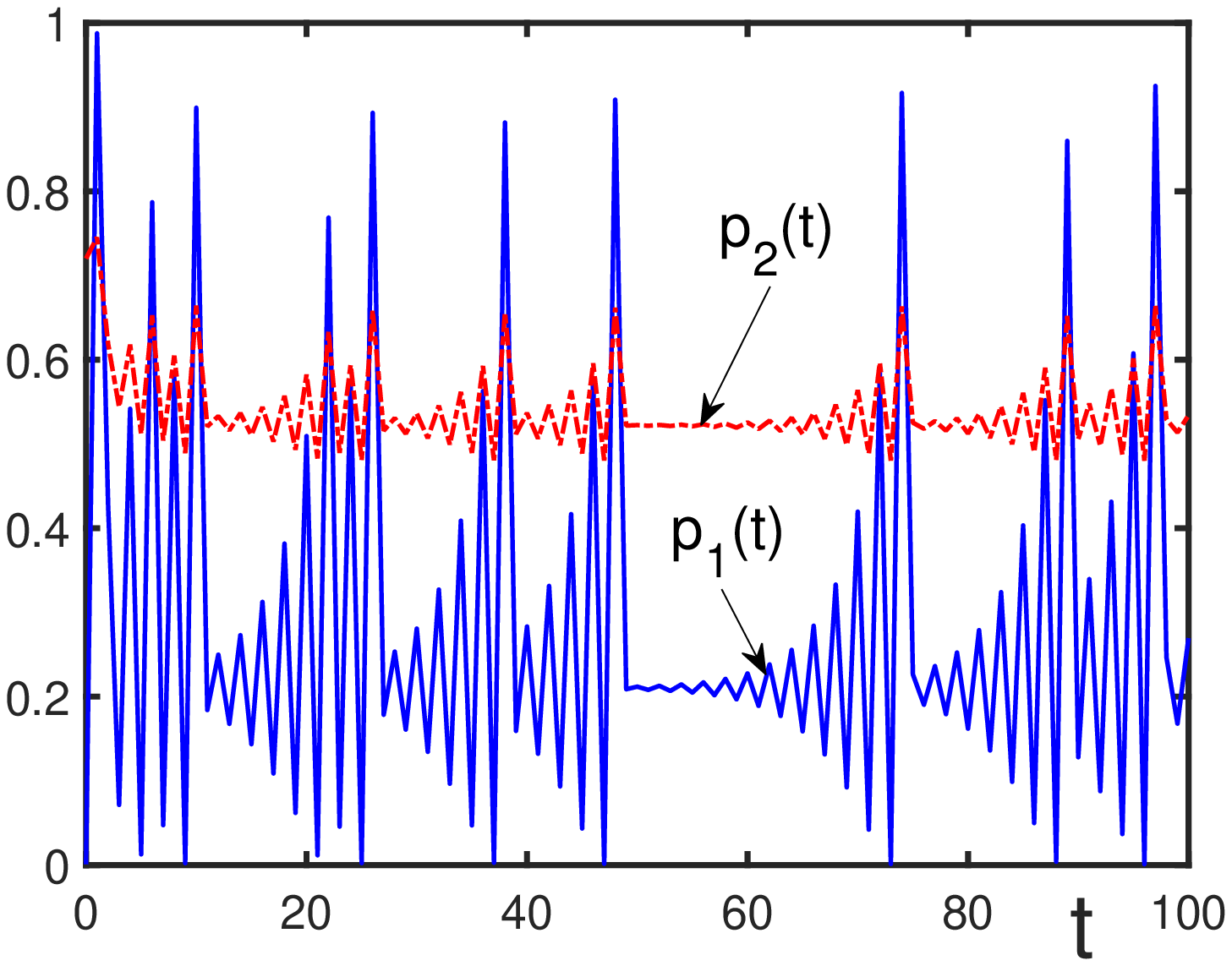}  }
\caption{\small Time dependence of the probabilities $p_1(t)$ (solid line) and 
$p_2(t)$ (dash-dotted line), for the initial conditions $f_1=0.6$, $q_1=0.3$ 
and $f_2=1$, $q_2=-0.999$, under the herding parameters $\ep_1=1$ and $\ep_2=0.8$. 
Both probabilities exhibit perpetual chaotic oscillations.
}
\label{fig:Fig.6}
\end{figure}

\vskip 2mm
The sudden appearance of self-excited periodic and chaotic oscillations is an 
interesting effect arising in a closed system. The herding behaviour smoothes 
the trajectories and suppresses both kinds of oscillations, periodic as well as 
chaotic. The appearance of large oscillations and chaotic fluctuations in decision
making can be understood as representing unstable decision making of mentally ill
individuals.

\subsection{Attenuation of emotion influence}

The existence of behavioural paradoxes, considered above, is caused by the 
presence of emotions of decision makers. The paradoxes arise when subjects make 
decisions without exchanging information with each other. However, when they are 
allowed to discuss the problem and to exchange their thoughts, the deviation of 
the behavioural probabilities from the utility factors diminishes, hence the 
attraction factors decrease. The paradox attenuation has been confirmed in numerous 
experimental studies \cite{Kuhberger_2001,Charness_2002,Blinder_2005,Cooper_2005,
Sutter_2005,Tsiporkova_2006,Charness_2007,Charness_Rigotti_2007,Chen_2009,Liu_2009,
Charness_2010}.

Thus a series of experiments \cite{Charness_2010} were accomplished, designed to 
test whether and to what extent individuals succumb to the conjunction fallacy 
after discussing the problem with each other. The experimental design of Tversky 
and Kahneman \cite{Tversky_1983} was used. It was found that, when subjects are 
allowed to consult between themselves, the proportion of individuals who violate 
the conjunction principle falls dramatically, especially when the size of the 
group rises. When individuals are forced to think, they recover in their minds 
additional information that has been forgotten or shadowed by emotions. The amount 
of received information increases with the size of the group. As a result, there 
is a substantially larger drop in the error rate when the group size is increased 
from two to three than when it is increased from one to two 
\cite{Charness_2007,Charness_2010}. These experiments confirm the earlier results 
\cite{Sutter_2005} that there is only a marginal difference between the choices 
of single individuals and two-person groups, but a significant difference between 
the choices of two-person and four-person groups in a game. In the experiment by 
Charness et al. \cite{Charness_2010} with groups of three members, the fraction 
of individuals giving the answers deviating from the utility factor dropped to 
$0.17$. The latter quantity, actually, is the measure of the attraction factor. 
Thus, after consultations providing additional information, that is after the 
repeated exchange of information, the attraction factor falls from about $0.25$ to 
$0.17$. It seems to be natural that, if the discussions and information exchange 
would continue, the attraction factor would drop further, but it is clear that  
it should not disappear completely, since the amount of received information is 
actually never infinite. 

The effect of the paradox attenuation, or in other words, the attenuation of 
the attraction factors after repeated exchange of information, under the conditions 
discussed above, is illustrated in Fig. 7 and Fig. 8 demonstrating the disjunction 
effect attenuation. Here the agents of both groups enjoy long-term memory, share 
the same utility factor, taken as in the experiments of Tversky and Shafir 
\cite{Tversky_Shafir_1992} $f_1 =f_2 = 0.64$, but differ by the initial conditions  
for the attraction factors. The probabilities $p_1(t)$ and $p_2(t)$ tend to 
the consensual limit equal to the utility factor, while the attraction factor, 
characterizing the paradox, tends to zero, in agreement with the paradox attenuation 
effect \cite{Kuhberger_2001,Charness_2002,Blinder_2005,Cooper_2005,Sutter_2005,
Tsiporkova_2006,Charness_2007,Charness_Rigotti_2007,Chen_2009,Liu_2009,Charness_2010}. 
The tendency of the attraction factors to zero is preserved under different herding 
parameters. 

\begin{figure}[ht]
\centerline{
\hbox{ \includegraphics[width=7.5cm]{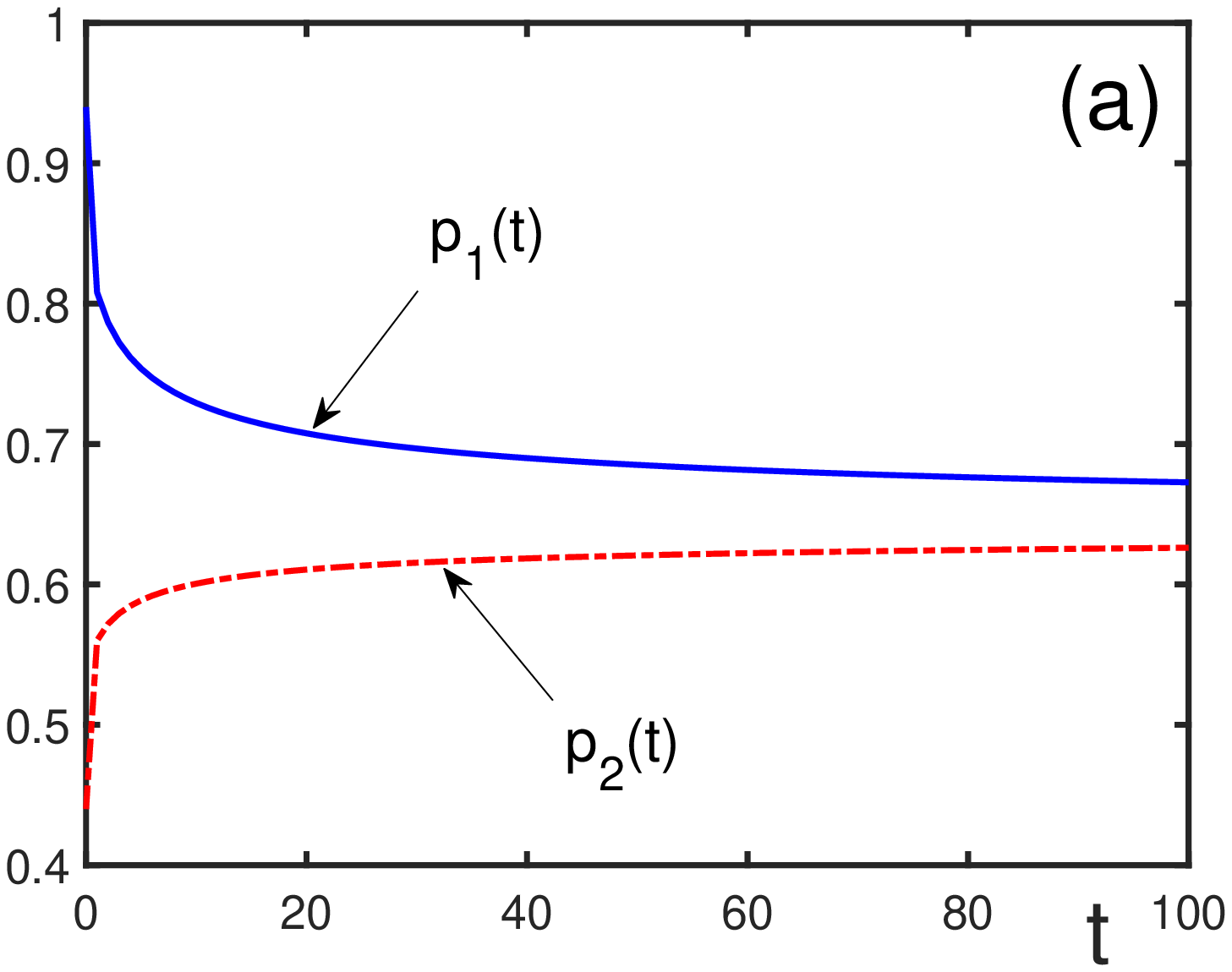} \hspace{0.5cm}
\includegraphics[width=7.5cm]{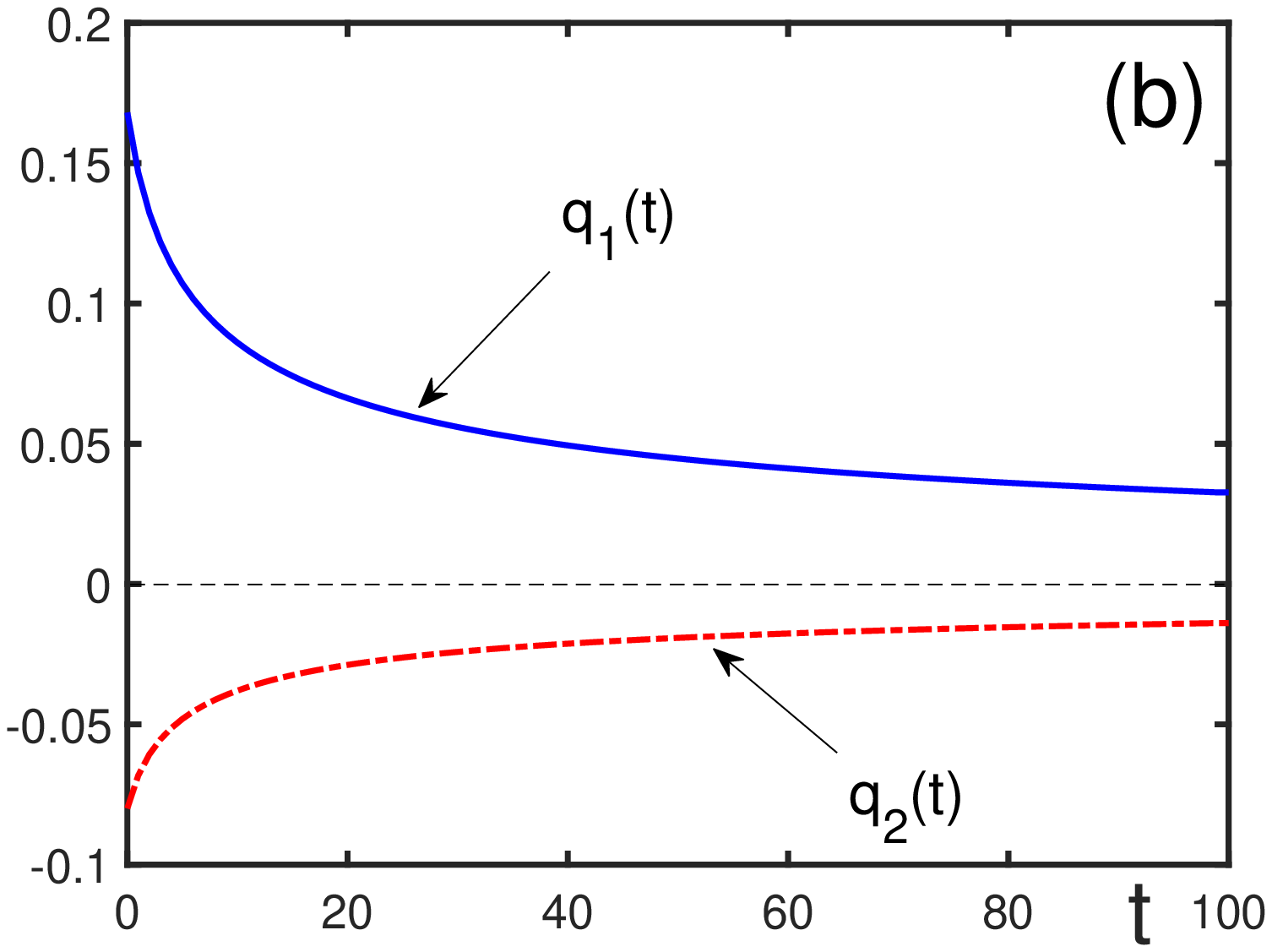}  } }
\caption{\small 
Dynamic disjunction effect. 
(a) Time dependence of the probabilities $p_1(t)$ (solid line) and $p_2(t)$ 
(dash-dotted line) for the initial conditions $f_1=f_2=0.64$, and $q_1=0.3$, 
$q_2=-0.2$, in the absence of herding $\ep_1=\ep_2=0$. Both probabilities 
tend to the consensual limit $p_1^*=p_2^*=0.64$; 
(b) Time dependence of the attraction factors $q_1(t)$ (solid line) and 
$q_2(t)$ (dash-dotted line) for the same conditions as in Fig.7a. Both attraction 
factors tend to $q_1^*=q_2^*=0$, demonstrating disjunction-effect attenuation. 
}
\label{fig:Fig.7}
\end{figure}

\begin{figure}[ht]
\centerline{
\hbox{ \includegraphics[width=7.5cm]{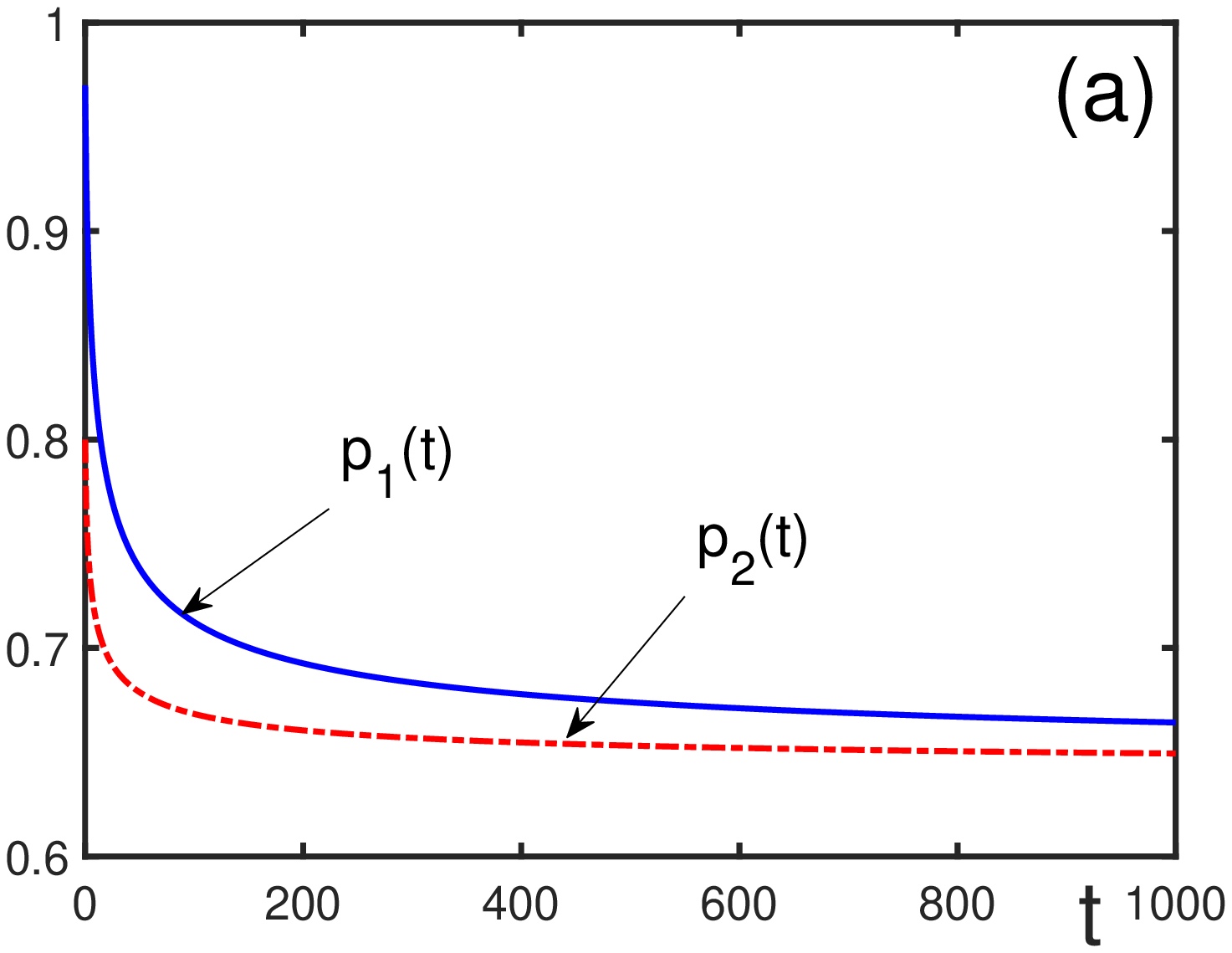} \hspace{0.5cm}
\includegraphics[width=7.5cm]{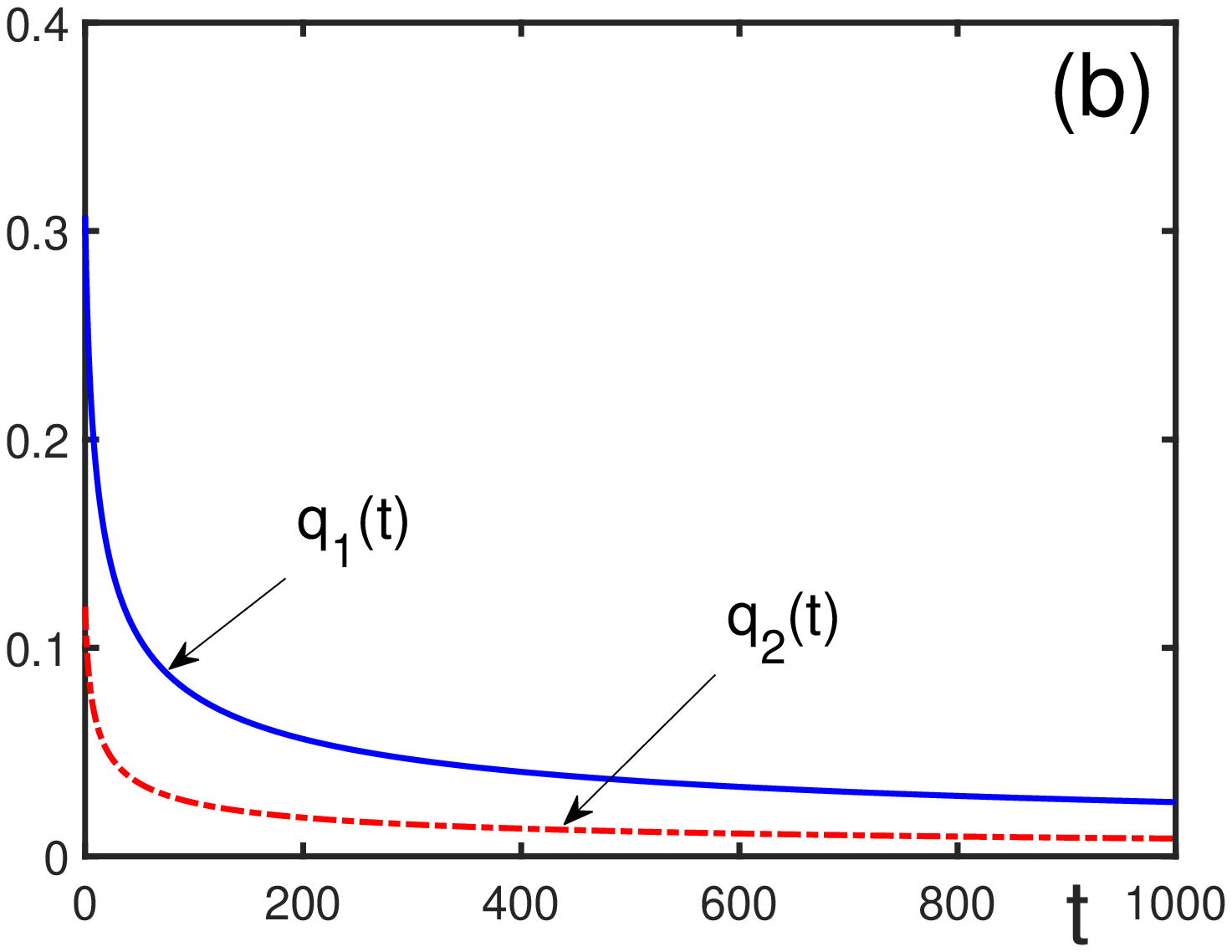}  } }
\caption{\small 
Dynamic disjunction effect. 
(a) Time dependence of the probabilities $p_1(t)$ (solid line) and $p_2(t)$ 
(dash-dotted line) for the initial conditions $f_1=f_2=0.64$, and $q_1=0.35$, 
$q_2=0.15$, for the herding parameters $\ep_1=0.1$ and $\ep_2=0.05$. Both 
probabilities tend to the consensual limit $p_1^*=p_2^*=0.64$; 
(b) Time dependence of the attraction factors $q_1(t)$ (solid line) and $q_2(t)$ 
(dash-dotted line) for the same conditions as in Fig.8a. Both attraction factors 
tend to $q_1^*=q_2^*=0$, demonstrating disjunction-effect attenuation. 
}
\label{fig:Fig.8}
\end{figure}

\subsection{Continuous decision making}

The evolution equations describing the dynamics of preferences deal with multistep 
decision making, which implies that decisions are taken in discrete moments of time.
The temporal interval between subsequent decisions is denoted by $\tau$. If the time
of observing the process of taking decisions by decision makers $t$ is much longer 
than $\tau$, then it seems admissible to expand the probability $p_j(A_n,t+\tau)$ 
over the small ratio $\tau/t \ll 1$. To first order, this gives
\be 
\label{5.36}
 p_j(A_n,t+\tau) \simeq p_j(A_n,t) + 
\frac{\prt p_j(A_n,t)}{\prt t}\; \tau \; .
\ee    
Measuring time in units of $\tau$, we have
\be
\label{5.37}
  p_j(A_n,t+1) - p_j(A_n,t) = \frac{\prt p_j(A_n,t)}{\prt t} \;  .
\ee
Then equation (\ref{5.5}) becomes a differential equation
\be
\label{5.38}
 \frac{\prt p_j(A_n,t)}{\prt t} = f_j(A_n,t) + q_j(A_n,t) +
h_j(A_n,t) - p_j(A_n,t) \; .
\ee

Equation (\ref{5.38}) describes the imaginary process of continuous decision 
making, assuming that agents can take decisions continuously in time. Of course, 
such a continuous process sounds not realistic. However, equation (\ref{5.38}), 
as it seems, is derived under a reasonable assumption that the observation time 
is much longer than the time of taking a single decision. It is therefore 
interesting whether the description of multistep decision making can be done 
by employing the continuous approximation, that is whether the solutions of the 
discrete in time equations, characterizing multistep decision making, could be 
approximated by solutions of continuous in time equation.          

The case of two groups deciding on two alternatives, characterized by 
Eq. (\ref{5.34}), in the continuous approximation reduces to the equation
\be
\label{5.39}
 \frac{d p_j(t)}{dt} = f_j + q_j(t) + h_j(t) - p_j(t) \;  .
\ee
Substituting here the herding factors (\ref{5.30}) yields the evolution equations
$$
 \frac{d p_1(t)}{dt} = ( 1 - \ep_1) [\; f_1 + q_1(t) \; ] + 
\ep_1  [\; f_2 + q_2(t) \; ] - p_1(t) \; ,
$$
\be
\label{5.40}
 \frac{d p_2(t)}{dt} = ( 1 - \ep_2) [\; f_2 + q_2(t) \; ] + 
\ep_2  [\; f_1 + q_1(t) \; ] - p_2(t) \;  ,
\ee
with the attraction factors (\ref{5.29}). Initial conditions are given by 
(\ref{5.35}). If we keep in mind the case of mixed memories, then one group 
possesses long-term memory (\ref{5.31}) that, using the Euler-Maclaurin 
formula,  
translates into
\be
\label{5.41}
M_1(t) = \int_0^t \mu_{12}(t') \; dt' + \frac{\mu_{12}(0) + 
\mu_{12}(t)}{2}  \; ,
\ee
while the other group has short-term memory (\ref{5.32}). For $t=0$, we have
\be
\label{5.42}
M_1(0) = \mu_{12}(0) \; , \; M_2(0) = \mu_{21}(0) \; .
\ee

The problem of whether the solutions of time-discrete equations are close to 
the solutions of their time-continuous approximation is interesting by its own. 
It is also important for the understanding which of the variants more correctly 
characterizes real multistep decision making.

\subsection{Discrete versus continuous processes}

Comparing the solutions to discrete and continuous in time equations 
(\ref{5.34}) and (\ref{5.40}), respectively, it is instructive to compare the 
solutions for the same parameters and the same initial conditions. As is well 
known, for two-dimensional dynamical systems, continuous differential equations 
usually produce more smooth solutions than those of discrete difference equations 
\cite{Kulenovic_2019}. This is confirmed by the results of numerical solution of 
the differential equations (\ref{5.40}) and their comparison with the solutions 
of discrete equations (\ref{5.34}). 

Figure 9 demonstrates that the solution to the continuous equation is smooth, 
and does not display oscillations as the solution to the discrete equation. 
When the solution to the discrete problem is smooth, then the solution to the 
continuous problem is also smooth and the related probabilities practically 
coincide, as it happens for the parameters of Fig. 2, when the solutions of 
discrete and continuous equations are not distinguishable, because of which we 
do not repeat them here.

\begin{figure}[ht]
\centerline{
\hbox{\includegraphics[width=7.5cm]{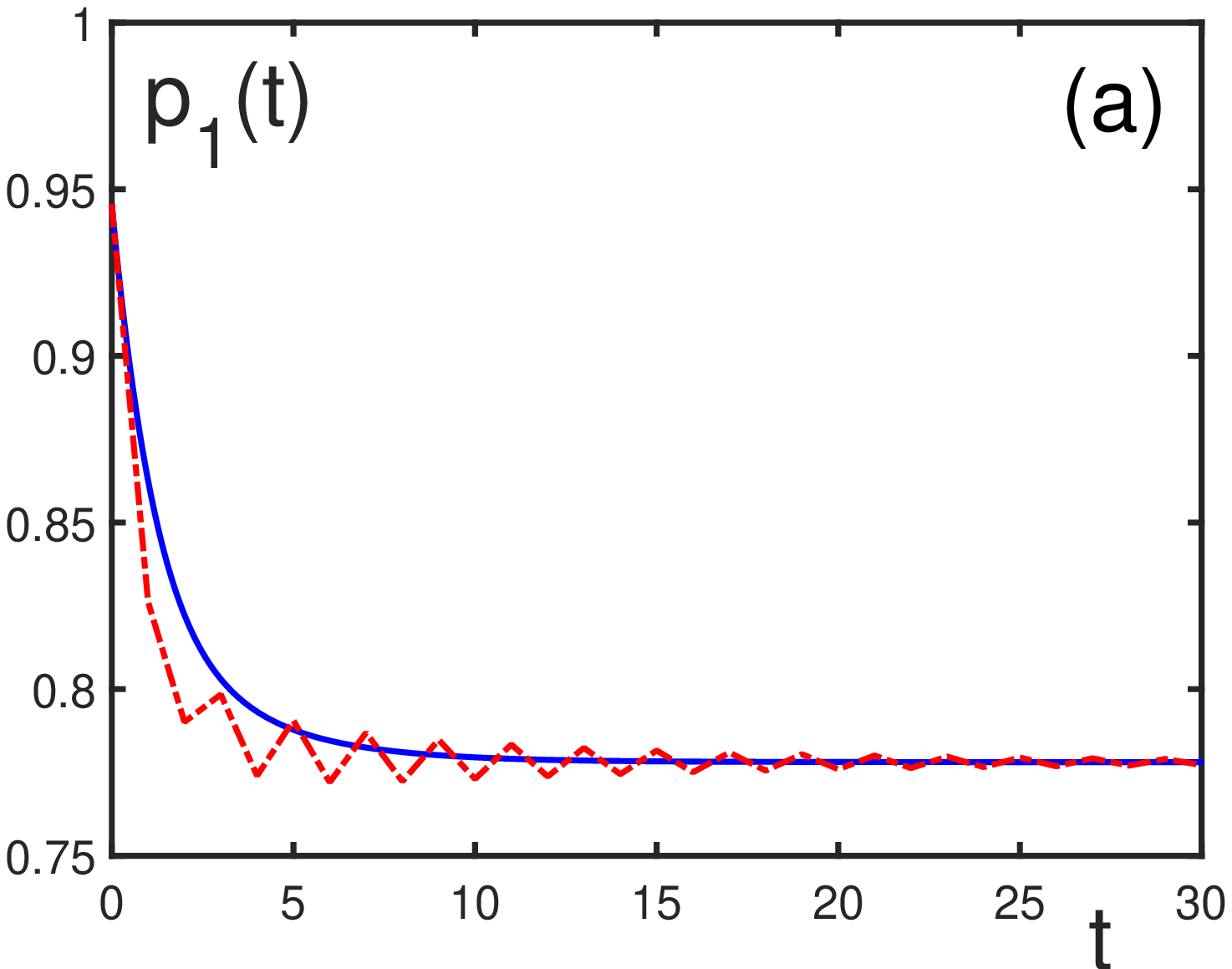} \hspace{0.5cm}
\includegraphics[width=7.5cm]{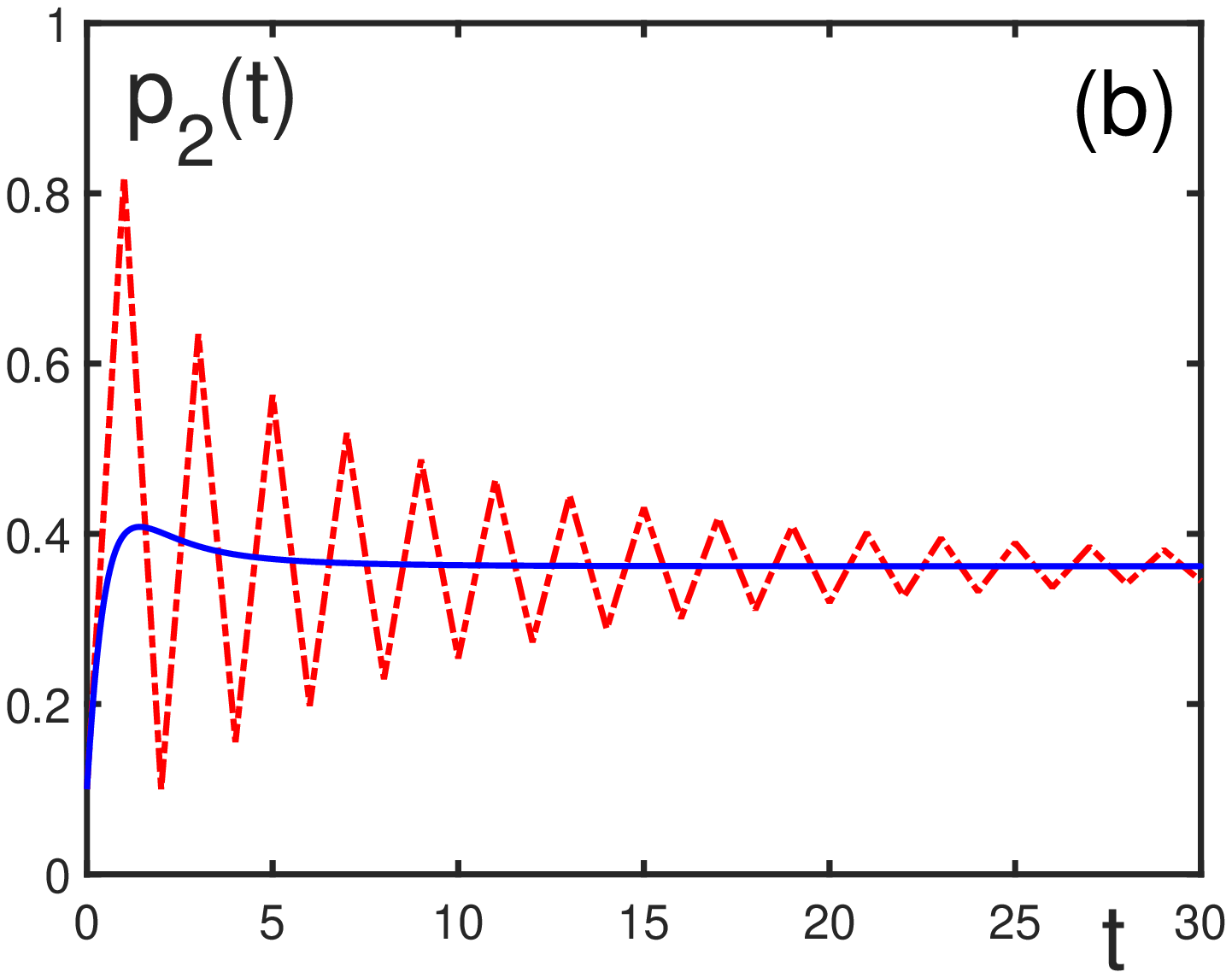}  } }
\caption{\small Comparison of the solutions for the probabilities $p_1(t)$ 
and $p_2(t)$ for the discrete and continuous decision making. The initial 
conditions $f_1=0.8$, $q_1=0.19$ and $f_2=0.9$, $q_2=-0.8$, under weak herding, 
with the parameters $\ep_1=0.05$ and $\ep_2=0$. The probabilities tend to the 
fixed points $p^*_1=0.778$ and $p_2^*=0.362$. (a) Solution $p_1(t)$ to continuous 
equations (solid line), as compared to $p_1(t)$ for discrete equations 
(dash-dotted line); 
(b) Solution $p_2(t)$ to continuous equations (solid line) as compared to 
$p_2(t)$ for discrete equations (dash-dotted line).
}
\label{fig:Fig.9}
\end{figure}

If the evolution equations possess fixed points, they are defined by the 
equations
$$
p_1^*=( 1 - \ep_1 )( f_1 + q_1^* ) + \ep_1( f_2 + q_2^* ) \; , 
\qquad
p_2^*=( 1 - \ep_2 )( f_2 + q_2^* ) + \ep_2( f_1 + q_1^* ) \; ,
$$  
where $q_j^* \equiv q_j(\infty)$.
    
Figures 10 and 11 show that the solutions to the continuous equations do not 
display oscillatory behaviour that exists for discrete equations. And Fig. 12 
evidences that the solutions to the continuous equations, as it should be 
expected, do not possess chaotic behaviour, demonstrated by the discrete 
equations.  

\begin{figure}[ht]
\centerline{
\hbox{\includegraphics[width=7.5cm]{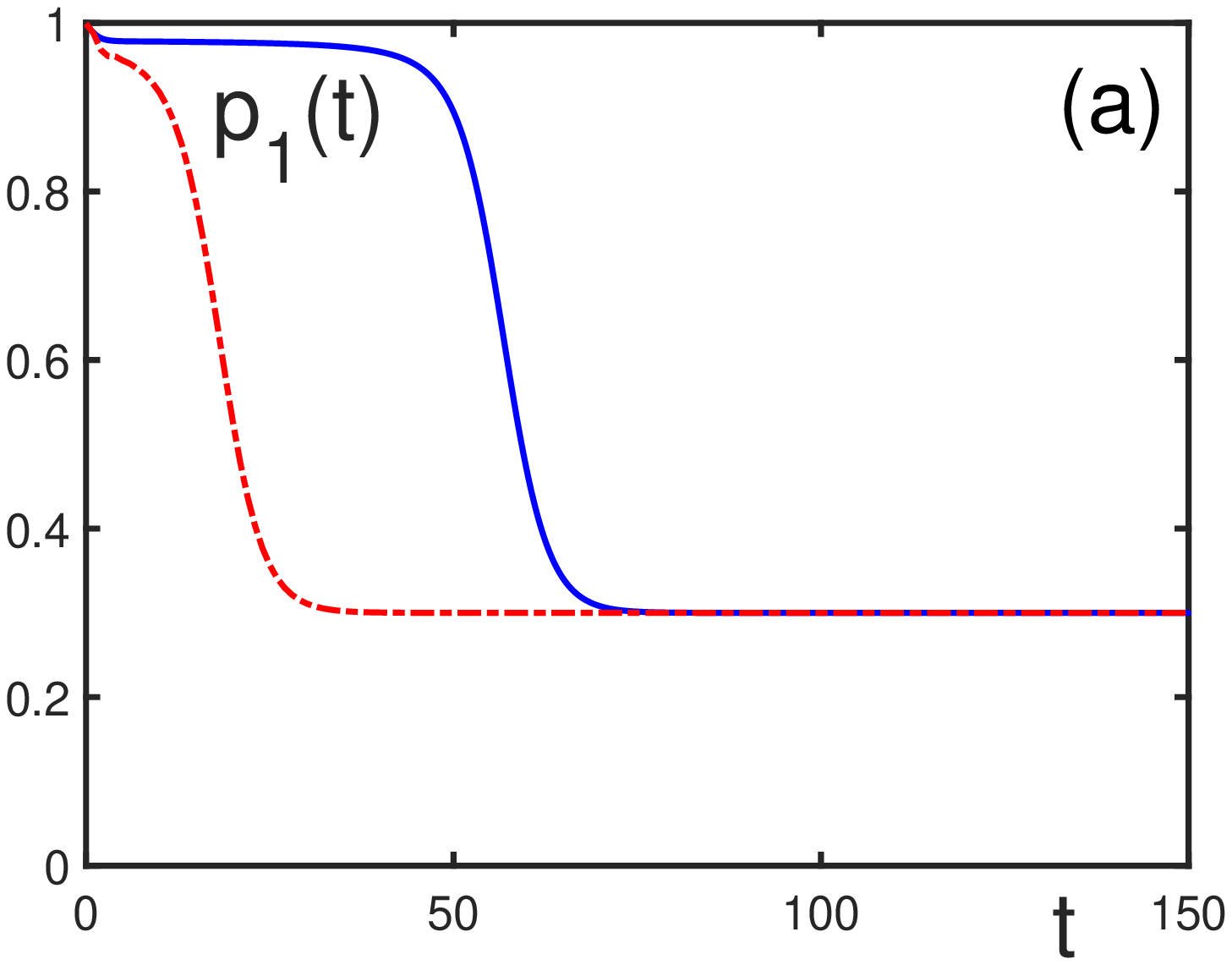} \hspace{0.5cm}
\includegraphics[width=7.5cm]{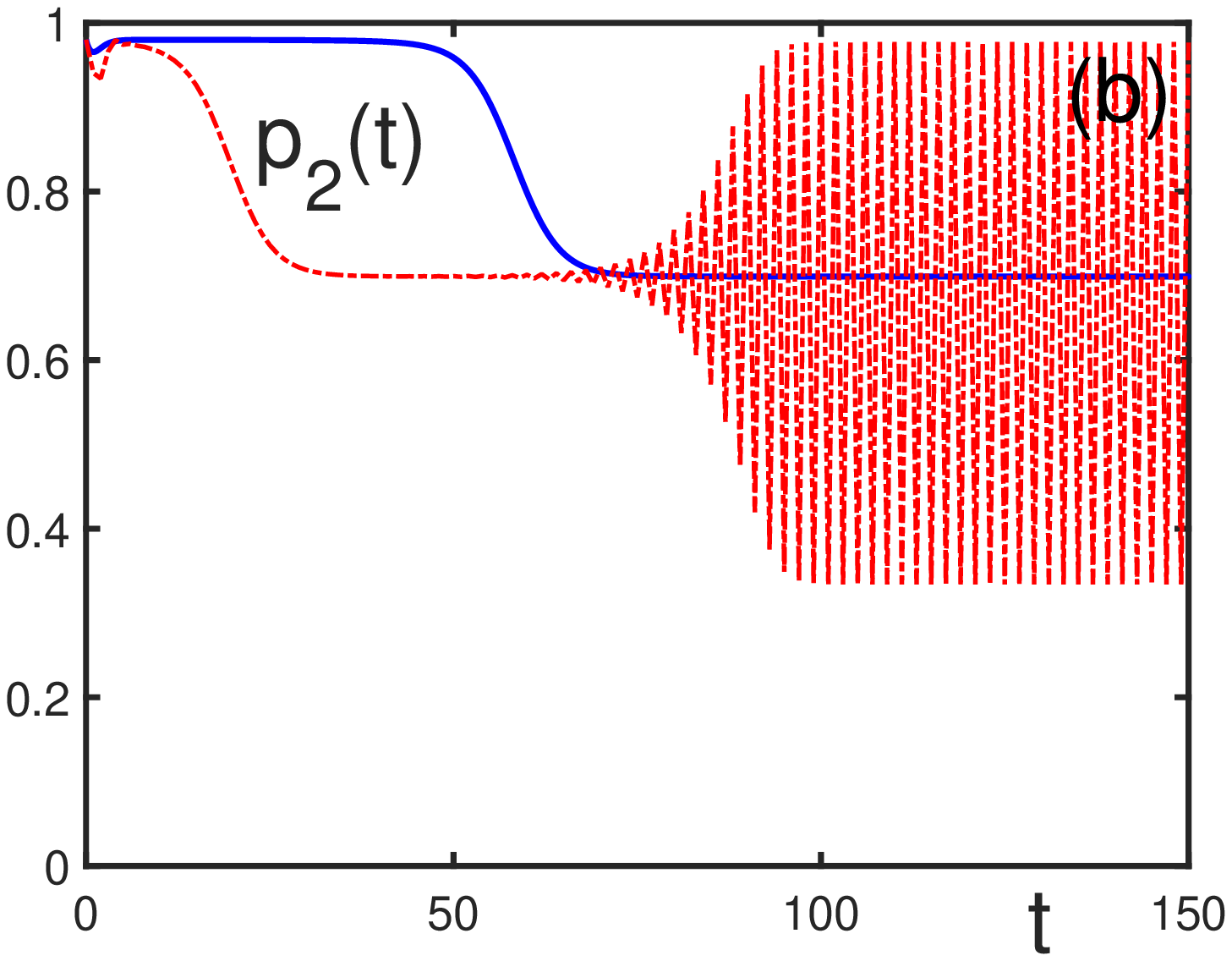}  } }
\caption{\small 
Comparison of the solutions for the probabilities $p_1(t)$ and 
$p_2(t)$ for the discrete and continuous decision making. The initial conditions 
are $f_1=0.3$, $q_1=0.699$ and $f_2=0$, $q_2=0.98$, in the absence of 
herding, with $\ep_1=\ep_2=0$. 
(a) Solution $p_1(t)$ to continuous equations (solid line), as compared to 
$p_1(t)$ for discrete equations (dash-dotted line). Both probabilities $p_1(t)$ 
tend to the same fixed point $p_1^*=0.3$; 
(b) Solution $p_2(t)$ to continuous equations (solid line), as compared to 
$p_2(t)$ for discrete equations (dash-dotted line). The continuous solution 
$p_2(t)$ tends to the fixed point $p_2^*=0.699$, whereas the discrete $p_2(t)$, 
after a period of smooth behaviour, starts periodically oscillating.
}
\label{fig:Fig.10}
\end{figure}

\begin{figure}[ht]
\centerline{
\hbox{\includegraphics[width=7.5cm]{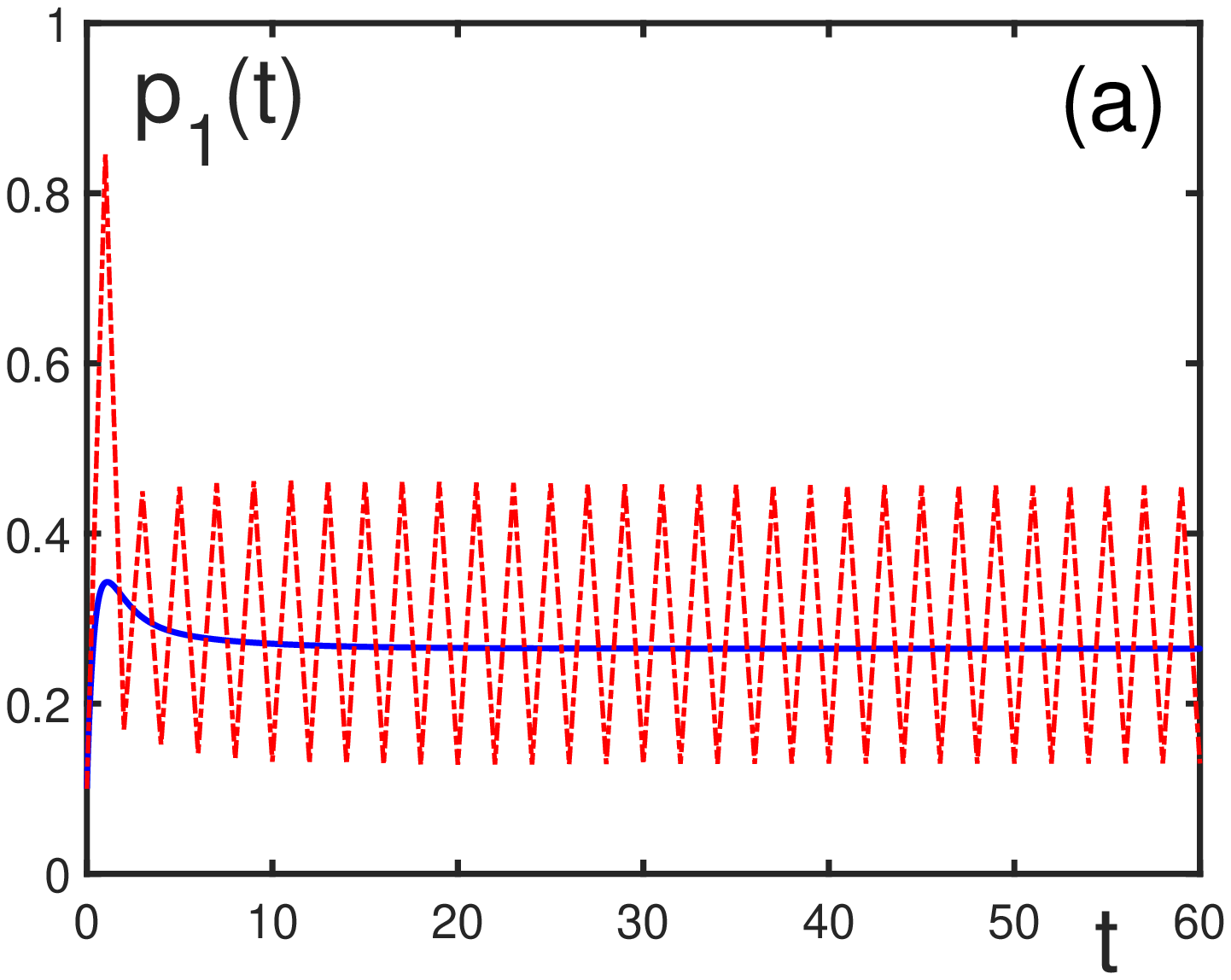} \hspace{0.5cm}
\includegraphics[width=7.5cm]{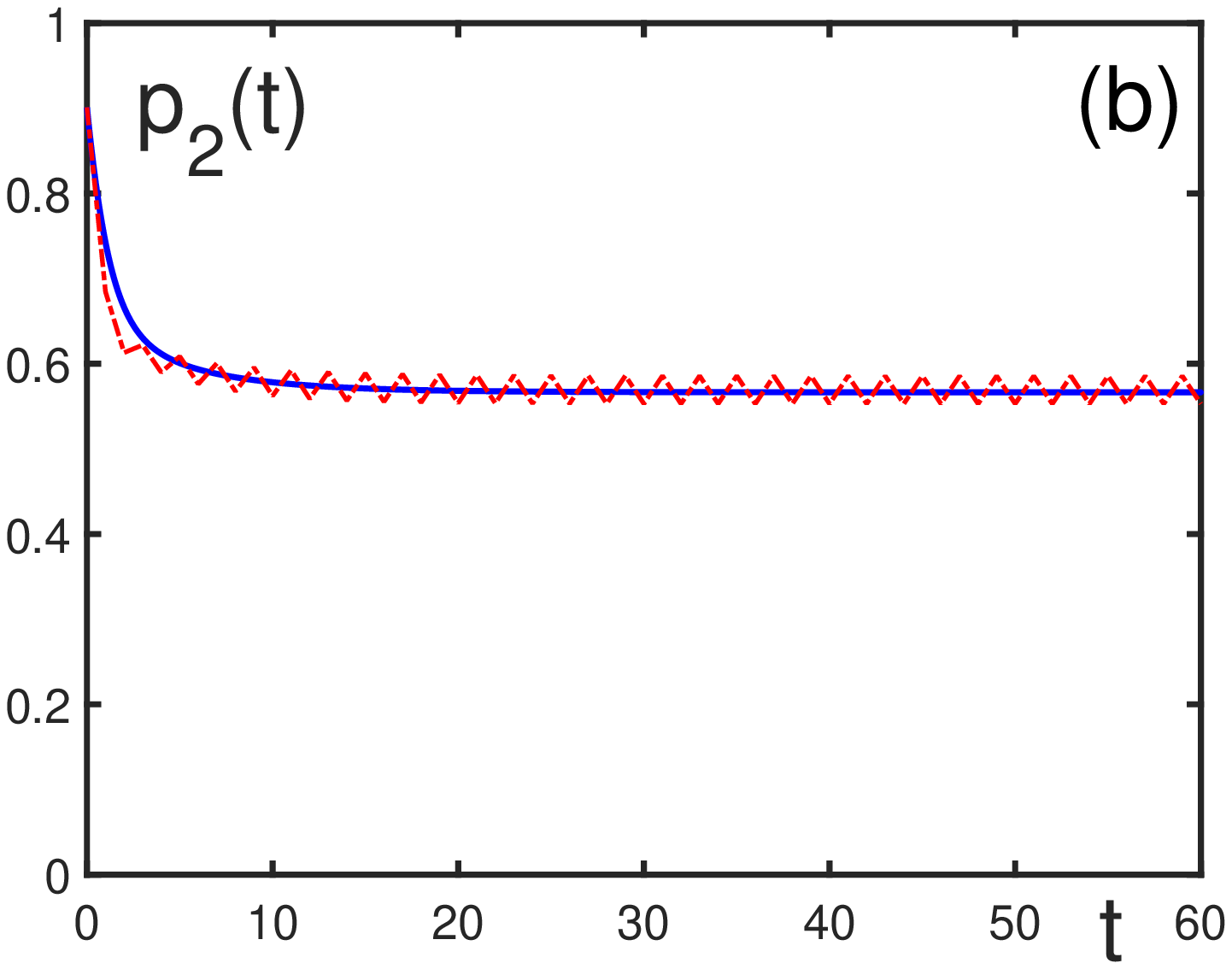}  } }
\caption{\small Comparison of the solutions for the probabilities $p_1(t)$ and 
$p_2(t)$ for the discrete and continuous decision making. The initial conditions 
are $f_1=0.6$, $q_1=0.39$, and $f_2=1$, $q_2=-0.9$, under strong herding, with 
the parameters $\ep_1=1$ and $\ep_2=0.9$. 
(a) Solution $p_1(t)$ to the continuous equations (solid line), as compared 
to $p_1(t)$ for the discrete equations (dash-dotted line). The continuous 
probability tends to the fixed point $p_1^*=0.265$, while the discrete 
probability perpetually oscillates; 
(b) Solution $p_2(t)$ to continuous equations (solid line), as compared to 
$p_2(t)$ for discrete equations (dash-dotted line). The continuous solution 
$p_2(t)$ tends to the fixed point $p_2^*=0.566$, while the discrete solution 
$p_2(t)$ permanently oscillates.
}
\label{fig:Fig.11}
\end{figure}

\begin{figure}[ht]
\centerline{
\hbox{\includegraphics[width=7.5cm]{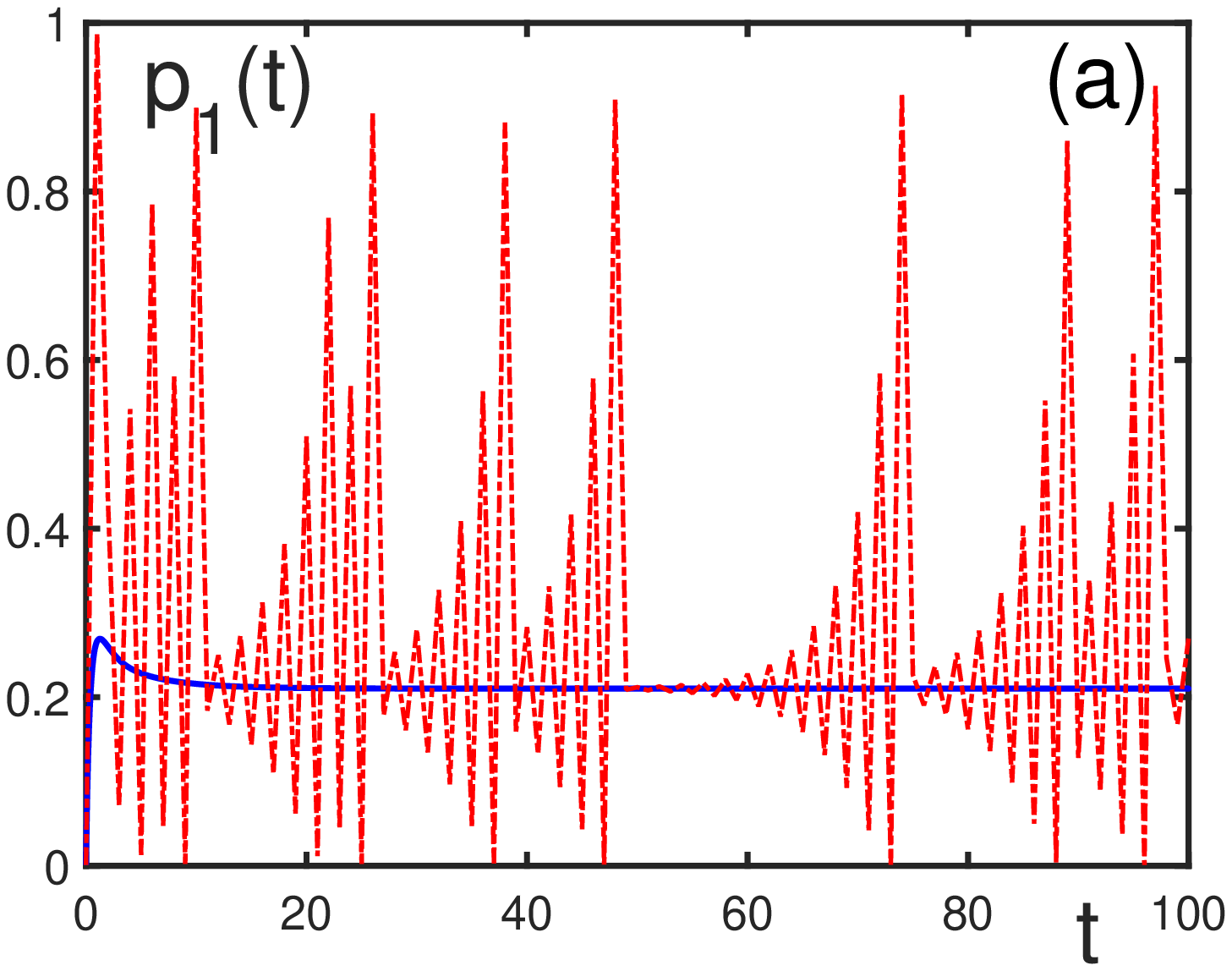} \hspace{0.5cm}
\includegraphics[width=7.5cm]{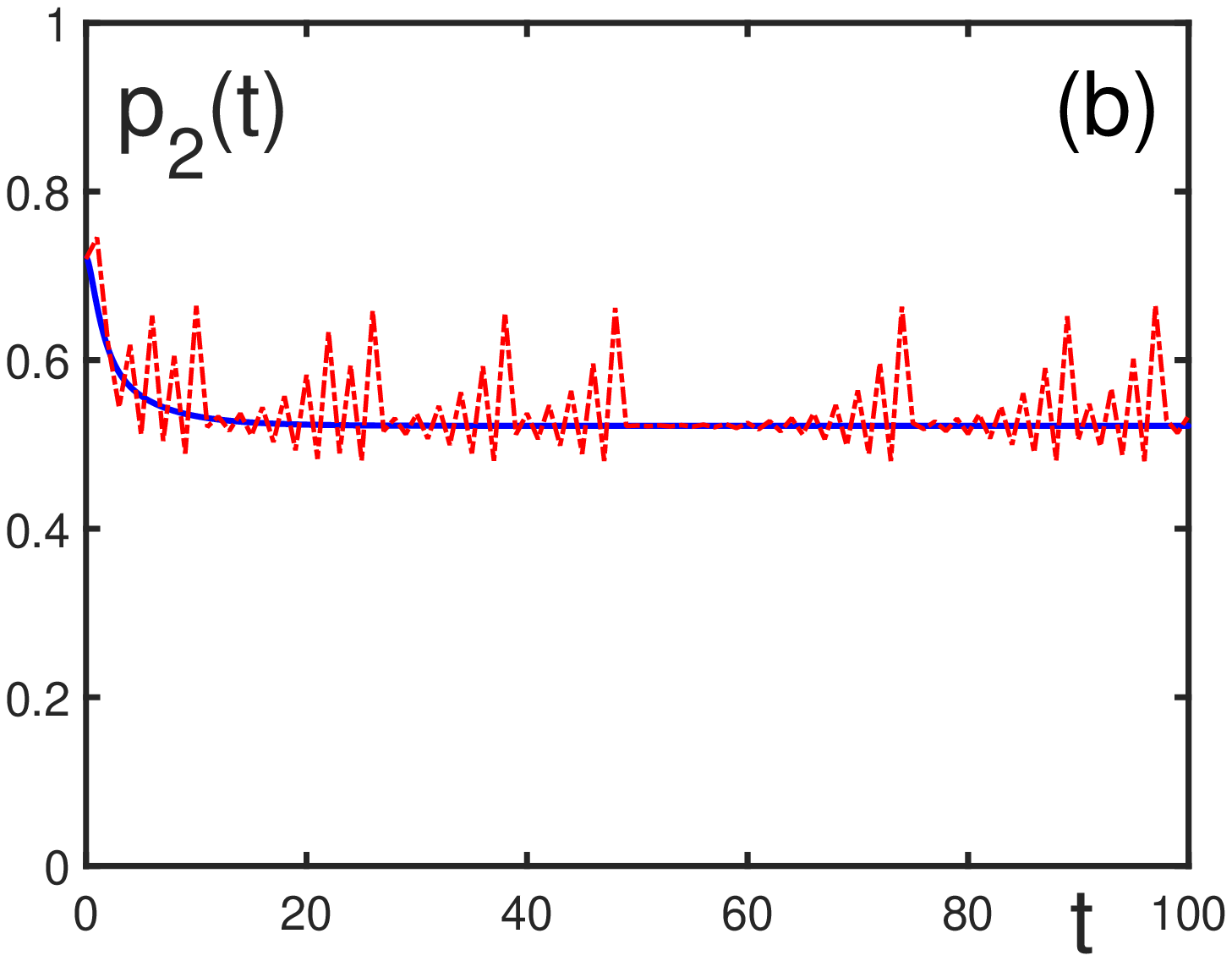}  } }
\caption{\small Comparison of the solutions for the probabilities $p_1(t)$ and 
$p_2(t)$ for the discrete and continuous decision making. The initial conditions 
are $f_1=0.6$, $q_1=0.3$ and $f_2=1$, $q_2=-0.999$, under the herding parameters 
$\ep_1=1$ and $\ep_2=0.8$. 
(a) Solution $p_1(t)$ to the continuous equations (solid line), as compared to 
$p_1(t)$ for the discrete equations (dash-dotted line). The continuous 
probability tends to the fixed point $p_1^*=0.210$, while the discrete probability 
chaotically oscillates; 
(b) Solution $p_2(t)$ to the continuous equations (solid line), as compared to 
$p_2(t)$ for the discrete equations (dash-dotted line). The continuous solution 
$p_2(t)$ tends to the fixed point $p_2^*=0.522$, while the discrete solution 
$p_2(t)$ chaotically oscillates.
}
\label{fig:Fig.12}
\end{figure}

A special case is illustrated in Fig. 13. The solutions $p_1(t)$ and $p_2(t)$ 
to the discrete equations sometimes can intersect, as is shown in Fig. 13b. 
However, the solutions to the continuous equations do not intersect, instead 
they glue to each other and tend to a common limit, while in the case of the 
discrete problem, one of the probabilities chaotically oscillates, as is shown 
in Fig. 14.

\begin{figure}[ht]
\centerline{
\hbox{\includegraphics[width=7.5cm]{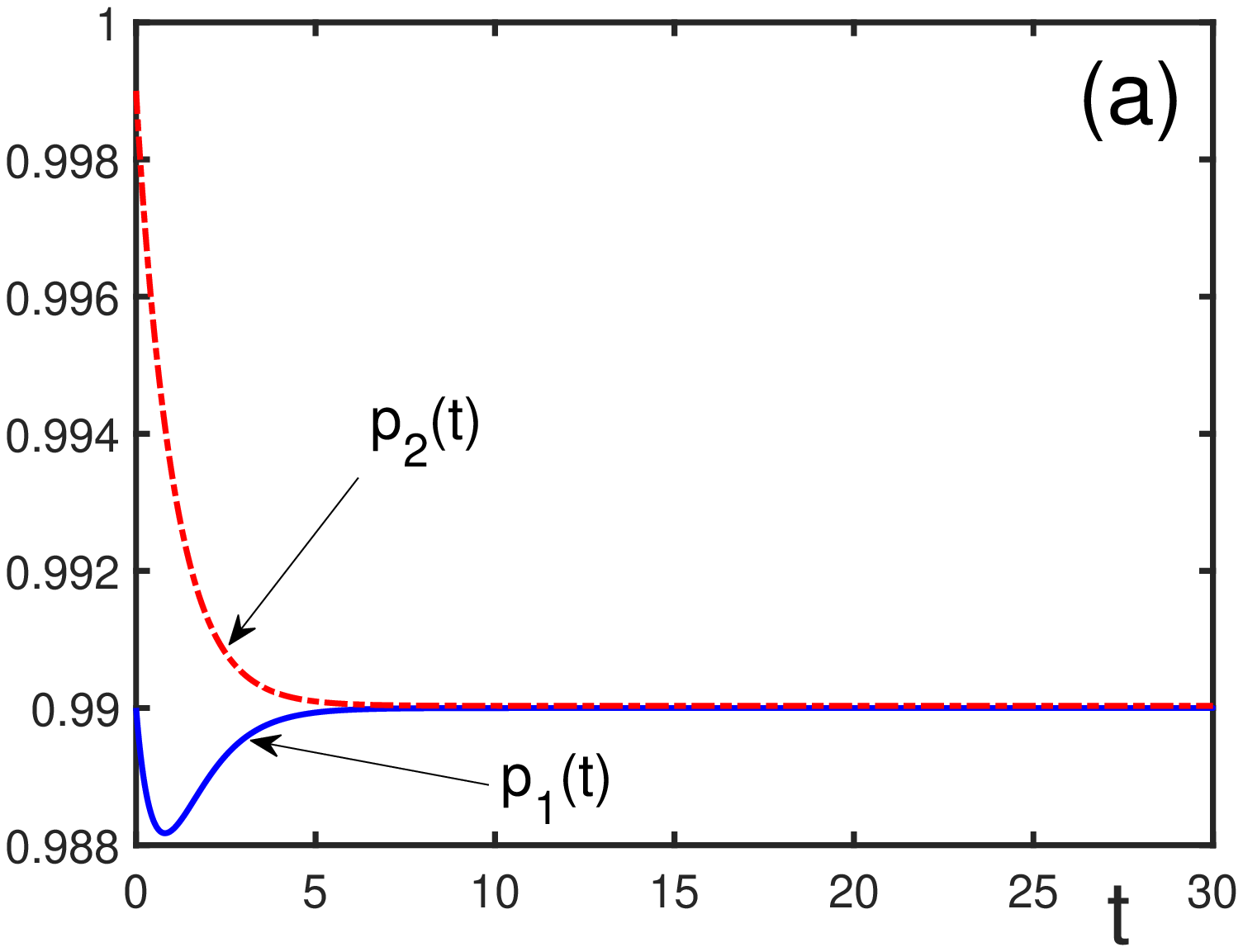} \hspace{0.5cm}
\includegraphics[width=7.5cm]{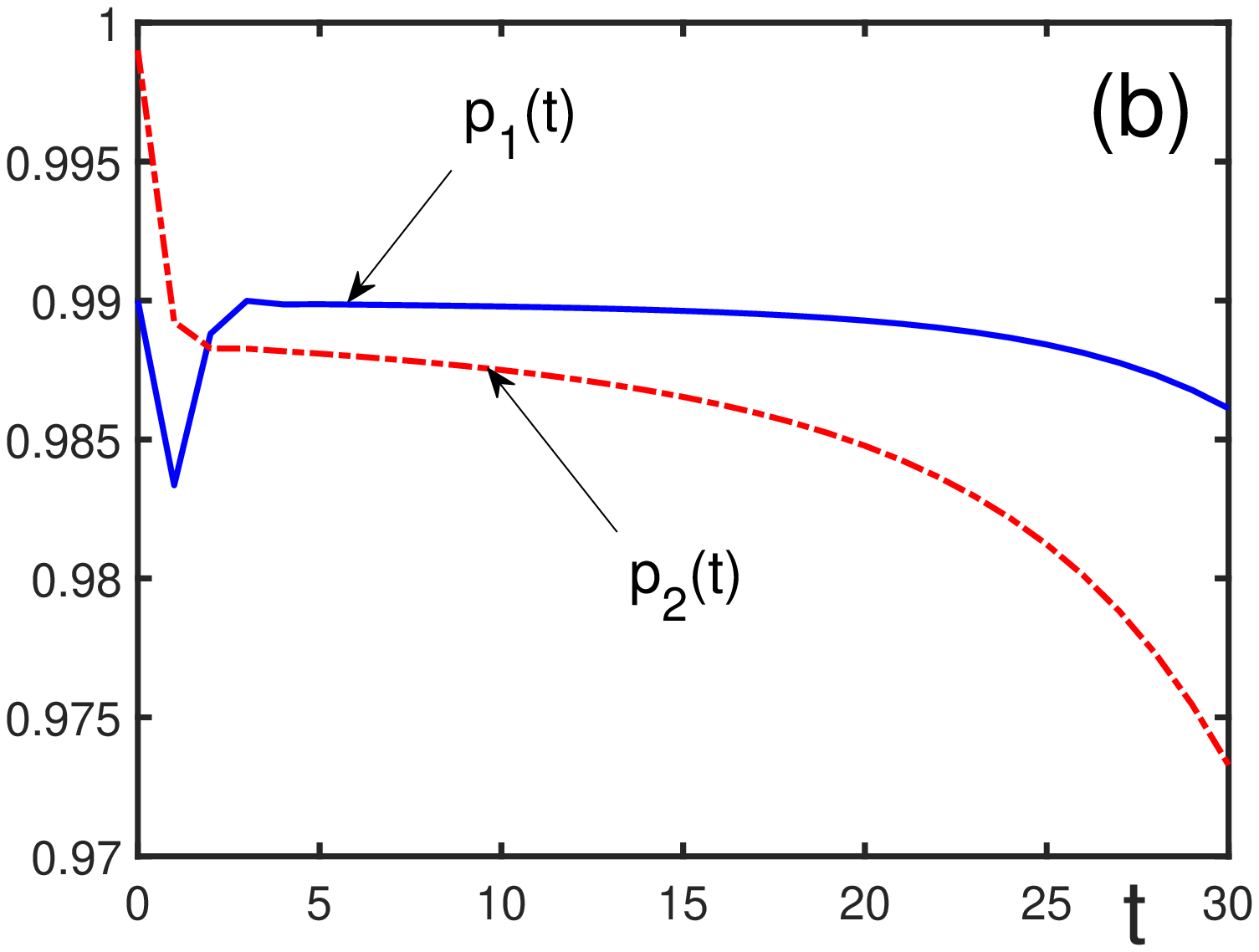}  } }
\caption{\small Comparison of the solutions for the probabilities $p_1(t)$ 
(solid line) and $p_2(t)$ (dash-dotted line) for the discrete and continuous 
decision making. The initial conditions are $f_1=0.3$, $q_1=0.699$ and $f_2=0$, 
$q_2=0.99$, with the herding parameters $\ep_1=\ep_2=1$. 
(a) Continuous solutions $p_1(t)$ (solid line) and $p_2(t)$ (dash-dotted line) 
at the beginning of the process for $t \in [0, 30]$; 
(b) Discrete solutions $p_1(t)$ (solid line) and $p_2(t)$ (dash-dotted line) 
at the beginning of the process for $t \in [0, 30]$. 
}
\label{fig:Fig.13}
\end{figure}

\begin{figure}[ht]
\centerline{
\hbox{\includegraphics[width=7.5cm]{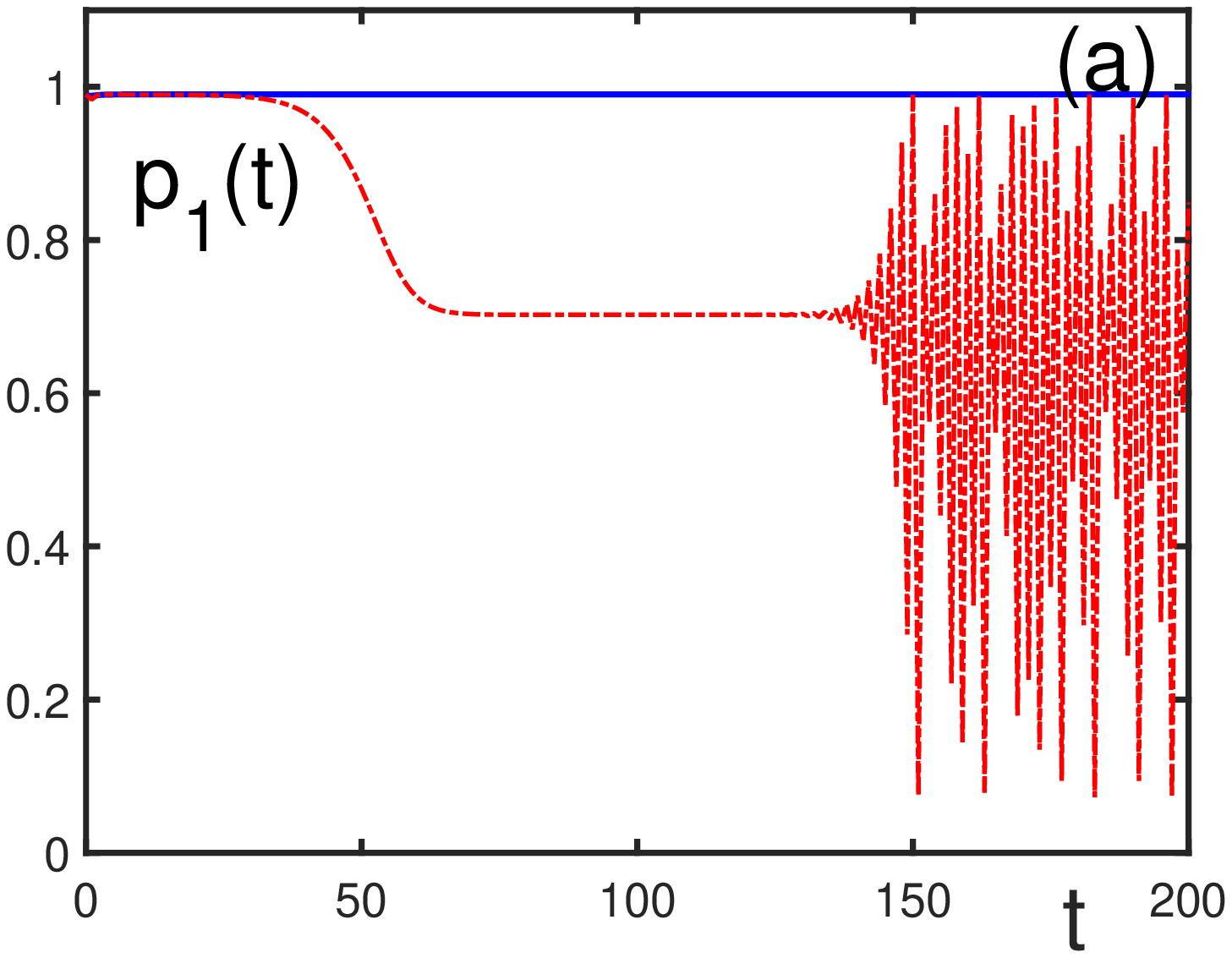} \hspace{0.5cm}
\includegraphics[width=7.5cm]{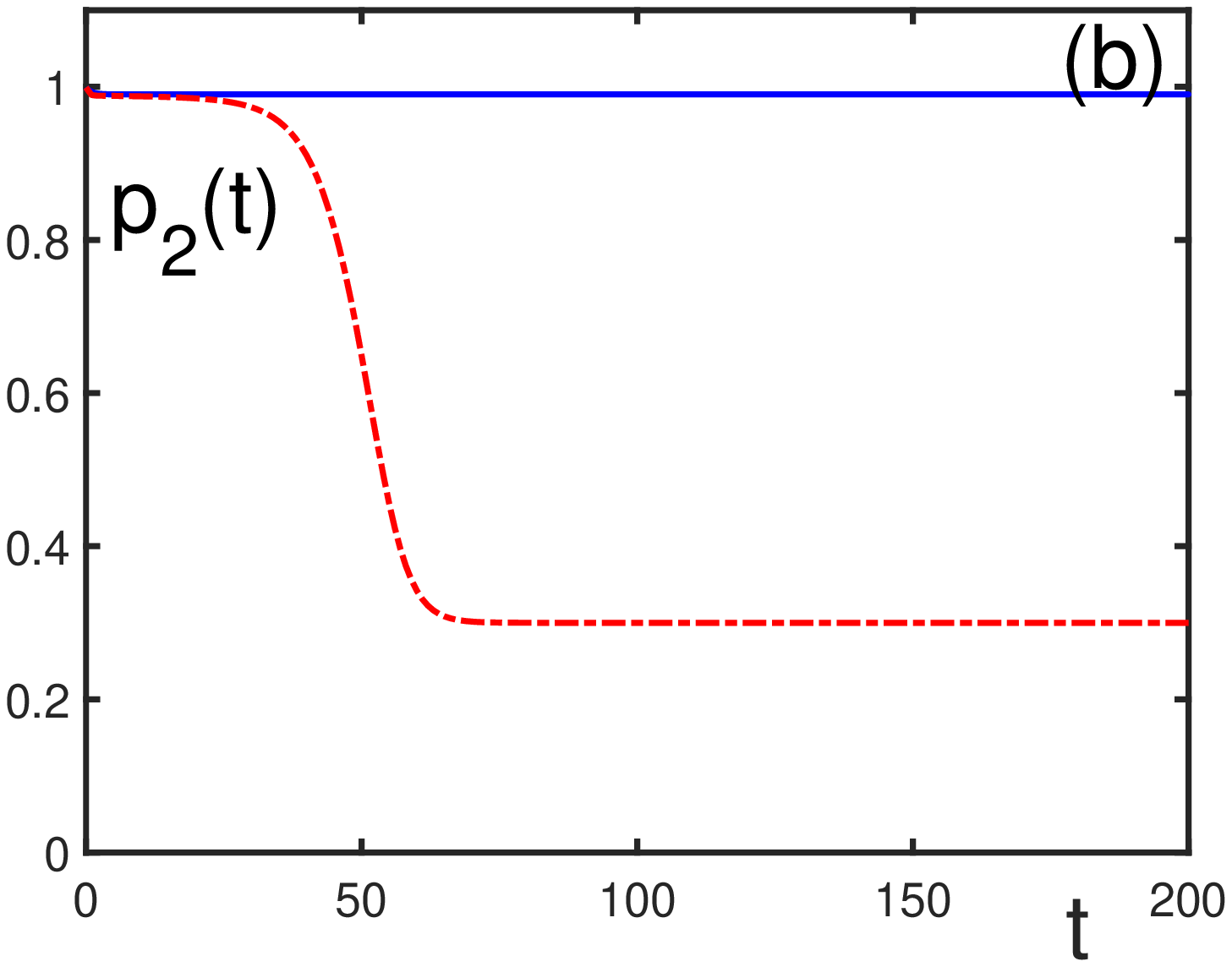}  } }
\caption{\small Comparison of the solutions for the probabilities $p_1(t)$ 
and $p_2(t)$ for the discrete and continuous decision making. The initial 
conditions are $f_1=0.3$, $q_1=0.699$ and $f_2=0$, $q_2=0.99$, with the 
herding parameters $\ep_1=\ep_2=1$. 
(a) Solution $p_1(t)$ to the continuous equations (solid line), as compared 
to $p_1(t)$ for the discrete equations (dash-dotted line). The continuous 
probability tends to the fixed point $p_1^*=f_2+q_2=0.99$, while the discrete 
probability chaotically oscillates; 
(b) Solution $p_2(t)$ to the continuous equations (solid line), as compared 
to $p_2(t)$ for the discrete equations (dash-dotted line). The continuous 
probability tends to the fixed point $p_2^*=f_2+q_2=0.99$, while the discrete 
probability tends to $p_2^* = f_1 = 0.3$.
}
\label{fig:Fig.14}
\end{figure}

Thus, the solutions to the continuous equations are always smooth and exhibit 
neither periodic oscillations nor chaotic fluctuations that exist for the discrete 
equations. The considered equations describe the process of decision making in 
a network of intelligent agents. The discrete equations correspond to multistep 
decision making, as it can be done in real life. The continuous equations serve 
as an approximation to the studied problem. This approximation is rather accurate 
when the discrete solutions are smooth. However, the continuous approximation does 
not catch such interesting cases as periodic oscillations and chaotic fluctuations. 
These oscillations and fluctuations can arise in realistic problems, where there 
are the groups of agents with different features, for instance when there occurs 
the co-existence of agents with different types of memory, say long-term and 
short-term memory.

\subsection{Time discounting of utility}

In the dynamical processes of preferences, considered above, the utility factor 
was treated as a constant quantity. This is a reasonable assumption based on the 
fact that the period of time when utility does not change is usually much longer 
than the time of information exchange between the agents of the network. For 
instance, prices often are not changed during years or at least months, while 
the information exchange discussing these prices can occur several times a day.

However, in order to have general understanding, let us consider how the utility 
factors could vary when taking account of time discounting of utility. Then we 
need to deal with discounted expected utility \cite{Frederick_2002,Read_2000}. 
In that case, the utility function $u(x)$ for a payoff $x$ is discounted at time 
$t$ as $D(t) u(x)$. Here $D(t)$ is a discount function with the properties
$$
0 < D(t) < 1 \qquad ( 0 < t < \infty) \; ,
$$
\be
\label{5.43}
D(0) = 1 \; , \qquad D(\infty) = 0 \; .
\ee
A typical example of a discount function is $D(t) = 1/(1+r)^t$, where $r$ is a 
discount rate and $t$ is a discrete time counted in some temporal units, usually 
in units of an year. With the discounted utility functions, the expected utility 
of a lottery $L_n$ becomes discounted as $D(t)U(L_n)$. Denoting the discounted 
expected utility of an alternative $A_n$ as
\be
\label{5.44}
U(A_n,t) = U(A_n) D(t) \; ,
\ee
for the discounted utility factor (\ref{3.25}), we obtain
\be
\label{5.45}
 f(A_n,t) = 
\frac{U(A_n,t)\exp\{\bt U(A_n,t)\} }{\sum_n U(A_n,t)\exp\{\bt U(A_n,t)\} } \; .
\ee

Using here the discounted utility (\ref{5.44}) gives
\be
\label{5.46}
f(A_n,t) = 
\frac{U(A_n)\exp\{\bt(t)U(A_n)\} }{\sum_n U(A_n)\exp\{\bt(t)U(A_n)\} } \; ,
\ee
with the initial expected utilities and the discounted belief function
\be
\label{5.47}
  \bt(t) = \bt D(t) 
\ee
enjoying the properties
$$
0 < \bt(t) < \bt \qquad ( 0 < t < \infty ) \; ,
$$
\be
\label{5.48}
 \bt(0) = \bt \; , \qquad \bt(\infty) = 0 \; .
\ee
As we see, the utility factor can be considered to be constant if $t \ll 1$, 
when $\beta (t) \simeq \beta$, if $t \gg 1$, when $\beta(t) \simeq 0$, and also 
if $\beta \ll 1$, when $\beta(t) \simeq 0$. 

In that way, the expected utilities are discounted with time, however they enter  
the utility factors so that the latter are almost constant for wide time periods,
and they do not vary at all for the neutral beliefs of decision makers, when
$\beta = 0$. Thus the assumption that the utility factor is practically constant 
has the wide range of applicability.

\subsection{Collective network operation}

The functioning of a network can be characterized in two ways, by considering the
actions of its separate parts and by studying the operation of the network as a whole.
Under a network, we can imply a society with the population structured into several
different groups or an artificial intelligence consisting of several groups of nodes
with different properties. That is, it is possible to study the decision making of
parts of a society or the collective decision making by the whole society. Otherwise,
it is possible to concentrate on the actions of different parts of a network or on
the collective functioning of the whole network.

Suppose, we consider a society (network) of $N$ agents (nodes). The society is
structured into $G$ groups (parts) enumerated by $j=1,2,\ldots,G$. Each group is 
characterized by its specific properties, say its type of memory and herding parameters. 
The group $j$ consists of $N_j$ agents, so that the group $j$ composition fraction is 
\be
\label{5.49}
c_j \equiv \frac{N_j}{N} \qquad ( j = 1,2, \ldots, G) \;   .
\ee 
Hence the normalization conditions are
\be
\label{5.50}
 \sum_{j=1}^G N_j = N \; , \qquad  \sum_{j=1}^G c_j = 1 \;  .
\ee

As early, the number of agents in a group $j$ choosing an alternative $A_n$ at time 
$t$ is $N_j(A_n,t)$. Since each agent of a group chooses one of the alternatives,
\be
\label{5.51}
 \sum_{n=1}^{N_A} N_j(A_n,t) = N_j \;   .
\ee
The probability that an agent of a group $j$ chooses an alternative $A_n$ at time 
$t$ is
\be
\label{5.52}
 p_j(A_n,t) \equiv \frac{N_j(A_n,t)}{N_j} \;  .
\ee

The number of agents in the whole society choosing the alternative $A_n$ reads as
\be
\label{5.53}
 N(A_n,t) = \sum_{j=1}^G N_j(A_n,t) \;  .
\ee
Then the probability that an agent of the whole society chooses an alternative $A_n$
at time $t$ is defined as
\be
\label{5.54}
  p(A_n,t) \equiv \frac{N(A_n,t)}{N} \;  .
\ee
The related normalization condition is
\be
\label{5.55}
\sum_{n=1}^{N_A} p(A_n,t) = 1 \;  .
\ee

In the above subsections, the temporal behaviour of the group probabilities 
(\ref{5.52}) has been investigated. The collective decision of a society, or the 
collective action of a network is described by the probability (\ref{5.54}). 
The group probabilities define the total network probability  
\be
\label{5.56}
p(A_n,t) = \sum_{j=1}^G c_j p_j(A_n,t) \;  ,
\ee
where the structure of the society (network) is prescribed by the given composition 
fractions $c_j$. The collective probability (\ref{5.56}) characterizes the behaviour 
of the society (network) as a whole.

\section{Conclusion}

The review discusses the principal points of how the operation of artificial 
intelligence has to be organized in order to simulate the so-called human-level 
artificial intelligence that is able to make decisions taking into account the 
utility of the considered alternatives as well as emotions accompanying the choice. 
In order to formalize the operation of such an affective intelligence, it is 
necessary to characterize the basic rules of human decision making. 

First of all, it is required to understand the general structure of affective 
decision making. The pivotal idea suggesting the main principles of this structure 
is the formal similarity between decision making under emotions and quantum 
measurements under intrinsic noise. Cognition-emotion duality in decision making 
is formally analogous to observable-noise duality in quantum measurements.    
   
It is in no way assumed that the brain is a quantum object or consciousness is 
in any sense quantum. The analogies between quantum measurements and decision 
making are purely nominal. It is possible to adduce a number of examples where 
physically very different processes are characterized by very close mathematics. 
For instance, many processes of self-organization in complex systems can be 
interpreted as decision making \cite{Yukalov_127}. As an example of two different 
in nature phenomena characterized by very similar mathematical description, it is 
possible to mention that the development and arising distribution of cities over 
their size can be described as the process of Bose-Einstein condensation 
\cite{Yukalov_138}. Although the growth of cities and Bose-Einstein condensation,
by their nature, are absolutely different processes, they can be described by 
similar mathematics. In the same vein one has to accept the analogies between 
quantum measurements and decision making. These analogies turned out to be 
rather productive suggesting the principal points governing the description of 
affective decision making.

To be concrete, Sec. 2 presents the general theory of quantum measurements under 
intrinsic noise. The goal is to analyze what ideas and methods could be accepted 
for the design of affective decision theory. However, there is no need to involve 
all that quantum machinery for decision theory because of the following:

\vskip 2mm
(a) From one side, quantum theory is overloaded by numerous notions having no 
direct meaning in decision theory. Thus, it is not defined what in decision 
making would be the form of Hamiltonians, of statistical operators, and so on. 

\vskip 2mm
(b) From the other side, the notions of quantum theory are not sufficient for 
suggesting quantitative theory of decision making. This is why the use of quantum 
techniques for decision making has produced practically no quantitative description
and no quantitative predictions. In order to develop a quantitative theory of 
decision making, one has to complement quantum formulas by quite a number of 
assumptions of purely classical origin and introduce several fitting parameters. 

\vskip 2mm
(c) In addition, involving rather complicated techniques of quantum theory, 
complementing them by a number of classical assumptions, and equipping with 
fitting parameters seems to be excessive if a simpler classical description is 
admissible.           
 
\vskip 2mm  
Therefore, borrowing some general ideas from the theory of quantum measurements 
under intrinsic noise, it looks beneficial to formulate an axiomatic approach to 
affective decision making in purely classical terms, given in Sec. 3. The ideas 
of quantum measurement theory serve rather as Ariadne's thread for formulating 
affective decision theory that can be used by artificial intelligence. It is in 
that sense the term ``quantum operation of artificial intelligence" is used. The 
intelligence is not quantum, but its operation imitates some rules specific for 
quantum theory. The main points borrowed from the theory of noisy measurements are:

\begin{enumerate}[label=(\roman*)]
\item
Decision making has to be treated as a probabilistic process that needs to be 
characterized by behavioural probabilities defining the fractions of decision 
makers choosing the related alternatives from a given set. 
  
\item
The role of emotions in decision making is similar to the role of intrinsic 
noise under quantum measurements.

\item
Behavioural probability, taking account of cognition-emotion duality, 
is represented by the superposition of two terms, a rational utility factor, 
measuring the utility of the alternatives, and an attraction factor, characterizing 
the role of emotions in the process of decision making. In particular cases, 
the attraction factor can be split into several terms responsible for different 
types of emotions. 

\item
Alternatives and accompanying emotions are entangled in the sense that they are 
connected with each other and both of them influence the choice.

\item
Different types of emotions interfere with each other producing in the behavioural 
probability the term named attraction factor.

\item
It is necessary to distinguish the joint probability of two consecutive 
events occurring at different moments of time and the joint probability of two 
events defined as a synchronous probability of two events at different locations. 
This latter probability, similarly to the probability of a single event, is 
represented by a sum of utility factor and attraction factor, if emotions are 
taken into account.

\item
Temporal dependence of behavioural probabilities can be modeled by the form 
arising from the consideration of time dependence of the averages corresponding 
to observable quantities under non-destructive repeated quantum measurements. 
\end{enumerate}

To have predictive power, these principal items can be complemented by 
non-informative priors characterizing the behaviour of a typical decision maker. 
All elaboration of affective decision theory can be given in classical terminology, 
with no quantum formulas explicitly implicated. The behaviour of real decision 
makers on the aggregate level is quantitatively well described by the formulated 
affective decision theory, using no fitting parameters. This is confirmed by 
treating a set of lotteries in Sec. 3. 

The famous behavioural paradoxes plaguing classical decision making find their 
natural resolution in the frame of the affective decision theory taking account 
of emotions. The paradoxes are not merely qualitatively explained but are in 
good quantitative agreement with empirical data on the aggregate level, without 
involving any fitting parameters.   

A family of affective intelligent agents composes a network where the agents 
interact through information exchange. The intelligent agents can differ by 
their initial conditions and by the type of memory, in the ultimate cases being 
long-term or short-term memory. There can exist the networks of agents with uniform 
memory, that is having the same type of memory, but different initial conditions. 
The most interesting and rich is the behaviour of a mixed network, where a part of 
agents enjoys long-term memory, while the other part has short-term memory. In 
these networks there can arise self-excited waves of preferences demonstrating 
either periodic or chaotic dynamics. Chaotic decision making reminds us the 
behaviour of patients suffering from mental diseases.

Discussing the multi-step decision making of society agents, we can keep in mind
either a human society composed of different groups or a network consisting of
several parts of nodes, or the brain composed of neurons.   

The described approach, from one side, helps understanding the behaviour of real 
humans making a choice between alternatives by evaluating their rational utility 
at the same time being subject to accompanying irrational emotions. From the other 
side, the formalized rules of the choice followed by a typical decision maker can 
help for implicating the related algorithms in the operation of a complex Affective 
Artificial Intelligence.

\section*{Acknowledgment}

I am very grateful for many useful discussions to D. Sornette and E.P. Yukalova. 
A part of the material composing this review has been presented in my lectures 
I have been giving for several years in Swiss Federal Institute of Technology 
(ETH Zurich), Zurich, Switzerland.   

\vskip 3mm

This research did not receive any specific grant from funding agencies in the public, 
commercial, or not-for-profit sectors.

\end{document}